\documentclass[MAL,biber]{nowfnt} 

\usepackage[utf8]{inputenc}
\usepackage{macros}
\usepackage{comment}

\title{Risk-Sensitive Reinforcement Learning 
\diff{via Policy Gradient Search}
}

\subtitle{}

\maintitleauthorlist{
	Prashanth L. A. \\
	Department of Computer Science and Engineering,\\ Indian Institute of Technology Madras. \\
	prashla@cse.iitm.ac.in
	\and
	Michael C. Fu \\
	Robert H. Smith School of Business \& \\Institute for Systems Research,\\
	University of Maryland, College Park. \\
	mfu@umd.edu
}

\issuesetup
{%
	copyrightowner={A.~Heezemans and M.~Casey},
	volume        = xx,
	issue         = xx,
	pubyear       = 2022,
	isbn          = xxx-x-xxxxx-xxx-x,
	eisbn         = xxx-x-xxxxx-xxx-x,
	doi           = 10.1561/XXXXXXXXX,
	firstpage     = 1, 
	lastpage      = xx
}

\usepackage{mwe}

\author[1]{L. A., Prashanth}
\author[2]{Fu, Michael C.}

\affil[1]{Indian Institute of Technology Madras; prashla@cse.iitm.ac.in}
\affil[2]{University of Maryland, College Park; mfu@umd.edu}

\articledatabox{\nowfntstandardcitation}

\begin{document}

\chapter*{Preface}
\thispagestyle{empty}
\vspace*{-100pt}

\diff{
Reinforcement learning (RL) is one of the foundational pillars of artificial intelligence and machine learning.
An important consideration in any optimization or control problem is the notion of risk, but its incorporation into RL has been a fairly recent development. 
This book surveys research on risk-sensitive RL that uses policy gradient search, i.e., 
policy optimization in a stochastic formulation, as opposed to robust optimization approaches and methods that focus on the value function. 

We have tried to make the exposition completely self-contained but also organized in a manner that allows expert readers to skip background chapters.
In particular, those readers already familiar with 
Markov decision processes (MDPs), risk measures, and stochastic gradient-based search (specifically, stochastic approximation) can skip Chapters 
\ref{sec:mdps},  
\ref{sec:risk-measures}, 
and
\ref{sec:background}, 
respectively. 

We have benefited from the feedback of many who read earlier drafts of the manuscript. 
We begin by thanking Prof. Vivek Borkar, who generously offered valuable detailed comments regarding the content, and provided material and references for the sections on the exponential cost formulation. 
Next, we thank Prof. Shalabh Bhatnagar for helpful discussions on the convergence analysis in the risk-as-constraint setting, 
and Prof. Armand Makowski for critical observations.
%
We'd also like to thank two anonymous reviewers, whose comments and suggestions helped us improve the exposition considerably. 
Lastly, we thank several of our Ph.D. students --- Xingyu Ren, Erfaun Noorani, Mehrdad Moharrami, Nithia Vijayan, Yi Zhou, and Mengting Chao, who read through various portions and stages of the manuscript and caught numerous typos. 
Any remaining errors are of course our responsibility alone. 

One final note: We have chosen to include references at the end of the chapter in bibliographic remarks rather than cite them in the main text, so as not to interrupt the expositional flow. 

}
\clearpage

\makeabstracttitle

\begin{abstract}
\diff{
The objective in a traditional reinforcement learning (RL) problem is to find a policy that optimizes the expected value of a performance metric such as the infinite-horizon cumulative discounted or long-run average cost/reward. In practice, optimizing the expected value alone may not be satisfactory, in that it may be desirable to incorporate the notion of risk into the optimization problem formulation, either in the objective or as a constraint. 
}
Various risk measures have been proposed in the literature, e.g., exponential utility, variance, percentile performance, chance constraints, value at risk (quantile), conditional value-at-risk, prospect theory and its later enhancement, cumulative prospect theory. 
In this \diff{book}, 
we consider risk-sensitive RL in two settings: one where the goal is to find a policy that optimizes the usual \diff{expected value objective while ensuring that a risk constraint is satisfied, and the other where the risk measure is the objective.
We survey some of the recent work in this area specifically where policy gradient search is the solution approach. 
}
In the first risk-sensitive RL setting, we cover popular risk measures based on variance, conditional value-at-risk, and chance constraints, and present a template for policy gradient-based risk-sensitive RL algorithms using a Lagrangian formulation.
\diff{
For the setting where risk is incorporated directly into the objective function, we consider an exponential utility formulation, cumulative prospect theory, and coherent risk measures.  
}
This non-exhaustive survey aims to give a flavor of the challenges involved in solving risk-sensitive RL problems \diff{using policy gradient methods}, as well as outlining some potential future research directions. 
\end{abstract}

\chapter{Introduction}
\label{sec:intro}

Markov decision processes (MDPs) 
provide a general framework for modeling a wide range of problems involving sequential decision making under uncertainty, which arise in many areas of applications, such as transportation, computer/communication systems, manufacturing, and supply chain management.
MDPs transition from state to state probabilistically over time due to chosen actions taken by the decision maker, 
incurring state/action-dependent costs/rewards at each instant.
The goal is to find a policy (sequence of decision rules) for choosing actions that optimizes a long-run objective function, e.g., the cumulative sum of discounted costs or the long-run average cost.

The traditional MDP setting assumes that 
(i) the transition dynamics (probabilities) and costs/rewards are fully specified/known, and 
(ii) the objective function and constraints involve standard expected value criteria.  
However, in a myriad of settings of practical interest, neither of these conditions holds, i.e., 
only \textbf{\textit{samples}} of transitions (and costs/rewards) can be observed (e.g., in a black-box simulation model or an actual system) 
and/or performance measures that incorporate \textbf{\textit{risk}} \diff{really need to be considered in the problem.} 
In the case of the former, reinforcement learning (RL) techniques can be employed,  
and in the latter setting, risk-sensitive approaches are appropriate. 
Although there is abundant research on both of these settings dating back decades, the work combining both aspects is more recent. 
Furthermore, the two settings have been predominantly pursued independently by different research communities, 
with RL a focus of CS/AI researchers and risk-sensitive MDPs a focus of stochastic control and operations research/management science/mathematical finance researchers.  

\diff{
\section*{Why risk? (Avoid merely expectations?)}
\vspace*{-4pt}
The focus of this monograph is not on why risk is important nor on what is the best way to incorporate it into decision making 
but rather on finding good risk-sensitive policies via RL policy gradient algorithms. 
However, to provide some motivation for incorporating risk into decision making, we briefly describe two everyday illustrative examples. 
The first example has to do with financial investments, where the primary objective is generally to \textit{maximize \textbf{expected} return}. 
Clearly, this is not sufficient for most decision makers, who would very much like to take into consideration the ``risk'' of the investments, in this case taken to mean mitigating the potential downside losses. 
The second example is your commute to work (which, as we finalize this manuscript, may seem less relevant in the midst of a pandemic). 
In this case, your primary objective is likely to \textit{minimize \textbf{expected} travel time}. 
However, if you have an important early morning meeting, you might want to reduce the ``risk'' of being late by choosing an alternative that has a higher expected travel time but is unlikely to suffer a huge delay from an unexpected but rare event such as an overturned tractor-trailer. 
A colleague of ours avoids taking the highway to/from work for this very reason (along with safety considerations). 
In other words, most decision makers consider more than merely expectations. 
Both of these examples also serve to illustrate the more general observation that real-world decisions involve multiple objectives, where at least one of them involves the notion of risk, extending beyond the usual expected value performance measures 
considered in standard MDP and RL models (including commonly used metrics for analysis purposes such as expected regret in multi-armed bandit models). 

\section*{Types of risk and ways to incorporate risk}
\vspace*{-4pt}
As in any multi-objective optimization problem, there are many ways to incorporate risk. Again, our focus is not on advocating for one formulation over another, but to provide several different alternatives, with a solution approach for each of them. 
Which formulation is ``better'' will depend on both the problem and the problem solver(s). 
We illustrate this concept by revisiting our two examples. 

One way to address risk in the investment problem is to minimize some measure of volatility, which could take the form of putting an upper bound on the {\it variance} of return. 
Thus, the decision problem becomes a constrained optimization of maximizing the objective of expected return {\it subject to a constraint} on the variance of return. 
This is the classic mean-variance portfolio optimization problem in finance for which Harry Markowitz was awarded the 1990 economics Nobel Prize. 

It can be easily argued that variance is not the best measure of risk for this problem, since it also penalizes excessive upside moves, so maybe focusing on one tail (the downside risk) is more appropriate. 
One way to address this would be to limit the probability of a high loss to some acceptable level such as 5\% or 1\% or even smaller. This is known as a {\it chance constraint}.
Conversely, one might have an upper bound on the amount of loss that might occur at a certain low probability, i.e., putting a constraint on a quantile of the loss distribution, which the financial industry defines as {\it value-at-risk} (VaR). 
A more sophisticated extension of VaR is {\it conditional value-at-risk} (CVaR), which also has some other nice properties that VaR does not, most notably that it is a {\it coherent risk measure}. 
Exponential utility is another way of capturing risk preferences and implicitly capturing higher moments beyond the second moment. 
Chapter \ref{sec:risk-measures} provides a more formal review of all of these risk concepts and metrics. 

Similarly, revisiting risk in the commuting problem where the objective is to minimize travel time,  
a constrained optimization problem formulation would be to minimize expected travel time subject to an upper bound on the variability of travel time, or alternatively, one could instead employ a chance constraint by specifying the probability of the travel time exceeding an acceptable threshold, e.g., requiring that at least 99\% of the time the travel time will be less than an hour.  

Realistic problems may involve multiple constraints that need to be satisfied concurrently, such as bounds on both the variability and the probability of a rare event. 
In our setting, this can be easily handled, but for the sake of simplicity we will only explicitly consider the case of a single constraint, as the extension using the policy gradient approach would just involve additional Lagrange multiplier gradient estimates, but the general approach would be the same. 

Finally, rather than formulating the problem with risk as a constraint, another approach is to try and include it in the objective function. 
Perhaps the simplest way would be as a weighted combination of the multiple objectives. 
While we don't address the weighed objectives formulation explicitly, it should be clear how it could also be handled as an easy special case using the techniques of this monograph.
Instead, we consider more general formulations: the use of expected utility (an exponential cost formulation), which modifies the output performance measure (corresponding to investment return or travel time in the two examples), and a risk measure called \textit{cumulative prospect theory} (CPT) that ``distorts'' the perceived probabilities due to the decision maker's view of the world. 
Demonstrating that prospect theory and CPT are able to model certain aspects of actual observed human behavior that utility theory was unable to capture was a key contribution for which (behavioral psychologist) Daniel Kahneman was awarded the 2002 economics Nobel Prize. 
Our treatment also extends the CPT formulation to a framework encompassing general coherent risk measures.
}

\vspace*{-3pt}
\section*{Objectives of this book}
\vspace*{-6pt}
\diff{
The main purpose of this book is to  
introduce and survey research results on \textit{\textbf{policy gradient methods for reinforcement learning with risk-sensitive criteria}}, 
as well as to outline some promising avenues for future research following the risk-sensitive RL framework. 
We consider both constrained formulations where the traditional expected value performance measure is augmented with a risk constraint 
and problem formulations where the risk measure is explicitly in the objective function being optimized.  
Some well-known examples of risk measures to be considered as constraints, 
most of which were illustrated by the two earlier examples,}
include variance (or higher moments), probabilities (in the form of chance constraints), 
quantiles or value-at-risk (VaR), and conditional value-at-risk (CVaR).
\diff{
As also mentioned in the examples, risk measures used explicitly as the objective function include exponential utility 
and some very recent work on using CPT with RL.
}

\diff{
To be specific, the constrained risk-sensitive RL problem will be an optimization problem of the following general form:
}
\vspace*{-4pt}
\boxed{
\begin{equation}
\min_{\theta \in \Theta} J(\theta) \triangleq \E\left[ D(\theta)\right]
\text{ subject to }\quad G(\theta)\leq\kappa,
\label{eq:gen-constrained-opt}
\end{equation}}
\vspace*{-2pt}
where $\theta$ denotes the policy parameter, $\Theta$ represents the policy space, 
\diff{
$D(\theta)$ is a (stochastic) cost function, $G(\theta)$ is a risk measure, and $\kappa$ denotes the acceptable risk level. 
In the MDP setting, the quantities may also depend on the initial state of the MDP, which is not indicated here.  
The most common choices for $D(\theta)$ in the MDP setting include the infinite-horizon cumulative discounted cost, total cost in a stochastic shortest path problem, and the long-run average cost.
Note that we will be minimizing cost (as in the commuting example), which is more common in MDP formulations than in the RL setting, which often focuses on maximizing reward (as in the investment example). 
The classic ``risk-neutral'' formulation  simply minimizes $J(\cdot)$ without the risk constraint in \eqref{eq:gen-constrained-opt}. 
Also, in contrast to the traditional setting of risk-sensitive control where $J$ and $G$ functions are analytically available in the MDP model, 
in the RL setting, $J$ and $G$ are unknown or cannot be calculated directly, but noisy estimates of $J$ and $G$ are available, e.g., samples of $D$ could provide an unbiased estimator of $J$. 
Thus, as in the usual RL setting, traditional MDP techniques cannot be applied, whereas RL algorithms suitably adapted provide one avenue to attack such risk-sensitive MDPs, i.e., a setting when the MDP model is unknown and all the information about the system is obtained from samples resulting from the decision maker's interaction with the environment. 
}

We propose to solve the constrained optimization problem \eqref{eq:gen-constrained-opt} by performing gradient descent search on the Lagrangian objective function.  
As depicted in Figure \ref{fig:algorithm-flow1},  
the risk-sensitive policy gradient algorithm requires estimators $\widehat \nabla J(\theta)$, $\widehat \nabla G(\theta)$ and $\widehat G(\theta)$ of $\nabla J(\theta)$, $\nabla G(\theta)$ and $G(\theta)$, respectively. Then,  two-timescale gradient-based search algorithms taking the following form will be developed (where $\lambda$ is the Lagrange multiplier to be optimized along with the policy parameter $\theta$):
\boxed{
	\vspace*{-10pt}
	\begin{align*}
		\lambda_{n+1} &= \left[\lambda_n + \zeta_1(n) \left(\widehat G(\theta_n) - \kappa\right)\right]^+, \label{eq:lambda-ascent}  \\
		\theta_{n+1} &= \Gamma \left[\theta_n - \zeta_2(n)  \left(\widehat\nabla J(\theta_n) + \lambda_n \widehat \nabla G(\theta_n)\right)\right],          
\end{align*}}
where $[x]^+=\max(0,x)$,
$\Gamma$ is a projection into $\Theta$, 
and
$\{\zeta_1(n), \zeta_2(n)\}$ are step-size sequences selected such that the $\theta$ update is on the faster timescale and the $\lambda$ update is on the slower timescale (see Section \ref{sec:template-constrained} for details). 

\begin{figure*}
	\centering
	\tikzstyle{block} = [draw, fill=white, rectangle,
	minimum height=5em, minimum width=6em]
	\tikzstyle{sum} = [draw, fill=white, circle, node distance=1cm]
	\tikzstyle{input} = [coordinate]
	\tikzstyle{output} = [coordinate]
	\tikzstyle{pinstyle} = [pin edge={to-,thin,black}]
	\scalebox{0.5}{\begin{tikzpicture}[auto, node distance=2cm,>=latex']
			\node (theta) at (-6,0) {$\bm{\theta_n,\lambda_n}$};
			\node [block,minimum height=7em, minimum width=7em, fill=blue!20,right=2cm of theta,label=below:{\color{bleu2}\bf Data Collection},align=center] (sample) {\textbf{Using policy $\mu^{\theta_n}$, }\\[1ex] \textbf{sample/simulate  }\\[1ex]\textbf{underlying system}}; 
			\node [block, fill=green!20,above right=3cm of sample,label=below:{\color{darkgreen!90}\bf Policy Gradient},align=center] (obj) {\textbf{Estimate} $\bm{\nabla J(\theta)}$};
			\node [block, fill=green!20,right=2cm of sample,label=below:{\color{darkgreen!90}\bf Risk Estimation},align=center] (estcvar) {\textbf{Estimate} $\bm{G(\theta)}$};
			\node [block, fill=green!20,below right=3cm of sample,label=below:{\color{darkgreen!90}\bf Risk Gradient},align=center] (cvar) {\textbf{Estimate} $\bm{\nabla G(\theta)}$};
			\node [block, fill=red!20,right=8cm of sample, minimum height=8em,label=below:{\color{violet!90}\bf Policy Update},align=center] (update) {\textbf{Update} $\bm{\theta_n}$  \\[1ex] \textbf{Update } $\bm{\lambda_n}$ };
			\node [right=2cm of update] (end) {$\bm{\theta_{n+1},\lambda_{n+1}}$};
			\draw [thick,-triangle 45] (theta) --  (sample);
			\draw [thick,-triangle 45] (sample) -- (obj);
			\draw [thick,-triangle 45] (sample) -- (estcvar);
			\draw [thick,-triangle 45] (sample) -- (cvar);
			\draw [thick,-triangle 45] (obj) -- (update);
			\draw [thick,-triangle 45] (estcvar) -- (update);
			\draw [thick,-triangle 45] (cvar) -- (update);
			\draw [thick,-triangle 45] (estcvar) -- (cvar);
			\draw [thick,-triangle 45] (update) -- (end);
	\end{tikzpicture}}
	\caption{Schematic of risk-sensitive policy gradient algorithm for constrained optimization (underlying system could be a simulation model or a real system).}
	\label{fig:algorithm-flow1}
\end{figure*}

\diff{In addition to the risk-constrained problem (\ref{eq:gen-constrained-opt}), we also consider a risk-sensitive problem where the risk measure is explicitly incorporated into the objective function, i.e., the following optimization problem:
\vspace*{-4pt}
\boxed{
\begin{equation}
\min_{\theta \in \Theta} G(\theta),
\label{eq:gen-unconstrained-opt}
\end{equation}}
\vspace*{-2pt}
where $G$ is a risk objective function involving exponential utility, CPT, or a coherent risk measure. 
}
For solving the problem (\ref{eq:gen-unconstrained-opt}), we propose a policy gradient algorithm that incorporates the following iterative update:
\boxed{
\vspace*{-10pt}
	\begin{align*}
		\theta_{n+1} &= \Gamma\big[\theta_n - \zeta(n)  \widehat \nabla G(\theta_n)\big],
\end{align*}} 
where $\{\zeta(n)\}$ is a step-size sequence,  $ \widehat \nabla G(\theta_n)$ is an estimate of $\nabla G(\theta_n)$, and $\Gamma$ is a projection operator that keeps the iterate $\theta_n$ bounded within the set $\Theta$ as in the case of the risk-constrained policy gradient algorithm above 
(see Section \ref{sec:template-unconstrained} for details).

\section*{Challenges in risk-sensitive RL}
\vspace*{-4pt}
Risk-sensitive RL is generally more challenging than its risk-neutral counterpart. For instance, for a discounted-cost MDP, there exists a Bellman equation for the variance of the return, but the underlying Bellman operator is not necessarily monotone, so that policy iteration is no longer guaranteed to lead to an optimal policy.
Moreover, finding a globally mean-variance optimal policy in a discounted-cost MDP is NP-hard, even in the classic MDP setting where the transition model is known.
Average-cost MDP problems also are generally NP-hard, e.g., consider a risk measure that is not the plain variance of the average cost and instead is a variance of a quantity that measures the deviation of the single-stage cost from the average cost.  
Finally, in comparison to variance/CVaR, CPT is a non-coherent and non-convex measure, ruling out the usual Bellman equation-based dynamic programming (DP) approaches when optimizing the MDP CPT-value. 

The computational complexity results summarized in the previous paragraph imply that finding guaranteed global optima of  
\diff{risk-sensitive MDP formulations described by \eqref{eq:gen-constrained-opt} or \eqref{eq:gen-unconstrained-opt}}
is not computationally practical, motivating the need for algorithms that approximately solve such MDP formulations. 
\diff{
{\it In this book, we focus on policy gradient-type learning algorithms where the policies are parameterized in a continuous space, and an iterative search for a better policy occurs through a gradient-descent update.} 
}
 Actor-critic methods are a popular subclass of policy gradient methods and were among the earliest to be investigated in RL. They are comprised of an {\em Actor} that improves the current policy via gradient descent (as in policy gradient schemes) and a {\em Critic} that incorporates feature-based representations to approximate the value function. The latter approximation is necessary to handle the curse of dimensionality. Regular policy gradient schemes usually rely on Monte Carlo methods for policy evaluation, an approach that suffers from high variance as compared to actor-critic schemes. On the other hand, function approximation introduces a bias in the policy evaluation.
A policy gradient/actor-critic scheme with provable convergence to a locally risk-optimal policy would require  careful synthesis of techniques from stochastic approximation, stochastic gradient estimation approaches, 
and importance sampling. 

Several of the constituent solution pieces require significant research for various risk measures. 
For example, consider the ``policy evaluation'' part of the overall algorithm in a risk-sensitive MDP, which requires estimating $J(\theta)$ and $G(\theta)$, given samples obtained by simulating the MDP with policy $\theta$. If $J(\theta)$ is one of the usual MDP optimization objectives such as discounted total cost, long-run average cost, or total cost (in a finite-horizon MDP), then estimating $J(\theta)$ can be performed using one of the existing algorithms. Temporal difference (TD) learning 
is a well-known algorithm that can learn the objective value along a sample path for a given $\theta$. 
However, estimating $G(\theta)$ using TD-type learning algorithms is infeasible in many cases. For instance, consider variance as the risk measure in a discounted-cost MDP. In this case, even though there is a Bellman equation, 
the operator underlying this equation is not monotone, ruling out a TD-type learning algorithm.
More recently, CVaR-constrained MDPs have been considered,  
though a variance-reduced CVaR estimation algorithm is still needed. 
In other words, there is no algorithm in an RL context that incorporates a variance reduction technique such as importance sampling and is provably convergent. 
Note that variance reduction is necessary, because CVaR is based on the tail of the distribution. 

Going beyond the prediction problem, designing policy gradient algorithms is challenging for a risk-sensitive MDP, as it requires estimating the (policy) gradient of the risk measure considered, a nontrivial task in the RL context. 
\diff{For instance, in a  discounted-cost MDP context, the policy gradient theorem variant that accounts for the variance of the cumulative discounted cost does not lend itself to an RL algorithm. An alternative is to apply a finite differences method such as simultaneous perturbation stochastic approximation (SPSA), which amounts to treating the MDP as a black box, and such an approach would ignore the underlying Markovian structure of the problem, which is the case with the existing policy gradient algorithms for optimizing the CPT-value in any of the MDP settings.}


\section*{Outline of the remaining chapters}
\vspace*{-4pt}
\diff{
Chapter \ref{sec:mdps} provides an overview of MDPs and outlines the standard formulations for discounted-cost and average-cost MDPs and stochastic shortest path total-cost MDP problems. Examples and basic theoretical results are included for the benefit of readers less familiar with MDPs.  Chapter \ref{sec:risk-measures} introduces all of the risk measures used in the book, namely \diff{exponential cost,} variance, CVaR, coherent risk measures, chance constraints, and CPT. Chapter \ref{sec:background}  provides an introduction to temporal difference learning and two gradient estimation techniques, namely simultaneous perturbation (stochastic approximation) and the likelihood ratio method.  Chapter \ref{sec:rpg-template} presents two templates for risk-sensitive policy gradient algorithms, one for the setting where the risk measure is the objective, and the other for the setting where the risk measure is featured in the constraint. This chapter also presents a convergence analysis of the template algorithms for both settings.
 Chapter \ref{sec:risk-algos} develops policy gradient algorithms for four special cases of risk-sensitive MDPs for the constrained optimization problem posed in \eqref{eq:gen-constrained-opt}, with variance, CVaR, and a chance constraint used as the risk measure constraint. Chapter \ref{sec:risk-algos-unconstrained} 
 \diff{develops policy gradient algorithms for three risk-sensitive MDP formulations in the unconstrained optimization setting of \eqref{eq:gen-unconstrained-opt} with risk explicitly as the objective: exponential cost, CPT, and coherent risk measures.} 
 Finally, Chapter \ref{sec:conclusions} provides concluding remarks and identifies some interesting future research directions.
 }

\clearpage
\diff{
\section*{A brief note on notation}
Throughout the book, 
the functions $J$, $G$, and $D$ may show one, two, or no arguments, depending on the context.
Specifically, the two possible arguments would be $\theta$, the policy parameter, as in  \eqref{eq:gen-constrained-opt} or \eqref{eq:gen-unconstrained-opt}, or a state of the MDP (e.g., $x_0, x, i, j$), 
as described in Chapter \ref{sec:mdps}.
This is particularly relevant to Chapters \ref{sec:rpg-template}, \ref{sec:risk-algos}, and \ref{sec:risk-algos-unconstrained}. 
The same ``convention'' is used for other analogous counterparts such as the variance and squared versions of these quantities. 
On the other hand, dependence on an entire MDP policy $\mu$ is represented as subscript, e.g., $J_{\mu}$.
Gradients represented by $\nabla$ are assumed to be with respect to $\theta$ unless otherwise indicated, 
e.g., $\nabla_{\lambda}$ denoting a gradient with respect to the Lagrange multiplier $\lambda$. 
Finally, all vectors will be assumed to be column vectors, and superscript ``$\mathsf{\scriptstyle T}$''  
will be used to denote the matrix/vector transpose operation. 
}

\section*{Bibliographic remarks}
MDPs have a long history dating back to the work of Richard E. Bellman. For a rigorous introduction, the reader is referred to the books by 
\textcite{puterman1994markov} and 
\textcite{Bertsekas2007}, and for reinforcement learning, the books by
\textcite{BertsekasT96,sutton2018reinforcement,csaba2011eorms}. 
Material in this book drawn from our own research includes
\textcite{l2013actor,prashanth2016mlj,prashanth2014cvar,prashanth2015cumulative,aditya2016weighted}.
Cumulative prospect theory (CPT) was introduced by 
\textcite{tversky1992advances}  
as a successor to prospect theory, which was one of the central contributions cited for Daniel Kahneman receiving the 
Nobel Memorial Prize in Economic Sciences in 2002. 

References for the various risk measures include the following:
mean-variance tradeoff \citep{markowitz1952portfolio}, exponential utility \citep{Arrow1971,Howard1972}, the percentile performance~\citep{Filar95PP}, the use of chance constraints
\citep{Prekopa2003}, stochastic dominance constraints \citep{dentcheva2003optimization}, value at risk (VaR), and conditional value-at-risk (CVaR) \citep{rockafellar2000optimization,Ruszczynski10RA,Shen13RS}. 
The concept of a coherent risk measure was introduced by \textcite{artzner1999coherent}, see also \textcite{Foellmer2004}, with the extension to multi-period settings treated in  \textcite{Riedel2004,Ruszczynski2006,Ruszczynski10RA,Cavus2014,Tallec2007,Choi2009}. 

The large body of literature utilizing the exponential utility formulation includes the classic formulation by 
\textcite{Howard1972}; related work includes \textcite{whittle1990risk,Browne95,fleming1995risk,HeMa96, Marcus1997,Fernandez-Gaucherand1997, Hernandez-Hernandez1999,
Coraluppi1999, Coraluppi1999a, Coraluppi2000,borkar2002risk,bauerle2014}. 
For a survey of risk-sensitive RL under the exponential utility formulation, the reader is referred to \textcite{borkar2010learning}.

\diff{Another approach to risk/uncertainty is the robust optimization approach. 
In the setting of Markov decision processes, \textcite{Iyengar2005} is an early seminal work in this area, where a robust optimal policy is defined relative to uncertainty in the underlying transition probabilities. 
We do not pursue the robust approach in this book. }

The existence of a Bellman equation for the variance of the return, where the underlying Bellman operator is not necessarily monotone, can be found in \textcite{Sobel82VD}.
The result that finding a globally mean-variance optimal policy in a discounted-cost MDP is NP-hard
can be found in \textcite{mannor2013algorithmic}.
The use of variance of a quantity that measures the deviation of the single-stage cost from the average cost can be found in \textcite{filar1989variance}.
The result that solving an average-cost MDP under this notion of variance is NP-hard is shown in \textcite{filar1989variance}.

Actor-critic methods investigated in RL are found in \textcite{Barto83NE,Sutton84TC}.
Temporal difference (TD) learning can be found in \textcite{sutton1988learning}. 
More recently, CVaR-constrained MDPs have been considered in \textcite{borkar2010risk,prashanth2014cvar,tamar2014optimizing}, though a variance-reduced CVaR estimation algorithm is still needed. 

The application of simultaneous perturbation stochastic approximation (SPSA) to policy gradient search for mean-variance optimization in discounted-cost MDPs is considered in \textcite{prashanth2016mlj} and for optimizing CPT-value in \textcite{prashanth2015cumulative}.

Prospect theory (PT) was introduced in \textcite{kahneman1979prospect}, 
and cumulative prospect theory (CPT) in \textcite{tversky1992advances}, 
with experiments on humans reported in \textcite{starmer2000developments,tversky1992advances}.  
More work adopting this approach includes \textcite{lin2013stochastic,Lin2013a,Lin2013b,Lin2018}; see also \textcite{Cavus2014}.

Variance as a risk measure in a discounted-cost and average-cost MDP, respectively, are based on \textcite{l2013actor,prashanth2016mlj}. 
CVaR as a risk measure is based on \textcite{prashanth2014cvar}. 
CPT as the risk measure is based on \textcite{prashanth2015cumulative,prashanth2018cpt}. 

A sampling but nowhere near exhaustive list of other risk-sensitive RL work includes the following. In \textcite{tamar2012policy}, variance as risk is considered in a stochastic shortest path context, and a policy gradient algorithm using the likelihood ratio method is provided. In \textcite{Mihatsch02RS}, a modified temporal differences algorithm is proposed and connected to the exponential utility approach. A general policy gradient algorithm that handles a class of risk measures that includes CVaR is presented in \textcite{TamarCGM15}.
An early work that considers a constrained MDP setting similar to that in \eqref{eq:gen-constrained-opt} is \textcite{borkar2005actor}, where the objective is average cost and the constraint is also an average-cost function different from the objective function. A modification of this formulation to a discounted-cost MDP, incorporating function approximation, was treated in \textcite{bhatnagar2010actor}. 
CVaR optimization in a constrained MDP setup was also explored in \textcite{borkar2010risk}, but the algorithm proposed there requires that the single-stage cost be separable. Optimization of risk measures that include CVaR in an unconstrained MDP setting using RL algorithms with function approximation  can be found in \textcite{jiang2017risk}.  

%
%
\chapter{Markov Decision Processes}
\label{sec:mdps}
\vspace*{-70pt}
A Markov decision process (MDP) is a discrete-time stochastic process that transitions from one state to the next in a probabilistic fashion after the execution of an action, on the way possibly accumulating costs (or rewards). 
The objective of an MDP is to minimize some cost function (or maximize some reward function) over the horizon of interest, which may be finite (possibly random) or infinite. 
In this chapter, we provide a basic overview of MDP theory for several of the most commonly used settings.
A reader familiar with the theory of MDPs can skip this chapter and move to the description of risk measures and risk-sensitive RL algorithms in the subsequent chapters.

We consider an MDP $\{x_t, t=0, 1, \ldots \}$ with state space $\X$ and action space $\A$ (both assumed to be finite), and starting in state $x_0$. Let $P(\cdot|x,a)$ denote the state transition probability distribution from state $x$ under action $a$, $k(x,a)$ denote the single-stage cost incurred in state $x$ under action $a$,
and $\A(x) \subset \A$ denote the set of feasible actions in state $x$ (all assumed to be stationary).
The sequence of actions $\{a_t, t=0, 1, \ldots \}$ follows a policy $\mu$ (also assumed to be stationary). 
For simplicity, we consider nonrandomized policies, 
i.e., $\mu: \X \longrightarrow \A$.
We extend to randomized policies at the end of the chapter when we introduce the policy parameterization to be considered in Chapter \ref{sec:rpg-template}. 
We introduce MDPs under three different (risk-neutral) objectives/settings: infinite-horizon discounted cost, stochastic shortest path total cost, and infinite-horizon average cost, described in Sections \ref{sec:mdp-discounted}, \ref{sec:mdp-ssp}, and \ref{sec:mdp-average}, respectively.

\vspace{0.5ex}


\section{Discounted-cost MDP}
\label{sec:mdp-discounted}
\vspace*{-4pt}
The infinite-horizon cumulative discounted cost for an MDP trajectory (or sample path) under policy $\mu$, starting in state $x_0$, is given by
\vspace*{-2pt}
\begin{align}
D_\mu(x_0)=\sum_{m=0}^\infty\gamma^m k(x_m,a_m),\label{eq:discounted-cost}
\vspace*{-5pt}
\end{align} 
where $\gamma \in [0,1)$ is the discount factor and $a_m = \mu(x_m)$.

In a risk-neutral setting, the performance measure (or value function) associated with a policy $\mu$ is the expected total discounted cost denoted by 
$$
J_\mu(x_0) \triangleq \E\left[D_\mu\left(x_0\right)\right],
$$
and the objective is to find the optimal cost or value function
\begin{align} 
J^* \triangleq \min_{\mu \in \Xi} \left\{ J_\mu\right\} 
\label{oe} 
\vspace*{-2pt}
\end{align}
and/or the associated optimal policy $\mu^* \triangleq \arg\min_{\mu \in \Xi} \left\{ J_\mu \right\}$, 
where $\Xi$ denotes the set of admissible policies.
A policy $\mu$ is {\it admissible} if it considers only feasible actions in any given state, i.e.,
$\mu(x) \in \A(x)$.

For an $|\X|$-dimensional vector $J \triangleq [J(x)]_{x\in\X}$, define the (Bellman optimality) operator $T$ as follows:
\begin{align}
	(TJ)(x) \triangleq \min_{a\in \A(x)} \bigg\{ k(x,a) + \gamma \sum_{y\in\X} P(y|x,a)J(y) \bigg\},\ \forall x\in\X,   \label{T_eqn_1}
\end{align}
where $\A(x)$ denotes the set of feasible actions in state $x$.
Next, we define another (Bellman policy) operator $T_{\mu}$ specific to a given policy ${\mu}$:
\begin{align}
	(T_{\mu} J)(x) \triangleq k(x,\mu(x)) + \gamma \sum_{y\in\X} P(y|x,\mu(x)) J(y),\ \forall x\in\X.
	\label{T_eqn_mu}
\end{align}

\vspace*{-4pt}
The first important result is the following, which forms the basis for the value-iteration algorithm for policy evaluation. 
\vspace*{-2pt}
\boxed{\begin{proposition} \label{prop-SSPfixed}
\vspace*{-8pt}
$ T_{\mu}^m J \longrightarrow J_\mu \mbox{~as~} m \rightarrow \infty,~\forall J \triangleq [J(x)]_{x\in\X}, $ \\
			where $J_{\mu}\triangleq [J_\mu(x)]_{x\in\X}$ is the unique solution of 
\begin{equation}
			J_{\mu} = T_{\mu} J_{\mu}. \label{eq:J-Bellman}
\end{equation}
\end{proposition}}
Value iteration relies on the result in Proposition \ref{prop-SSPfixed}, as it repeatedly applies the $T_{\mu}$ operator for the expected cost $J_{\mu}$ defined by \eqref{T_eqn_mu}, starting with an arbitrary $J_0$, i.e., the following update rule for computing $J_{\mu}$: 
\begin{align}
	J_{k+1}(x) = k(x,\mu(x)) +  \gamma\sum_{y\in\X} P(y|x,\mu(x)) J_k(y), \ \forall x\in \X.
\end{align}

Important properties of the Bellman optimality operator $T$ and its relationship to the optimal value function $J^*$ are summarized here.
\boxed{\begin{proposition} \label{disc:Bellman}
\vspace*{-8pt}
~\\
(i) $J^* \triangleq [J^*(x)]_{x\in\X}$ is the unique solution to $J^* = T J^*.$ \\
(ii) For any $J\triangleq [J(x)]_{x\in\X}$,    
$ T^m J \longrightarrow J^* \mbox{~as~} m \rightarrow \infty, $ \\
(iii) A policy ${\mu}$ is optimal if and only if
$ T_{\mu} J^* = T J^*.$
\end{proposition}}
The fixed-point equation in (i) is referred to as the Bellman (optimality) equation, and  
the corresponding value iteration update for computing the optimal value $J^*$ is the following:
\begin{align}
	J^*_{k+1}(x) = \min_{a\in \A(x)} \left(k(x,a) +  \gamma\sum_{y\in\X} P(y|x,a) J^*_k(y)\right), \ \forall x\in \X.
\end{align}

We now present a simple illustrative example where the optimal policy is found using the Bellman equation.
\begin{example}\textbf{\textit{(machine replacement problem)}}\\
A machine can be in any of $n$ states, denoted $1,2,\ldots,n$, where state $1$ corresponds to a machine in perfect condition, and higher-valued states indicate deteriorating conditions (e.g., age of the machine), for which it is more costly to operate the machine. 
The goal is to decide when to replace the machine to minimize the long-run discounted operating cost. 

Let $k(i)$ denote the cost of operating the machine in state $i$ in any time period (in this case with no direct dependence on the action), with 
\[ k(1) \le k(2) \ldots \le k(n). \]
The machine deteriorates stochastically according to transition matrix ${\bf P} = [p_{ij}]$, where $p_{ij}$ is the probability that the machine goes from state $i$ to $j$. Assume the machine can never improve its state without intervention, which implies $p_{ij} =0$ for $j<i$. 
The state of the machine at the beginning of each time period is known, and there are just two possible actions: (i) do nothing, or 
(ii) replace the machine with a new machine (state $1$) at cost $R$.
The problem is to choose the actions to minimize the infinite-horizon discounted cost. Intuitively, it makes little sense to replace the machine when it is almost new, i.e., the state is close to $1$, and it turns out that a policy that uses a threshold to determine whether to replace or not is optimal.  

For a machine in state $i$, if the action chosen is to replace the machine, then the machine will go to state 1 for the next period, incurring immediate cost $R+k(1)$ and a future discounted (optimal) expected cost from state 1; otherwise, it deteriorates to state $j \geq i$ according to the state transition probabilities $p_{ij}$, incurring immediate cost $k(i)$ and a corresponding future discounted (optimal) expected cost. 
Putting these together for the RHS of \eqref{T_eqn_1}, the Bellman equation is obtained by applying Proposition \ref{disc:Bellman}(i):
$$
J^{*}(i) = \min\bigg\{R + k(1) + \gamma J^{*}(1), k(i) + \gamma \sum_{j=i}^n p_{ij}J^{*}(j)\bigg\}.
$$
Letting ${\bf P}_{i,:}$ denote the $i$th row of ${\bf P}$, so $\sum_{j=1}^n p_{ij}J^{*}(j) = {\bf P}_{i,:} \cdot J^{*}$,
the optimal action is to replace a machine in state $i$ if 
$$
R + k(1) + \gamma J^{*}(1) \leq k(i) + \gamma {\bf P}_{i,:} \cdot J^{*},  
$$
and do nothing otherwise.
\end{example}


To establish that the optimal policy is threshold-based requires the following additional assumption (besides the previous assumption that the machine cannot get better on its own):
\vspace*{-3pt}
$$
p_{ij} \leq p_{(i+1)j}, ~i < j,
\vspace*{-4pt}
$$ 
which implies that the machine is more likely to jump to a worse state from a state that is closer to it than from a state that is farther away.
For $J$ satisfying $J(1) \leq J(2) \leq \ldots \leq J(n)$, and $\forall i \in \{1,2,\ldots,n-1\}$, 
\vspace*{-2pt}
$$
{\bf P}_{i,:} \cdot J \leq {\bf P}_{i+1,:} \cdot J
\implies 
k(i) + \gamma {\bf P}_{i,:} \cdot J \leq k(i+1) + \gamma {\bf P}_{i+1,:} \cdot J,
\vspace*{-2pt}
$$
where the RHS inequality used the assumption that $k(i)$ is nondecreasing. 
Thus, $(TJ)(i)$ is nondecreasing in $i$ if $J$ is nondecreasing. Using this observation together with the fact that $T$ is a monotone operator, $(T^m J)(i)$ is nondecreasing in $i$ for any $ m \ge 1$, which implies that $J^{*}(i) = \lim_{m \to \infty} (T^m J)(i)$  is nondecreasing in $i$. 

The foregoing implies that the function $h(i)= k(i) + \gamma {\bf P}_{i,:} \cdot J^{*}$ is nondecreasing in $i$. 
Let $i^*$ denote the smallest state satisfying
$R + k(1) + \gamma J^{*}(1)  \leq k(i) + \gamma {\bf P}_{i,:} \cdot J^{*}$.
As illustrated in Figure \ref{tikz}, it is easy to infer that the following policy is optimal for the machine replacement problem:
replace if $i \geq i^{*}$; else, do nothing. 
%

\begin{figure}[htb]
	\centering
	\vspace*{-1pt}
	\begin{tikzpicture}[scale=2,>=latex,x=1.5cm]
		\begin{scope}[domain=0:1]
			\draw[->] (-0.2,0) -- (1.8,0) node[above left]{\footnotesize state $(i)$};
			\draw[->] (0,-0.2) -- (0,1.8) node[below right]{\footnotesize $h(i) = k(i) + \gamma \sum_{j=i}^n p_{ij}J^{*}(j)$};
			\draw[-,blue!30!green] (0,0.2)to[out=0,in=225](0.9,0.9);
			\draw[-,blue!30!green] (0.9,0.9)to[out=45,in=190](1.5,1.5);
			\node at (-0.45,0.9) {\footnotesize $R + k(1) + \gamma J^{*}(1)$};
			\draw[-] (0,0.9) to (0.9,0.9);
			
			\node at (0.9,-0.1) {\footnotesize $i^*$};
			\draw[-] (0.9,-0.05) to (0.9,0.9);
			\node at (1.5,-0.1) {\footnotesize $n$};
			\draw[-] (1.5,-0.05) to (1.5,0.05);
			
			\draw[<->] (0,-0.3)--(0.5,-0.3)node [below]{\scriptsize do nothing} to (0.9,-0.3) ;
			\draw[<->] (0.9,-0.3)--(1.3,-0.3)node [below]{\scriptsize replace~~~} to (1.5,-0.3)  ;
			\end{scope}
	\end{tikzpicture}
	\vspace*{-4pt}
	\caption{The optimal threshold-based policy for the machine replacement problem (For ease of illustration, $h(i)$ is plotted as a continuous function.).} 
	\label{tikz}
\end{figure}

\section{Stochastic shortest path MDP}
\label{sec:mdp-ssp}
In a stochastic shortest path (SSP) problem, the horizon is finite but unknown (random), as the MDP terminates when it enters a predetermined state (or possibly set of states in a generalized setting). 
The basic version of a shortest path problem is to traverse from a source (origin) node to a sink (destination) node, hence the term stochastic shortest path problem.

An \textit{episode} is a sample path using a policy $\mu$ that starts in state $x_0\in \X$, visits $\{x_1,\ldots, x_{\tau-1}\}$ before ending in the (cost-free) {\it absorbing} state $0\in \X$, where $\tau$ is the first passage time to state $0\in \X$.
	Let 
	$ D_\mu(x_0) = \sum\limits_{m=0}^{\tau-1} k(x_m,a_m)$ denote the total cost from an episode,
	with the actions $\{a_m\}$ chosen according to policy $\mu$, i.e., $a_m=\mu(x_m)$.
In a risk-neutral setting, the performance measure (or value function) associated with a policy $\mu$ is 
$$
J_\mu(x_0) \triangleq \E\left[D_\mu\left(x_0\right)\right].  
$$
We consider policies that ensure that the absorbing state $0$ is recurrent and the remaining states transient in the underlying Markov chain. Such policies are referred to as ``proper'', and we formalize this notion in the following definition:
\begin{definition} \label{policy-proper}
	A stationary policy $\mu$ is {\it proper} if $\forall x \in \X$, there exists $M>0$ s.t. 
	\[ \rho_\mu = \max_{x\in \X} \Prob{x_M\ne 0 \mid x_0=x,\mu} < 1.\]
\end{definition}

\vspace*{10pt}
The condition in Definition \ref{policy-proper} ensures that there exists a path of positive probability from any state to the absorbing state $0$. 
To understand the need for the notion of a proper policy, consider a three-state deterministic shortest path example shown in Figure \ref{fig:SSPex}, where the costs (rather than transition probabilities) are shown on the arcs.
%
\begin{figure}[h!] 
\begin{center}
	\scalebox{0.59}{
		\begin{tikzpicture}[->,>=stealth',shorten >=1pt,auto,node distance=4cm,
			semithick]
			\tikzstyle{every state}=[fill=red!60,draw=none,text=white]
			
			\node[state] (A)   {$1$};
			\node[state]         (B) [below of=A] {$2$};
			\node[state]         (D) [above right =3cm of B] {$0$};
			
			\path (A) edge [bend left]              node {{\LARGE $\epsilon$}} (B)       
			(A) edge node {$1$} (D)
			(B) edge node[below]  {$1$} (D)
			(B) edge [bend left] node {{\LARGE $\epsilon$}} (A);
	\end{tikzpicture}}
\end{center}
\caption{Three-state deterministic shortest path example (costs on the arcs).}
\label{fig:SSPex}
\end{figure}

\noindent
It is obvious that the least-cost path from node $1$ (or $2$) to $0$ follows the edge connecting it to $0$ in one step, with a total cost of $1$. However, one improper policy is a policy that leads to a path that loops between $1$ and $2$. Since the edge costs are positive, it is clear that such an improper policy would incur infinite cost. On the other hand, having zero cost edges between $1$ and $2$ (i.e., letting $\epsilon \rightarrow 0$) would mean the improper policy would incur zero cost in the long run -- one type of pathological scenario that we would like to avoid, motivating the following assumptions for the analysis of SSPs:
\begin{assumption}
	\label{ass:ssp-proper}
There exists at least one proper policy.
\end{assumption}
\begin{assumption}
	\label{ass:ssp-improper}
For every improper policy $\mu$, the associated cost $J_\mu(x_0)$ is infinite for at least one state $x_0$.
\end{assumption}

The risk-neutral objective in an SSP context is to minimize the expected total cost, i.e.,
\begin{align} 
\min_{\mu \in \Xi} \left\{ J_\mu(x_0)\right\},
\label{eq:ssp-cost} 
\end{align}
where $\Xi$ denotes the set of admissible policies that satisfy \ref{ass:ssp-proper} and \ref{ass:ssp-improper}. 

As in the discounted-cost setting, 
for an $|\X|$-dimensional vector $J \triangleq [J(x)]_{x\in\X}$, define the (Bellman optimality) operator $T$:
\begin{align}
	(TJ)(x) \triangleq \min_{a\in \A(x)} \bigg\{ k(x,a) + \sum_{y\in\X} P(y|x,a)J(y) \bigg\},\ \forall x\in\X,   \label{T_eqn}
\end{align}
and also the operator specific to a given policy ${\mu}$, $T_{\mu}$, defined as
\begin{align}
	(T_{\mu} J)(x) \triangleq k(x,\mu(x)) + \sum_{y\in\X} P(y|x,\mu(x)) J(y),\ \forall x\in\X.
	\label{T_eqn_muSSP}
\end{align}

Now, we state a few important properties of the $T_{\mu}$ operator in the proposition below.
\boxed{\begin{proposition}
		\label{prop:ssp-tpi}
		Assume \ref{ass:ssp-proper} and \ref{ass:ssp-improper}. Then, for any proper policy ${\mu}$ and any $J \triangleq [J(x)]_{x\in\X}$, 
			\begin{align}
				\lim_{m\rightarrow \infty} (T_{\mu}^m J)(x) = J_\mu(x), \ \forall x\in\X,\label{eq:tpilim}  
			\end{align}
			where $J_{\mu}\triangleq [J_\mu(x)]_{x\in\X}$ is the unique solution of the fixed-point equation 
			\[J_{\mu} = T_{\mu} J_{\mu}. \]
\end{proposition}}
As in the discounted case, the convergence result given by \eqref{eq:tpilim} can be used to establish convergence of the value iteration algorithm, which repeatedly applies the $T_{\mu}$ operator given by \eqref{T_eqn_muSSP}, starting with an arbitrary $J_0$.

We now turn our attention to the Bellman optimality operator $T$, and state a few important properties concerning this operator.
First we define the optimal cost (value) function:
$$
J^*(x_0) \triangleq \min_{\mu \in \Xi} J_{\mu}(x_0), \forall x_0 \in \X. 
$$
\boxed{\begin{proposition}
		Assume \ref{ass:ssp-proper} and \ref{ass:ssp-improper}. Then, 
		\begin{enumerate}
			\item $J^* \triangleq [J^*(x)]_{x\in\X}$ is the unique solution to the fixed-point equation
			\begin{align} 
				J^* = T J^*.\label{eq:be-ssp} 
				\end{align}
			\item For any $J\triangleq [J(x)]_{x\in\X}$,    
			\begin{align}
				\lim_{m\rightarrow \infty} (T^m J)(x) = J^*(x), \ \forall x\in\X.\label{eq:tlim}  
			\end{align}
			\item A stationary policy ${\mu}$ is optimal if and only if
			\[ T_{\mu} J^* = T J^*.\]
		\end{enumerate}
\end{proposition}}
We illustrate the usage of Bellman equation \eqref{eq:be-ssp} for finding the optimal policy in the example below.
\begin{example}\textbf{\textit{(the spider and the fly)}}\\
	A spider hunts a fly on the one-dimensional line of integers $..., -1, 0, +1, ...$ In each period/stage, the fly jumps forward or backward $1$ unit with probability $p$ and remains in the same position with probability $1 - 2p$.
	The spider jumps (1 unit) towards the fly if the distance between them is greater than $1$ unit. If the distance between them is exactly $1$ unit, the spider can choose to stay in its position hoping the fly will come to it or go $1$ unit forward.
	The game (and the fly) ends when the spider is in the same position as the fly. The goal is to decide the actions of the spider to minimize the expected time to catch the fly. 

	This problem can be formulated as an SSP, with the state as the distance between the spider and the fly. The terminal state is state $0$, which is reached when the spider and fly are in the same position.
	Note that given an initial distance between the spider and the fly, the subsequent distance between them can never be greater than this distance, so that the number of states is finite. 
	Specifically, assuming that the spider and fly are initially at a separation of $n$, the state space is 
$		\X = \{ 0,1 \dots n \}$.
	The  transition probabilities are obtained as follows:
	When the state is $2$ or more, the spider has  to jump towards the fly, leading to
	\begin{align*}
		&p_{i,i-2} = p, \ 
		p_{i,i-1} = 1 - 2p, \textrm{ and }
		p_{i,i} = p, \textrm{ for } i\ge 2, 
	\end{align*}
	where $p_{i,j}$ denotes the probability that the system transitions from state $i$ to $j$, when the spider jumps, for $i\ge 2$. Since the optimal action for the spider is to jump in this situation, we drop the dependence on the spider's action in the transition probabilities for $i\ge 2$.

	When the state is $1$, the spider has two possible actions. 
	Denoting $M$ and  $\overline M$ as the actions ``move''  and ``don't move'', respectively, the transition probabilities as a function of the spider's action are given by
	\begin{align*}
		&p_{1,1}(M) = 2p, \quad
		p_{1,0}(M) = 1 - 2p, \textrm{and }\\
		&p_{1,2}(\overline{M}) = p, \
		p_{1,0}(\overline{M}) = p, \textrm{ and }
		p_{1,1}(\overline{M}) = 1 -2p.
	\end{align*}
	To find the minimum expected time to catch the fly, we set all the single-stage costs to $1$.
	
	We now derive the Bellman equation \eqref{eq:be-ssp} for this problem by finding the RHS of \eqref{T_eqn} for each state.
	As there is only one action to take for states $i>1$, the Bellman equation for $i>1$ is
	\begin{align*}
		J^*(i) 
		& = 1 + p_{i,i-2}J^*(i-2) + p_{i,i-1} J^*(i-1) + p_{i,i} J^*(i) \\
		& = 1 + p J^*(i-2) + (1-2p)J^*(i-1)+ p J^*(i).
	\end{align*}
	For state $1$, we have to take the minimum over two actions, leading to
	\begin{align*}
		J^*(1) 
		&= 1+ \min \left(2pJ^*(1),\: pJ^*(2)+(1-2p)J^*(1) \right).
	\end{align*}
	Therefore, we have 
	\begin{align*}
		J^*(i) = 
		\begin{cases}
			1 + pJ^*(i-2) + (1-2p)J^*(i-1) + pJ^*(i) & ~i \geq 2, \\
			1 + \min(2pJ^*(1), \:pJ^*(2) + (1-2p)J^*(1)) & ~i = 1, \\
			0 \textrm{~~(cost-free absorbing state)} & ~i = 0. \label{spidfly}
		\end{cases}
	\end{align*}
Straightforward calculations involving $J^*(2)$ and $J^*(1)$ lead to the spider's optimal action in state $1$ being to move if the probability of the fly jumping is under 1/3, leading to the spider's overall optimal policy (for $p=1/3$, either action is optimal in state $i=1$):
	\begin{equation}
		\label{eq:spflyopt}
		\mu(i) = \begin{cases}
			M &\textrm{$i = 1$ and $p < \frac{1}{3}$}, \\
			\overline{M} &\textrm{$i = 1$ and $p > \frac{1}{3}$}, \\
			M &\textrm{$i \geq 2$}.
		\end{cases}
	\end{equation}
\end{example}

\vspace*{-4pt}
\section{Average-cost MDP}
\label{sec:mdp-average}
\vspace*{-6pt}
The average cost under policy $\mu$ is defined as
\begin{align}
J_\mu(x) \; \triangleq \; \lim_{T\rightarrow\infty}\frac{1}{T}\E\left[\sum_{m=0}^{T-1} k(x_m,a_m) \middle\vert x_0=x\right],
\label{eq:avg-cost}
\end{align}
where $a_m=\mu(x_m)$.

Under the following unichain assumption, the infinite-horizon average cost in \eqref{eq:avg-cost} is identical for all initial states $x_0 \in \X$:
\vspace*{-2pt}
\begin{assumption}
	\label{ass:avgcost-same}
The Markov chain generated by any policy $\mu$ is irreducible and positive recurrent.
\end{assumption}
Under \ref{ass:avgcost-same}, we can drop the dependence in \eqref{eq:avg-cost} on the initial state and simply use $J_\mu$ to denote a \textbf{\textit{scalar}} rather than a vector as in the two previous settings.
 
The goal in the standard (risk-neutral) average-cost formulation is 
\[\min_{\mu\in\Xi}J_\mu,\]
where, as before, $\Xi$ denotes the set of admissible policies.

For any policy $\mu$, we associate an expected differential value function defined as follows: 
\vspace*{-2pt}
\begin{align}
\label{eq:DifV-Q}
Q_\mu(x,a)&\triangleq \sum_{m=0}^\infty\E\big[k(x_m,a_m)-J_\mu \mid x_0=x,a_0=a,\mu\big], \\
V_\mu(x) &\triangleq Q_\mu(x,\mu(x)),
\label{eq:DifV}
\end{align}
where $V_\mu(x)$ is the expected sum of the differences between the single-stage cost and the average cost $J_\mu$ under the policy $\mu$ with initial state $x \in \X$, and $Q_\mu(x,a)$ has an analogous connotation. 
$V_\mu$ and $Q_\mu$ are referred to as the differential value and Q-value functions, respectively.

\boxed{
\vspace*{-6pt}
\begin{proposition}  \label{prop:Poission}
Assume \ref{ass:avgcost-same}.	The differential value and Q-value functions satisfy the following Poisson equations:
\begin{align}
J_\mu+V_\mu(x) &=  k(x,\mu(x)) +\sum_{y \in \X}P(y|x,\mu(x))V_\mu(y),~~~~~~~  \label{eq:DifV-Q-Poisson}
\\
J_\mu+Q_\mu(x,a)&=k(x,a)+\sum_{y \in \X}P(y|x,a)V_\mu(y).
\end{align}
\end{proposition}}

A well-known algorithm for computing the differential value function of a given policy using the Poisson equation, is `relative value iteration', which employs the following update rule:
\begin{align}
	V_{m+1}(x) = k(x,\mu(x)) + \hspace{-6pt}
	\sum_{y \in \X} \hspace{-4pt} P(y|x,\mu(x)) V_m(y) - V_m(x_f),~\forall x \in \X,
\end{align}
where $x_f$ is a fixed state. $V_m$ converges asymptotically to the differential value function $V_\mu$, although the proof is more delicate than in the discounted-cost or SSP settings, where the $T_\mu$ operator is a contraction, which is not the case for the average-cost setting. 
In addition to computing the differential value, relative value iteration can be used to obtain the average cost $J_\mu$, since the term $V_m(x_f)$ converges to $J_\mu$.

We now turn our attention to the optimal average cost, and its associated Bellman (optimality) equation.
\boxed{
\vspace*{-6pt}
\begin{proposition}
Assume \ref{ass:avgcost-same}. Let $J^* = \min_{\mu \in \Xi} J_\mu$ denote the optimal average cost. Let $\mu^*$ denote the optimal policy, and let $V^*=V_{\mu^*}$ denote the differential value function associated with the optimal policy $\mu^*$. Then, $(J^*,V^*(x)),x\in \X$, satisfy the following Bellman equation:
\begin{align}
	J^* + V^*(x) = \min_{a \in \A(x)} \Big\{ k(x,a) +  
	\sum_{y\in\X}P(y|x,a) V^*(y) \Big\}, \forall x \in \X.
	\label{bf-a}
\end{align}
\end{proposition}
}
The `relative value iteration' for computing the optimal average cost employs the following update rule:
\begin{align}
	V^*_{m+1}(x) = \min_{a\in \A(x)} \bigg(k(x,a) + \hspace{-8pt}
	\sum_{y \in \X} \hspace{-4pt} P(y|x,a) V^*_m(y)\bigg) - V^*_m(x_f),~\forall x \in \X,
\end{align}
where $x_f$ is a fixed state. Again, $V_m$ converges asymptotically to the optimal differential value function $V^*$, and $V_m(x_f)$  converges to $J^*$.

\smallskip
\begin{example}\textbf{\textit{(machine replacement problem revisited)}}\\
	Consider the machine replacement example again, with the average-cost objective in place of discounted cost. Using the same notation, the Bellman equation for the average-cost problem is given by
	$$
		J^* + V^*(i)  = \min \bigg\{ R+k(1) + V^*(1), k(i) +
		\sum_{j=1}^n p_{ij} V^*(j)\bigg\}, i=1,\ldots,n,
	$$
where the action that minimizes the RHS of the Bellman equation is the optimal action in state $i$.

What is needed to ensure that the Bellman equation is solvable for the average-cost version of this problem? First, observe that not all polices necessarily satisfy the unichain assumption now, i.e., there exists policies that do not satisfy \ref{ass:avgcost-same}. To see this, consider a policy that replaces the machine in all states $i<n$, and in state $n$ does nothing. Such a policy clearly results in two disjoint recurrent classes, and hence the underlying Markov chain is not unichain. A more general assumption that ensures solvability of the Bellman equation is the following: 
\begin{assumption}
\label{ass:avgcost-prime}
The state space can be partitioned into two disjoint classes $\X_1$ and $\X_2$ such that (i)  all states in $\X_1$ are transient in the Markov chain generated by any policy $\mu$; and (ii) there exists a policy, say $\tilde\mu$, and a positive integer $M$ such that $\Prob{x_M=x'  | x_0=x,\tilde \mu} >0,$ $~\forall x,x' \in \X_2$. 
\end{assumption}

This assumption is an intuitive extension of the unichain assumption, allowing the presence of transient states, whereby the MDP eventually acts as if it were unichain once it leaves the set of transient states.  

Under the assumption above, the average cost $J$ is identical for all initial states, as in the unichain setting. Moreover, it can be easily verified that the machine replacement problem satisfies \ref{ass:avgcost-prime}, implying the Bellman equation is solvable. Furthermore, the arguments used for discounted-cost MDPs to show that the form of the optimal policy is threshold-based can be extended to the average-cost setting, as well, and we omit the details.
\end{example}

\section{Randomized policies and policy parameterization}

Nonrandomized policies $\mu$ specify a single action for a given state, i.e., $\mu(x),~x \in \X$, whereas 
a randomized policy $\mu$ specifies a {\it probability distribution} over the feasible action space, i.e., $\mu(\cdot | x)$ will be used to denote a distribution over $\A(x)$.
For the discrete state/action space setting, $\mu(a | x)$ is simply the probability of taking action $a$ in state $x$, and a randomized policy $\mu$ is admissible if the policy puts nonzero probability on only feasible actions in any given state, i.e.,
$\mu(a|x)>0$ implies $a \in \A(x)~\forall x,a$.
Various definitions would be adjusted accordingly, by summing over the action space. 
For instance, for the discounted-cost setting, the $T_{\mu}$ operator given by \eqref{T_eqn_mu} becomes
$$
(T_{\mu} J)(x) \triangleq \sum_{a\in \A(x)} \mu(a| x) \Big[ k(x,a) + \gamma \sum_{y\in\X} P(y|x,a) J(y) \Big],\ \forall x\in\X, 
$$
and for the average-cost setting, Equations \eqref{eq:DifV} and \eqref{eq:DifV-Q-Poisson} become
$$
V_\mu(x) \triangleq \sum_{a \in \A(x)} \mu(a | x) Q_\mu(x,a)  
$$
and
$$
J_\mu+V_\mu(x) = \sum_{a \in \A(x)} \mu(a | x)  \Big[ k(x,a) +\sum_{y \in \X}P(y|x,a)V_\mu(y) \Big],  
$$
respectively. 

Our focus in policy gradient risk-sensitive RL will be on parameterized randomized policies $\mu^\theta$, 
where 
$\theta$ denotes the policy parameter, which appears in the distribution. 
For example, in the machine replacement problem, a randomized policy would specify the probability of replacing the machine (or doing nothing), 
and in `the spider and the fly' example, a randomized policy would specify the probability of moving (or not moving) in each state, e.g., specified as follows: 
$$
\mu^\theta(M|i) = \theta^i,~~ \mu^\theta(\overline{M}|i) = 1-\theta^i,~~ 
\theta = [\theta^0 ~\theta^1 ~\cdots~ \theta^n],
$$
where $\theta^i$ denotes the probability of the spider moving when in state $i$.
Policy gradient methods require estimators for $\nabla J(\theta)$,
the gradient of the objective function (i.e., total cost or average cost) with respect to the policy parameters.

\section{Bibliographic remarks}
MDPs have a long history dating back to the seminal work of Richard E. Bellman in the 1950s \citep{Bellman1957}. For a more complete treatment, the reader may refer to the books by 
\textcite{Derman1970,Ross1983,puterman1994markov,Bertsekas2007,bertsekas2012dynamic,sutton2018reinforcement}, where the last text focuses on reinforcement learning.
We assume finite state and action spaces throughout, 
as the infinite, especially uncountable, space setting involves deeper mathematics beyond the scope of this work; 
a comprehensive review in the average-cost setting is \textcite{Arapostathis1993}.
The presentation of the discounted-cost MDP in Section \ref{sec:mdp-discounted} is based on material from Chapter 1 of \textcite{bertsekas2012dynamic}. 
In Section \ref{sec:mdp-ssp}, the notion of a proper policy is based on the treatment of SSPs from \textcite[Chapter 2]{bertsekas2012dynamic}. The latter chapter is also the source for the presentation of Bellman operator and `the spider and the fly' example in Section \ref{sec:mdp-ssp}.
For the Poisson and Bellman equations in the average-cost MDP from Section \ref{sec:mdp-average}, the reader can refer to either \textcite{puterman1994markov} or 
\textcite[Chapter 4]{bertsekas2012dynamic}. The unichain Assumption \ref{ass:avgcost-same}, as well as its relaxation in Assumption \ref{ass:avgcost-prime}, is also based on material from Chapter 4 of \textcite{bertsekas2012dynamic}.
For simulation-based approaches, see \textcite{chang2007simulation, chang2013simulation, Gosavi2003}.

%
\chapter{Risk Measures}
\label{sec:risk-measures}
\diff{
In this chapter, we introduce risk measures that can be incorporated into risk-sensitive RL, whether explicitly in the objective function or implicitly as a constraint. The first risk measure involves a modified functional form of exponential utility serving as the objective function in an average-cost MDP, which we will refer to throughout as exponential cost. This measure has been analyzed in depth in the risk-sensitive stochastic control community, but there remains a dearth of computationally practical policy gradient algorithms for high-dimensional state-space RL settings. The second and third risk measures are based on two different interpretations of variance in discounted-cost and average-cost MDPs, described in Sections \ref{sec:risk-discounted} and \ref{sec:risk-average}, respectively. The fourth risk measure, described in Section \ref{sec:risk-cvar}, is conditional Value-at-Risk (CVaR), widely used in finance. The fifth risk measure, detailed in Section \ref{sec:risk-chance}, are chance constraints commonly used in stochastic optimization problem formulations, as briefly mentioned in the commuting example of Chapter \ref{sec:intro}. The sixth risk measure, described in Section \ref{sec:risk-coherent}, is the class of coherent risk measures, which includes CVaR. The final risk measure, described in Section \ref{sec:risk-cpt}, is based on cumulative prospect theory (CPT), which has been found to model human decision making well. }

\diff{
\section{Exponential cost in average-cost MDPs}
\label{sec:exp-cost}

As mentioned in Chapter \ref{sec:intro}, an exponential utility function is one way commonly used to capture risk preferences.
For average-cost MDPs, the corresponding risk measure takes the following form (referred to henceforth as \textbf{\textit{exponential cost}}):
\boxed{
	\begin{align}
		G_\mu
	& \triangleq \limsup_{T\rightarrow\infty}\frac{1}{T}\frac{1}{\beta}\log \E\left[\exp\left(\beta\sum_{n=0}^{T-1}k(x_n,a_n)\right)\right],
	\label{eq:expcost-variant}
\end{align}
}
where $\beta$ is a parameter that controls risk sensitivity. 
Assuming that the single-stage costs are positive, i.e., $k(\cdot,\cdot)>0$, 
$\beta>0$ corresponds to the risk-averse setting and $\beta<0$ to the risk-seeking setting, 
and in the limit $\beta \rightarrow 0$, the exponential cost approaches the classic (risk-neutral) average cost.
If the Markov chain generated by the policy $\mu$ is irreducible and positive recurrent, 
i.e., satisfies \ref{ass:avgcost-same}, 
then the $\limsup$ in the definition (\ref{eq:expcost-variant}) can be replaced by an ordinary limit. 
}

\section{Variance in discounted-cost MDPs}
\label{sec:risk-discounted}
For discounted MDPs with starting state $x_0$, we consider a {\em variability} risk measure defined as follows:
\boxed{
\vspace*{-8pt}
\begin{align}
\label{eq:V1}
G_\mu(x_0)& \triangleq \Var\left[D_\mu(x_0)\right] = U_\mu(x_0)-J_\mu(x_0)^2, \\
\textrm{ where~} J_\mu(x) &\triangleq \E\left[D_\mu(x)\right] \textrm{ and }U_\mu(x) \triangleq \E\left[(D_\mu(x))^2\right].\nonumber
\end{align}
}
This risk measure is the {\em overall} variance of the cumulative discounted cost. An alternative is to consider {\em per-period} variance, i.e., the deviations of the single-stage costs, which we will consider for the average-cost MDP in Section \ref{sec:risk-average}. Setting aside the question of which is the most appropriate notion of variability for the underlying MDP, we shall design algorithms for overall variance in the discounted-cost case and per-period variance in the average-cost case in Sections \ref{sec:discounted-MDP-algo}--\ref{sec:avg-MDP-algo}, although the constituent pieces of these algorithms can also be easily incorporated to handle the cases of per-period variance in a discounted cost MDP and overall variance in an average cost MDP. 

\diff{
The variance risk measure $G(x)$ defined by (\ref{eq:V1}) satisfies the following fixed-point equation for any $x \in \X$:
	\begin{align}
 G_\mu(x)&=\chi_\mu(x)+ \gamma^2 \sum_{y\in\X}P(y | x, \mu(x)) G_\mu(y), \label{eq:G-fixedpoint}
\vspace*{-8pt}
\end{align}
where \\
$\chi_\mu(x)=\gamma^2 \left(\sum_{y\in\X}P(y | x, \mu(x)) J_\mu(y)^2-\left(\sum_{y\in\X}P(y | x, \mu(x)) J_\mu(y)\right)^2\right)$.
}

\begin{remark}
The exponential cost risk measure implicitly incorporates the mean-variance trade-off. 
To see this, let $D=\sum_{n=0}^{T-1}k(x_n,a_n)$ denote the total cost r.v.
Using a Taylor's series expansion, we have
$\frac{1}{\beta}\log\mathbb{E}[e^{\beta D(\theta)}]
= \mathbb{E}[D(\theta)] + \frac{\beta}{2}\text{Var}[D(\theta)]+O(\beta^2)$; 
hence, 
the exponential cost risk measure incorporates all higher moments of $D$. 
On the other hand, a mean-variance constrained optimization formulation would involve a variance constraint in \eqref{eq:gen-constrained-opt}, while minimizing the usual expected cost objective function. 
Each formulation has its advantages and disadvantages. 
For example, the constraint formulation eschews choosing the risk parameter $\beta$, although its value can be imputed from the choice of the constraint threshold $\kappa$ in \eqref{eq:gen-constrained-opt}, where $\kappa$ may have a more intuitive meaning in many practical applications. However, the fixed-point relation in \eqref{eq:G-fixedpoint} lacks monotonicity, ruling out policy iteration as a candidate for optimizing variance, whereas the exponential cost satisfies a multiplicative form of Bellman equation (see Section \ref{sec:expcost-algo}), making it more amenable to dynamic programming algorithms.
\end{remark}
\section{Variance in  average-cost MDPs}
\label{sec:risk-average}
\vspace*{-1pt}
For average-cost MDPs, we consider the variance defined by deviations of single-stage cost from the average cost (as opposed to the variance of the average cost itself), viz., 
\boxed{
\vspace*{-8pt}
\begin{align}
G_\mu 
& \triangleq \lim_{T\rightarrow\infty}\frac{1}{T}\E\left[\sum_{n=0}^{T-1}\big(k(x_n,a_n)-J_\mu\big)^2\right],\label{eq:V}
\end{align}}
where the actions $a_n$ are governed by policy $\mu$, and $ J_{\mu}$ is the average cost of the policy $\mu$.
To see the rationale behind the definition above for variability, consider two stream of cost: a policy $\mu^1$ results in $(0,0,0,0,\ldots)$, while another policy $\mu^2$ gives $(100,-100,100,-100,\ldots)$. The average cost as well as the variance of the average cost is zero for both policies. On the other hand, from the point of the variance as defined by \eqref{eq:V}, policy $\mu^1$ is better than $\mu^2$. 

A straightforward calculation yields
\begin{align*}
G_\mu &= \eta_\mu - J_\mu^2,
\end{align*}
where $\eta_\mu=\sum_{x,a}\pi(x,a)k(x,a)^2$ is the average squared cost, with $\pi(x,a)$ denoting the stationary distribution of the state-action pair $(x,a)$ under policy $\mu$.

\diff{Along the lines of \eqref{eq:DifV-Q} and \eqref{eq:DifV}, the
	squared-cost counterparts $W$ and $U$  of $Q$ and $V$ are defined as follows:
\begin{align}
	\label{eq:DifW-U}
	W_\mu(x,a)& \triangleq \sum_{n=0}^\infty\E\big[k(x_n,a_n)^2-\eta_\mu\mid x_0=x,a_0=a,\mu\big], \\
	U_\mu(x) & \triangleq \sum_a\mu(a|x)W_\mu(x,a).
\end{align}
These differential-value and Q-value functions $U$ and $W$ for the square cost, which are defined above, satisfy 	
the	Poisson equations given by
\begin{align*}
\eta_\mu+U_\mu(x) &= \sum_a\mu(a|x)\big[k(x,a)^2+\sum_{y}P(y|x,a)U_\mu(y)\big], \label{eq:DifU-W-Poisson} \\
\eta_\mu+W_\mu(x,a) &= k(x,a)^2+\sum_{y}P(y|x,a)U_\mu(y).
\end{align*}}

\section{Conditional Value-at-Risk (CVaR)}
\label{sec:risk-cvar}

For any random variable \diff{(r.v.)} $X$, the Value-at-Risk (VaR) at level $\alpha\in\left(0,1\right)$ is defined as
\[
\text{VaR}_{\alpha}(X) \triangleq \inf\left\{\xi \mbox{ }|\mbox{ }\mathbb{P}\left(X\leq \xi\right)\geq\alpha\right\},
\]
which mathematically is just \diff{an $\alpha$-quantile}, since if $F$ is the \diff{cumulative distribution function (c.d.f.)} of $X$, VaR is equivalently defined as 
\[
\text{VaR}_{\alpha}(X) \triangleq \inf\left\{\xi \mbox{ }|\mbox{ } F(\xi) \geq\alpha\right\} = F^{-1} (\alpha).
\]
VaR is a commonly used risk measure in the financial industry, where it represents a level of assets needed to cover a potential loss. 
VaR as a risk measure has several drawbacks, which precludes using standard stochastic optimization methods; 
most prominently, VaR is not a coherent risk measure (see Section \ref{sec:risk-coherent} below for a definition).
On the other hand, another closely related risk measure also widely used in the financial industry called CVaR is coherent and thus lends itself to stochastic programming techniques. 
CVaR is a conditional mean over the tail distribution as delineated by the VaR, defined as follows:
\[
\text{CVaR}_{\alpha}(X) \triangleq \mathbb{E}\left[X | X \geq \text{VaR}_{\alpha}(X)\right].
\]

In a stochastic shortest path problem, CVaR is defined as:
\boxed{
\vspace*{-10pt}
\begin{align}
G_\mu(x_0)\triangleq \text{CVaR}_{\alpha}\left[ \left. \sum\limits_{m=0}^{\tau-1} k(x_m,a_m)\right|x_0 \right],\label{eq:cvar-ssp}
\end{align}}
where $\tau$ is the first visit time to absorbing state $0$  and $a_m \sim \mu\left(\cdot | x_m\right)$. 

\section{Chance constraints}
\label{sec:risk-chance}
A chance constraint takes the form
\begin{align}
\mathbb{P}(g(X) \geq 0) \leq \alpha, \label{eq:chance}    
\end{align}
where $X$ is again some random variable of interest and $\alpha$ is a small number (e.g., 0.1, 0.05, 0.001) referred to as the risk tolerance level.  
This fits into the general constrained formulation given by (\ref{eq:gen-constrained-opt}) by taking
$G\triangleq \mathbb{P}(g(X) \geq 0)$. 

\section{Coherent risk measures}
\label{sec:risk-coherent}
A risk measure $\rho(\cdot)$ is coherent if it satisfies the following conditions (where $X$ and $Y$ are random variables): 
\begin{itemize}
	\item Monotonicity:
	If $X\leq Y$ a.s. \diff{(almost surely)}, then $\rho(X)\leq \rho(Y).$
	\item Sub-additivity: $\rho(X+Y)\leq \rho(X)+\rho(Y).$ 
	\item Positive homogeneity: $\rho(\lambda X)=\lambda \rho(X)$ for any $\lambda\geq 0.$
	\item Translation invariance: For constant $a>0,$ $\rho(X+a)=\rho(X)+a.$ 
\end{itemize}
The sub-additivity requirement is vital, and in the context of portfolio optimization implies diversification cannot lead to increased risk. Note that VaR violates this condition, whereas CVaR is a coherent risk measure. 

In the context of an MDP, the r.v.~$X$ could correspond to the total cost in an SSP problem, or the cumulative cost in a discounted MDP, or the long-run average cost. 

\section{Cumulative prospect theory (CPT)}
\label{sec:risk-cpt}

CPT is a risk measure that captures human attitudes towards risk. 
For any r.v.~$X$, the CPT-value is defined as 
\boxed{
\vspace*{-5pt}
\begin{align}
\C(X) &\triangleq \intinfinity w^+\left(\Prob{u^+(X)>z}\right) dz  - \intinfinity w^-\left(\Prob{u^-(X)>z}\right) dz, \label{eq:cpt-general-def}
\end{align}}
where
$u^+,u^-:\R\rightarrow \R_+$ are the utility functions that are assumed to be continuous,
with $u^+(x)=0$ when $x\le 0$ and increasing otherwise, and with $u^-(x)=0$ when $x\ge 0$ and decreasing otherwise,
$w^+,w^-:[0,1] \rightarrow [0,1]$ are weight functions assumed to be continuous, non-decreasing and satisfy $w^+(0)=w^-(0)=0$ and $w^+(1)=w^-(1)=1$.
Note that
CPT-value is a generalization of the classic expected value, which can be seen by taking $w^+,w^-$ as the identity functions,  
$u^+(x)=x^+$ and $u^-(x)=x^-$, where $x^+ \triangleq \max(x,0)$, $x^- \triangleq \max(-x,0)$, 
leading to 
$\C(X) =  \EE{ X^+ } - \EE{ X^- }.$

\begin{figure}[t!]
\centering
\scalebox{0.999}{\begin{tikzpicture}
   \begin{axis}[width=13cm,height=6.5cm,legend pos=south east,
            axis lines=middle,
            xmin=-5,     
            xmax=5,    
            ymin=-4,     
            ymax=4,   
            ylabel={\large\bf Utility},
            xlabel={\large\bf Gains},
            x label style={at={(axis cs:4.7,-0.8)}},
            y label style={at={(axis cs:-0.8,4.8)}},
            xticklabels=\empty,
            yticklabels=\empty
            ]
         \addplot[domain=0:5, green!35!black, very thick,smooth] 
              {pow(abs(x),0.8)}; 
            \addplot[domain=-5:0, red!35!black,very thick,smooth] 
              {-2*pow(abs(x),0.7)}; 
               \addplot[domain=-5:5, blue, thick]           {x};               
               \path(axis cs:0,0) -- (axis cs:5,5);
 		\path(axis cs:-5,-5) -- (axis cs:0,0);
                \path(axis cs:0,0) -- (axis cs:5,0);
                 \path(axis cs:-5,0) -- (axis cs:0,0);
                 \path(axis cs:0,0) -- (axis cs:0,5);
                 \path(axis cs:0,-5) -- (axis cs:0,0);
 \node at (axis cs:  -3.7,-0.45) {\large\bf Losses};
 \node at (axis cs:  4,2.5) {\large $\bm{u^+}$};
 \node at (axis cs:  -1,-3) {\large $\bm{-u^-}$};
   \end{axis}
 \end{tikzpicture}}\\[1ex]
\vspace*{-10pt}
\caption{An example of a utility function. A reference point on the $x$ axis serves as the point of separating gains and losses.
 For losses, the disutility $-u^-$ is typically convex and for gains, the utility $u^+$ is typically concave; 
 both functions are non-decreasing and take the value of zero at the reference point.}
\label{fig:u}
\end{figure}

\begin{figure}[b!]
\centering
\scalebox{0.999}{\begin{tikzpicture}
  \begin{axis}[width=11cm,height=6.5cm,legend pos=south east,
           grid = major,
           grid style={dashed, gray!30},
           xmin=0,     
           xmax=1,    
           ymin=0,     
           ymax=1,   
           axis background/.style={fill=white},
           ylabel={\large Weight $\bm{w(p)}$},
           xlabel={\large Probability $\bm{p}$}
           ]
          \addplot[domain=0:1, red, thick,smooth,samples=1500] 
             {pow(x,0.69)/pow((pow(x,0.69) + pow(1-x,0.69)),1.44)}; 
             \node at (axis cs:  0.8,0.35) (a1) {\large $\bm{\frac{p^{0.69}}{(p^{0.69}+ (1-p)^{0.69})^{1/0.69}}}$};           
             \draw[->] (a1) -- (axis cs:  0.7,0.6);
                 \addplot[domain=0:1, blue, thick]           {x};                      
  \end{axis}
  \end{tikzpicture}}\\[1ex]
\vspace*{-10pt}
\caption{An example of a weight function.
A typical CPT weight function inflates small probabilities and deflates large probabilities, capturing the 
tendency of humans doing the same when faced with decisions of uncertain outcomes.}
\label{fig:w}
\end{figure}

The human preference to play safe with gains and take risks with losses is captured by a concave gain-utility
$u^+$ and a convex disutility $-u^-$.
The weight functions $w^+, w^-$ capture the observed empirical behavior that 
the value seen by a human subject is nonlinear in the underlying probabilities.
In particular, humans deflate high probabilities and inflate low probabilities. 
Examples of utility and weight functions used in practice are shown in Figures \ref{fig:u} and \ref{fig:w}, respectively. 

A risk measure based on CPT in a typical MDP setting could apply the CPT-functional to a risk-neutral objective. 
For instance, take the r.v. $X$ in \eqref{eq:cpt-general-def} to be either the total cost in an SSP problem or the infinite-horizon cumulative cost in a discounted MDP.
However, to carry out dynamic programming would require a Bellman optimality equation, which is not readily available, given the non-convex structure of the CPT-value. 
A different approach is to use a nested formulation, together with the CPT-style probability distortion.
Basically, the formulation is equivalent to optimizing the sum of CPT-value period costs rather than the CPT-value of the sum, and by doing so guarantees the existence of a Bellman optimality equation.  
Intuitively, it makes sense to incorporate CPT for the total reward rather than applying it separately to the reward in each period. 

\section{Bibliographic remarks}
\diff{
What we defined as the ``exponential cost'' risk measure in Section \ref{sec:exp-cost} to be used primarily in Chapter \ref{sec:risk-algos-unconstrained} has an extensive literature in the control and the finance/economics/operations research communities, usually referred to as simply risk-sensitive control and exponential utility, respectively; see the bibliographic remarks in Chapter \ref{sec:intro} for numerous references.
}

The risk measure of Section \ref{sec:risk-discounted} for discounted-cost MDPs was introduced by \textcite{Sobel82VD}, who also derived a fixed-point equation for it. However, as shown there, the operator underlying this equation for variance lacks the monotonicity property. 
The approach of deriving a fixed-point equation for the square value function $U$ and using it to estimate the variance was introduced in an SSP context in \textcite{Tamar13TD}, and later extended to the discounted MDP context in \textcite{prashanth2016mlj}. 
Establishing $T$ as a contraction mapping can be found in  \textcite[Lemma 2]{prashanth2016mlj}.
For the average-cost MDP, the single-stage variance definition can be found in  \textcite{filar1989variance}.

For CVaR and chance constraints, see \textcite{rockafellar2000optimization} and \textcite{NeSh07}.  
Chance-constrained optimization problems
were introduced in \textcite{ChCoSy58}; see also \textcite{MiWa65,Pr70}.
The seminal paper introducing coherent risk measures is \textcite{artzner1999coherent}.

Empirical evidence on human behavior such as the observation that they deflate high probabilities and inflate low probabilities can be found in \textcite{tversky1992advances,Barberis:2012vs}. 
For the weight functions,  \textcite{tversky1992advances} recommend $w(p) = \frac{p^{\eta}}{{(p^{\eta}+ (1-p)^{\eta})}^{1/\eta}}$, whereas \textcite{prelec1998probability} recommends $w(p) = \exp(-(-\ln p)^\eta)$, with $0 < \eta <1$. 
Figure \ref{fig:w} is an example of the former with $p=0.69$. 
In both forms, the weight function has an inverted-s shape,
which is seen to be a good fit from empirical tests on human subjects, as reported by numerous researchers \citep{conlisk1989three,camerer1989experimental,camerer1992recent,harless1992predictions,sopher1993test,camerer1994violations,gonzalez1999shape,abdellaoui2000parameter}.  

The nested formulation approach for CPT was adopted by Lin in his PhD dissertation \citep{lin2013stochastic}; see also \textcite{Lin2013a,Lin2013b,Lin2018,Cavus2014}. 

Finally, we note that there are several risk measures that have not been explored directly in the risk-sensitive RL literature. For instance, spectral risk measure (SRM) \citep{acerbi2002spectral} and utility-based shortfall risk (UBSR) \citep{follmer2002convex}. SRM generalizes CVaR, while retaining coherency. In particular, SRM employs a `risk-aversion' function to weigh losses. Note that VaR gives a zero weight for each loss beyond a given quantile, while CVaR assigns a constant weight in the tail region beyond VaR. On the other hand, the risk-aversion function in SRM allows one to assign larger weights to higher losses, and thus model a user's risk attitude better. Moreover, a positive, increasing risk-aversion function that integrates to one would imply coherency of SRM. Finally, SRMs are also equivalent to the class of distortion risk measures (DRMs), when the risk-aversion function satisfies the aforementioned properties \citep{balbas2009properties}. DRM employs a weight function to distort probabilities, and a concave weight function ensures coherency of DRMs. The CPT risk measure that we presented is more general than DRMs, as the weight function employed there is neither convex nor concave.  Next, UBSR is a special instance of a convex risk measure, which is a generalization of coherency. This can be inferred from the fact that subadditivity and positive homogeneity -- properties imposed for coherency --  imply convexity. UBSR involves a utility function that can be chosen to encode the risk associated with each value of the r.v. $X$, allowing more flexibility in modeling a user's risk attitude, as compared to CVaR. 
See \textcite{follmer2016stochastic} for a textbook introduction to the class of convex risk measures in general and  UBSR in particular. 

%
%
\chapter{Background on Policy Evaluation and Gradient Estimation}
\label{sec:background}
TD-learning and gradient estimation serve as building blocks for policy gradient algorithms, 
in both the risk-neutral and risk-sensitive MDP contexts. 
This chapter provides the basic background on these two topics needed to understand the remainder of the book.  
Two of the most commonly used gradient estimation approaches in policy gradient algorithms, 
simultaneous perturbation stochastic approximation (SPSA) and the likelihood ratio (LR) method, are covered, 
as they are employed in the policy gradient algorithms of Chapters \ref{sec:rpg-template}, \ref{sec:risk-algos}, and \ref{sec:risk-algos-unconstrained}. 
A reader familiar with this background material can skip this chapter. 

\section{Stochastic approximation (SA)}
\label{sec:sa}
\diff{

The goal of stochastic approximation (SA) is to find a root of an unknown real-valued function, denoted here by $H:\R^d \longrightarrow \R$. 
Specifically, SA aims to find a $\theta^*\in \R^d$ that solves the equation $H(\theta^*)  = 0$, where only noisy estimates of $H$ are available, i.e., an estimator $\hat{H} = H + \xi$, where 
$\xi$ is a random variable representing the noise. 
If $\xi$ is zero mean, then the estimator $\hat{H}$ is said to be unbiased.
The primary setting of interest in this book is where $H$ represents a gradient such that a zero of $H(\theta)$, say $\theta^*$, corresponds to a (local) optimum.
}
\subsection{Basic algorithm}
The seminal Robbins-Monro (RM) algorithm solved this problem by employing the following SA iterative update rule:

\boxed{
\vspace*{-10pt}
\begin{align}
\label{eq:sa}
\theta_{n+1} = \theta_n + \zeta(n) (H(\theta_n)+\xi_n),
\end{align}
}
where $\{\zeta(n)\}$ is a step-size sequence; commonly used choices include $\zeta(n) = c/n$ and $\zeta(n)=c$, for some constant $c>0$. In the original RM setting, $\{ \xi_n \}$ is an i.i.d. zero-mean sequence, but 
more generally it could be a martingale-difference sequence (see \ref{ass:noise-sa} below).
\diff{
In the optimization setting where an unbiased estimator can be obtained such as through the LR method described in Section \ref{sec:lr}, the parameter update will take the form \eqref{eq:sa}, with the gradient estimator given by $\hat{H} = H + \xi$.

The RM algorithm can converge even if the function measurements contain an additional bias term that vanishes asymptotically, 
in which case the SA update iteration takes the following form in place of \eqref{eq:sa}:

\boxed{
\vspace*{-10pt}
\begin{align}
\theta_{n+1} = \theta_n + \zeta(n) (H(\theta_n)+\xi_n +\beta_n),\label{eq:sa-general}
\end{align}
}
where $\beta_n$ is an asymptotically vanishing bias term (see \ref{ass:betaVanish} below).
In applications involving biased function measurements, such as SPSA discussed in Section \ref{sec:spsa}, 
the parameter update will take the form \eqref{eq:sa-general}, with (biased) gradient estimator $\hat{H} = H + \xi +\beta$, 
where the bias term $\beta_n$ vanishes asymptotically. 


%
\subsection{Asymptotic convergence}
For the asymptotic analysis of \eqref{eq:sa} and \eqref{eq:sa-general}, we follow the ordinary differential equation (ODE) approach,
where the main idea is to show that the algorithm in \eqref{eq:sa} or \eqref{eq:sa-general} is a noisy discretization of the following ODE:
		\begin{align}
	\dot{\theta}(t) = H(\theta(t)).\label{eq:theta-ode-stoapprox}
\end{align}
In the absence of $\xi_n$ and $\beta_n$, it is apparent that \eqref{eq:sa-general} is a Euler discretization of the ODE defined above, and hence the algorithm \eqref{eq:sa-general} would converge to the equilibria of the aforementioned ODE.
The analysis under the ODE approach would show that the algorithm \eqref{eq:sa-general} tracks the above ODE, even in the presence of noise $\xi_n$ and bias $\beta_n$. For establishing this claim, we make the following assumptions:
\begin{assumption}
\label{ass:LipschitzF}
$H:\R^d \longrightarrow \R$ is Lipschitz continuous.
\end{assumption}
	\begin{assumption}
\label{ass:betaVanish} The sequence $\{\beta_n\}$ is a bounded random sequence with
	$\beta_n \rightarrow 0$ almost surely (a.s.) as $n\rightarrow \infty$.
\end{assumption} 
	\begin{assumption}
	\label{ass:stepsize-sa}
The sequence $\{\zeta(n)\}$ satisfies
	$  \zeta(n)\rightarrow 0 $ and 
	$\sum_{n=0}^{\infty} \zeta(n)=\infty.$
\end{assumption} 
	\begin{assumption}
	\label{ass:noise-sa}
 $\{\xi_n\}$ is a sequence such that for any $\epsilon>0$,
	\[ \lim_{n\rightarrow\infty} \mathbb{P}\left( \sup_{m\geq n}  \left\|
	\sum_{i=n}^{m} \zeta(i) \xi_i\right\| \geq \epsilon \right) = 0. \]
\end{assumption} 
\begin{assumption}
\label{ass:stable}
 $\sup_{n} \l \theta_n \r < \infty$~ a.s.
\end{assumption}
We now discuss these assumptions.
\ref{ass:LipschitzF} ensures that the ODE \eqref{eq:theta-ode-stoapprox} is well-posed. \ref{ass:betaVanish} ensures that the bias $\beta_n$ vanishes asymptotically. \ref{ass:stepsize-sa} contains standard stochastic approximation conditions on the step sizes $\{\zeta(n)\}$. The condition $\sum_n \zeta(n)=\infty$ ensures that the entire time axis is covered, since $\zeta(n)$ can be seen as the discrete time steps, while $\zeta(n) \rightarrow 0$ ensures the discretization errors can be ignored. \ref{ass:noise-sa} imposes conditions on the noise $\xi_n$ that ensure the effect of noise is asymptotically negligible. 

\medskip
\noindent
A typical SA convergence result for the RM algorithm is as follows. 
}
\boxed{
\vspace*{-10pt}
\begin{theorem}
\label{thm:sa-conv}
Assume \ref{ass:LipschitzF}--\ref{ass:stable}.
Then $\theta_n$ governed by \eqref{eq:sa-general} converges a.s.~to the set $\{\theta^* \mid H(\theta^*)  = 0\}$.
\end{theorem}
}
\diff{
Note that for the original RM algorithm given by \eqref{eq:sa}, where the bias term is absent, \ref{ass:betaVanish} is automatically satisfied.
This particular result is often referred to in the SA literature as the Kushner-Clark Lemma. 
Convergence to a set is interpreted as follows. 
If the set consists of a single point, then the convergence would be to that point. 
If all the elements in the set are disconnected, then convergence would be to a single point in the set, with the specific point to which the algorithm converges depending on the initial condition, the step-size sequence, and the noise. 
If some of the points are connected, then the algorithm could ``bounce'' between such points and not converge to a single point.  
These possibilities are illustrated in Figure \ref{fig:setconv}.


\begin{figure}
\begin{tabular}{ccc}
\scalebox{0.72}{\begin{tikzpicture}
	\begin{axis}[
		xlabel=$\theta$,
		ylabel={$f_1(x)$},
		axis x line=middle, 
		axis y line=none,
		xmin=-5, xmax=3,
		ymin=-5,ymax=5,
		xticklabels=\empty,
		yticklabels=\empty,
		]
		\addplot[domain=-5:3,samples=200,thick,blue] {(x^3+3*x^2 -6*x-8)/3 };
		 \node at (axis cs:  -4,4) {$H(\theta)$};
		 \node[circle,fill=red,inner sep=0pt,minimum size=5pt] at (axis cs: -4,0) {};
		 \node[circle,fill=red,inner sep=0pt,minimum size=5pt] at (axis cs: -1,0) {};
		 \node[circle,fill=red,inner sep=0pt,minimum size=5pt] at (axis cs: 2,0) {};
	\end{axis}
\end{tikzpicture}}
&&
\scalebox{0.72}{
	\begin{tikzpicture}[
		declare function={
			func(\x)= (\x<=3) * ( (\x-3)*(\x-3))   +
			and(\x>3, \x<=5) * (0)     +
			(\x>5) * (-8*(\x-5)*(\x-5));
		}
		]
		\begin{axis}[
			axis x line=middle, 
			axis y line=none,
			ymin=-5, ymax=5, xticklabels=\empty,yticklabels=\empty, ylabel=$f_2(x)$,
			xmin=1, xmax=6, xlabel=$\theta$,
			]
			\addplot[blue, domain=1:6,  thick,smooth]{func(x)};
     		\node[ellipse,draw = red] at (axis cs: 4,0) {~~~~~~~~~~~~~~};
			\node at (axis cs:  1.5,4) {$H(\theta)$};
		\end{axis}
	\end{tikzpicture}
}
\end{tabular}
\caption{Functions illustrating the two main types of SA convergence to the zero(s) of the function. In the left graph, the SA algorithm would converge to one of the three points at which the function crosses the x-axis (indicated by the large red circles), where which one it reaches depends on the starting point and the noise. In the right graph, the SA algorithm would eventually bounce between points in the interval (circled in red) on the x-axis unless the noise goes to zero.}
\label{fig:setconv}
\end{figure}

The noise sequence $\{ \xi_n \}$ is generally assumed to be a martingale difference, i.e., $\E(\xi_n \mid \F_n) = 0$, where $\F_n = \sigma(\theta_m, m\le n)$ denotes the underlying $\sigma$-field, 
in which case, \ref{ass:stepsize-sa} and \ref{ass:noise-sa} can be replaced with the following:
}

\begin{assumption}
	\label{ass:stepsize-squaresummable}
	The sequence $\{ \zeta(n) \}$ satisfies
	$  \sum_{n=0}^{\infty} \zeta(n)=\infty, 
	\sum_{n=0}^{\infty} \zeta(n)^2<\infty.$
\end{assumption}	

\begin{assumption}
	\label{ass:martingalegrowth}
$\{\xi_{n}\}$ is a square-integrable martingale-difference sequence satisfying
$\E [\l \xi_{n} \r^2 \mid \F_n] \le C_0 (1 + \l \theta_n \r^2)~ \forall n \geq 0$, 
for some constant $C_0$.
\end{assumption}	

\medskip
To see that the above two assumptions in conjunction with \ref{ass:stable} imply \ref{ass:noise-sa}, we use Doob's martingale inequality, given as follows:  \\
For a martingale sequence $\{W_m\}$,
\begin{align}
	\Prob{\sup_{m\geq 0}   \left\|W_m\right\| \geq \epsilon} \le \dfrac{1}{\epsilon^2} \lim_{m\rightarrow \infty} \E \left\|W_m\right\|^2. 
	\label{eq:doob}
\end{align}
Apply the inequality above to 
$\{\sum_{n=k}^{l} \zeta(n) \xi_{n}\}_{l\ge k}$ to obtain
\begin{align*}
	\lim_{k\rightarrow \infty}	\mathbb{P}\left( \sup_{l\geq k}   \left\|\sum_{n=k}^{l} \zeta(n) \xi_{n}\right\| \geq \epsilon \right) \le & \dfrac{1}{\epsilon^2}\lim_{k\rightarrow \infty} \sum_{n=k}^{\infty} \zeta(n)^2 \E\left\| \xi_{n}\right\|^2 \\
	\le & \dfrac{\textrm{const}}{\epsilon^2} \lim_{k\rightarrow \infty}\sum_{n=k}^{\infty} \zeta(n)^2 =0, \end{align*}
where the final inequality used the following facts: (i) $\E\left\| \xi_{n}\right\|^2$ is bounded above since the iterate $\theta_n$ is bounded a.s. from \ref{ass:stable} and the noise satisfies a linear growth condition as specified in \ref{ass:martingalegrowth}; (ii) the step sizes are square summable from \ref{ass:stepsize-squaresummable}. Thus, \ref{ass:noise-sa} is satisfied.

SA is useful in solving several subproblems in risk-sensitive RL. For example, TD-learning is an instance of an SA algorithm that incorporates a fixed-point iteration. While regular TD-learning is useful in estimating $J(\theta)$, a variant will be useful in estimating variance indirectly (see Section \ref{sec:discounted-MDP-algo}). Moreover, VaR estimation is performed using an SA scheme that features a stochastic gradient descent-type update iteration, while CVaR estimation is a plain averaging rule that can be done through SA, as well.

\diff{If $H$ represents a gradient, say $H=\nabla h$, the RM algorithm becomes a stochastic gradient descent (for minimization problems) scheme with the following SA iterate updating equation:
\boxed{
\vspace*{-10pt}
\begin{align}
\label{eq:sg}
\theta_{n+1} = \theta_n - \zeta(n) \widehat\nabla h(\theta_n),
\end{align}}
where $\widehat\nabla h(\theta_n) = \nabla h(\theta_n)+\xi_n+\beta_n$ denotes the gradient estimator. 

\begin{figure}
\begin{tabular}{ccc}
\scalebox{0.72}{\begin{tikzpicture}
	\begin{axis}[
		xlabel=$\theta$,
		axis x line=bottom, 
		axis y line=left,
		xmin=-5, xmax=3,
		ymin=-5,ymax=5,
		xticklabels=\empty,
		yticklabels=\empty,
		]
		\addplot[domain=-5:3,samples=200,thick,blue] {(x^4/4+x^3 -3*x^2-8*x+4)/3 };
		 \node at (axis cs:  -4,4) {$h(\theta)$};
		 \node[circle,fill=red,inner sep=0pt,minimum size=5pt] at (axis cs: -4,-4) {};
		 \node[circle,fill=purple,inner sep=0pt,minimum size=5pt] at (axis cs: -1,2.7) {};
		 \node[circle,fill=red,inner sep=0pt,minimum size=5pt] at (axis cs: 2,-4) {};
	\end{axis}
\end{tikzpicture}}
&&
\scalebox{0.72}{
	\begin{tikzpicture}[
		declare function={
			func(\x)= (\x<=3) * (0.5+2*(\x-3)*(\x-3))   +
			and(\x>3, \x<=4) * (0.5)     +
			and(\x>4, \x<=5) * (0.5+(\x-4)*(\x-4)) + 
			(\x>5) * (2.5+1*(\x-6)*(\x-6)*(\x-6));
		}
		]
		\begin{axis}[
			axis x line=bottom, 
			axis y line=left,
			ymin=0, ymax=5, xticklabels=\empty,yticklabels=\empty, 
			xmin=2, xmax=8, xlabel=$\theta$,
			]
			\addplot[blue, domain=1:8,  thick,smooth]{func(x)};
	     		\node[ellipse,draw = red] at (axis cs: 3.5,0.5) {~~~~~~};
			\node[circle,fill=red,inner sep=0pt,minimum size=5pt] at (axis cs: 6,2.5) {};
			\node at (axis cs:  2.5,4.5) {$h(\theta)$};
		\end{axis}
	\end{tikzpicture}
}
\end{tabular}
\caption{Two graphs illustrating the types of SA convergence for stochastic optimization. In the left graph, an SA algorithm for minimization would converge to one of the two local minima or the local maximum indicated by the filled (red) circles, where which one it reaches depends on the starting point and the noise.  In the right graph, the SA algorithm could converge to the saddle point indicated by the filled (red) circle or would eventually bounce between points in the circled (in red) interval unless the noise goes to zero. As long as the gradient estimate remains appropriately noisy, the SA algorithm would eventually move away from the local maximum in the left graph and away from the saddle point in the right graph.}
\label{fig:setconv2}
\end{figure}

In the context of RL, the policy parameter updates in a policy gradient algorithm for solving risk-neutral/risk-sensitive MDPs are of this form.
A straightforward specialization of the result in Theorem \ref{thm:sa-conv} leads to the following result:
\vspace*{-4pt}
\boxed{
\vspace*{-10pt}
\begin{theorem}
\label{thm:sa-conv2}
Assume \ref{ass:LipschitzF}--\ref{ass:stable}.
Then $\theta_n$ governed by \eqref{eq:sg} converges a.s. to the set $\{\theta^* \mid \nabla h(\theta^*)  = 0\}$.
\end{theorem}
}
As in the root-finding setting, 
if the set consists of a single point, then the convergence would be to that point. 
Otherwise, the meaning of convergence to a set is depicted by two graphs in Figure \ref{fig:setconv2}. 
If all the elements in the set are disconnected, then convergence would be to a single point in the set, with the specific point to which the algorithm converges depending on the initial condition, the step-size sequence, and the noise, as illustrated in the left graph of Figure \ref{fig:setconv2}, which contains two local minima and one local maximum. 
If some of the points are connected, then the algorithm could ``bounce'' between such points and not converge to a single point, as illustrated in in the right graph of Figure \ref{fig:setconv2}, which contains a flat local minimal region and a saddle point.  
``Unstable'' points such as local maxima (in minimization problems) and saddle points can be avoided by ensuring that the gradient estimate is suitably noisy, to be described in more detail now.

Since the ODE tracked by the iteration \eqref{eq:sg} is $\dot{\theta}(t) = \nabla h(\theta(t))$, we know that its stationary points will be local maxima or minima, saddle points, or points of inflection. If these points are isolated, then the algorithm \eqref{eq:sg} will a.s. converge to a sample path-dependent stationary point. 
Under additional assumptions, one can ensure convergence to a local minimum, i.e., avoid local maxima and saddle points. One such assumption is that the stationary points are hyperbolic, i.e., the Hessian $\nabla^2 h$ does not have eigenvalues on the imaginary axis. Then locally, it has a ‘stable manifold’ of dimension equal to the number of eigenvalues in the left half plane and an unstable manifold with the complementary dimension. A trajectory on the former converges to the stationary point along the stable manifold, whereas one on the latter moves away from it on the unstable manifold. A trajectory initiated anywhere else also eventually moves away. Thus, if there is at least one unstable eigenvalue, the trajectories move away from the stationary point except on the stable manifold, a set of zero Lebesgue measure. Hence, if the noise is omnidirectional, i.e., rich in all directions in a certain precise sense, the iterations will be pushed away from the stable manifold often enough for the iterates to move away from the stationary point for good, a.s. Then the iterates will a.s. converge to a local minimum, where there are no unstable directions. In case the conditions on noise cannot be verified for the problem at hand, one can always add extraneous i.i.d. zero mean noise, i.e., an SA update iteration of the form
\begin{align}
	\label{eq:sg-noiseextra}
	\theta_{n+1} = \theta_n - \zeta(n) (\widehat\nabla h(\theta_n) + \varphi_{n}),
\end{align}
where $\varphi_{n}$ is extraneous noise added to ensure that the algorithm avoids saddle points/local maxima. A simple choice is to sample $\varphi_n$ from the $d$-dimensional unit sphere uniformly.
In practice, it may not be necessary to add such a noise factor extraneously, since the algorithm has an inherent noise component in the gradient estimates.
}

\diff{
\subsection{Projected stochastic approximation}
\label{sec:projSA}
Theorem \ref{thm:sa-conv} imposes a stability requirement on the iterates, i.e., the condition $\sup_{n} \l \theta_n \r < \infty$.  This requirement is not easy to ensure in RL applications,  where one considers a policy gradient-type algorithm for finding the optima of a non-convex objective function. In such situations, an alternative is to employ projections to artificially ensure stability of iterates. 
	
A projected stochastic approximation algorithm would involve the following update iteration:
\boxed{
\vspace*{-4pt}
	\begin{equation}
	\label{eq:kush-cla}
	\theta_{n+1} = \Gamma(\theta_{n} + \zeta(n)(H(\theta_n) + \xi_n + \beta_n)), 
\end{equation}
}
where $\Gamma$ is a projection into a compact and convex set, say $\Theta \subset \R^d$. 

The ODE associated with (\ref{eq:kush-cla}) is given by
\begin{equation}
	\label{eq:kushcla-ode}
	\dot \theta = \check{\Gamma}(H(\theta)),
\end{equation}
where $\check\Gamma$ is a projection operator that keeps the ODE evolution within the set $\Theta$, defined as follows:
For any bounded continuous function $f(\cdot)$,
\begin{align}
	\label{eq:Pi-bar-operator-bg}
	\check{\Gamma}\big(f(\theta)\big) = \lim\limits_{\tau \rightarrow 0}
	\dfrac{\Gamma\big(\theta + \tau f(\theta)\big) - \theta}{\tau}.
\end{align}
The limit defined above exists because $\Theta$ is convex. 
Furthermore, for $\theta$ in the interior of $\Theta$, the projection $\check{\Gamma}(f(\theta)) = f(\theta)$, while for $\theta$ on the boundary of $\Theta$, $\check{\Gamma}(f(\theta))$ is the projection of $f(\theta)$ onto the tangent space of the boundary of $\Theta$ at $\theta$.

The following theorem presents an asymptotic convergence result for the projected SA iteration \eqref{eq:kush-cla}, with assumptions similar to those used in Theorem \ref{thm:sa-conv} (sans the stability requirement \ref{ass:stable}). 
\boxed{
\vspace*{-10pt}
	\begin{theorem}\textit{(Projected stochastic approximation)}
	\label{thm:kc}
	Assume \ref{ass:LipschitzF}--\ref{ass:noise-sa}.
	Let $\Theta^* = \{ \theta \mid \check{\Gamma}(H(\theta))) = 0 \}$ denote the set of limit points of the ODE \eqref{eq:kushcla-ode}. 
	Then $\theta_n$ governed by \eqref{eq:kush-cla} converges a.s. to the set
	$\Theta^*$.
\end{theorem}}}
\vspace*{-2pt}
\noindent
\diff{
This result is the projected form of the Kushner-Clark Lemma.
}

\diff{
	\subsection{A stability result}
	\label{sec:stability}
\vspace*{-5pt}
Recall that the asymptotic convergence result in Theorem \ref{thm:sa-conv} requires that the iterate $\theta_n$ remains bounded a.s.
The variant in Theorem \ref{thm:kc} ensured stability through a projection operator $\Gamma$. However, one can do away with the projection operator under certain conditions, and infer both boundedness as well as convergence. The result in this section presents conditions for ensuring stability, and these conditions are usually satisfied in the context of policy evaluation, esp. through TD learning methods (see Section \ref{sec:td}).

\begin{assumption}
	\label{ass:scaled-ode}
 For any $\eta \in \R$, define 
 \begin{align}
 	H_{\eta}(\theta) = H(\eta \theta)/\eta.
 	\label{eq:scaled-ode}
 	\end{align}
 Then, there exists a continuous function $H_\infty$ such that $H_{\eta} \rightarrow H_{\infty}$ as $\eta \rightarrow \infty$ uniformly on compact sets. Furthermore, $\theta^*$ is the (unique) globally asymptotically stable equilibrium for the ODE
\begin{align}\label{eq:asym-q}
	\dot \theta(t) =  H_\infty(\theta(t)).
\end{align}
\end{assumption}


Assumption \ref{ass:scaled-ode} can be interpreted intuitively as follows: 
Consider the scaled ODE \eqref{eq:scaled-ode}, which arises by scaling the iterate to lie within a unit ball and linearly interpolating between the scaled iterate values. The assumption requires that the limit of these scaled functions $H_\eta$ exist, and the limiting ODE has a globally asymptotically stable equilibrium. Under these conditions, together with \ref{ass:martingalegrowth}, which implies the effects of the noise is asymptotically negligible, we obtain the following stability result for the original (unprojected) SA algorithm.  

\boxed{\begin{theorem}
\vspace*{-10pt}
		\label{thm:borkarmeyn}
		Assume \ref{ass:LipschitzF}, \ref{ass:martingalegrowth}, and \ref{ass:scaled-ode}.
		Then for $\theta_n$ governed by \eqref{eq:sa},  
		$\sup_n\l \theta_n \r <\infty$ a.s. for any $\theta_0$.
		Furthermore,
        $\theta_n$ converges a.s.~to the set $\{\theta^* \mid H(\theta^*)  = 0\}$.
	\end{theorem}}	
}

\diff{\section{Contractive stochastic approximation}
\label{sec:contractive-SA}
In many RL problems, the underlying operator is contractive in nature, and the goal is to find the fixed point of such a contraction mapping by observing a sample path of the underlying MDP. Stochastic approximation facilitates finding such a fixed point, and we formalize this claim below.

Given a vector $\nu =(\nu(1),...,\nu(|\X|))$, with $\nu(i)>0,\ \forall i$, define the weighted maximum norm of a vector $\theta = (\theta(1), \ldots, \theta(|\X|))$ by 
\begin{align*}
	\left\Vert \theta \right\Vert_{\nu} = \max_{i} \frac {|\theta(i)|} {\nu(i)}.
\end{align*}
If $ \nu(i)=1,\  \forall  i $, then $\left\Vert \cdot \right\Vert_{\nu}$ is the max-norm or $\ell_\infty$ norm.

Suppose that  $H$ is a weighted max-norm contraction, i.e. $\exists$ a positive vector $\nu =(\nu(1),...,\nu(|\X|))$, and a constant $\beta \in [0, 1)$ such that
\begin{align*}
	\left\Vert H(\theta) - H(\theta')\right\Vert_{\nu} \leq \beta \left\Vert \theta - \theta'\right\Vert \hspace{1cm} \forall \theta, \theta' \in \R^{|\X|}.
\end{align*}
It is well known that there exists an $\theta^{*}$ that is the unique fixed point of the contraction mapping $H$, i.e. $H(\theta^{*})=\theta^{*}$.

To see the connection of the norm defined above, recall from the theory of MDPs in Chapter \ref{sec:mdps} that (i) in an SSP with all policies proper, the Bellman operator $T$, as well as the policy-specific operator $T_\mu$, are contractions under a weighted-max norm; and (ii) in a discounted MDP with bounded single-stage cost, the Bellman operator $T$, as well the policy-specific operator $T_\mu$, are contractions under the max-norm. For the Bellman operator, the fixed point $\theta^*$ would correspond to the optimal cost, while $\theta^*$ would be the expected total cost for the policy-specific operator $T_\mu$.

An SA iteration for finding $\theta^*$ would take the following form:
\boxed{
\vspace*{-10pt}
\begin{align}{\label{eq:stItAlgo}}
	\theta_{n+1} = \theta_{n} + \zeta(n) (H(\theta_{n}) + \xi_{n} - \theta_n),
\end{align}
}
where $\xi_n$ is the noise element and $\zeta(n)$ is the step size, as in the previous section. The SA algorithm uses the noisy observation $H(\theta_{n})(i) + \xi_{n}(i)$ to perform an incremental update in each component $i$. In RL settings, each component would denote a state, and $\theta_n$ would be an estimate of the value function or the optimal value function, based on whether the problem is value prediction or control, respectively.

We now state an asymptotic convergence result for the iterate $\theta_n$ to the fixed point $\theta^*$ of $H$. 
	\boxed{
\vspace*{-5pt}
		\begin{theorem}{\label{thm:contract-sto-approx}}
			Assume  \ref{ass:stepsize-sa}, \ref{ass:noise-sa}, \ref{ass:stable}, and that $H$ is a weighted max-norm contraction.
			Then $\theta_n$ governed by \eqref{eq:stItAlgo} converges a.s. to the fixed point of $H$, 
			i.e., $\theta_n \longrightarrow \theta^*$ a.s., where $\theta^*=H(\theta^{*})$.
		\end{theorem}
	}
}

\section{Temporal-difference (TD) learning}
\label{sec:td}
A key algorithm for policy evaluation in RL is TD learning. The objective of TD-learning is to estimate the value function $J_\mu(x)$ for a given policy $\mu$.
In this section, we first present tabular TD learning, i.e., a setting where the state space is small allowing one to store a lookup table with an entry for each state. Subsequently, we cover TD learning with linear function approximation -- an algorithm that can handle large state spaces by employing feature-based representations.

\subsection{Tabular TD learning}
\diff{For ease of exposition, we consider the TD(0) algorithm in a discounted-cost MDP setting. 
Recall that the value function $J_\mu(x)$ satisfies the following fixed point relation:
\begin{equation} 
\label{eq:TD-fixed_point}
	J_\mu(x) = \EE{k(x,a)+\gamma J_\mu(y)},
\end{equation}
where the expectation is over the random action $a$ chosen according to $\mu(\cdot | x)$, and the next state $y$, which is sampled from $P(\cdot | x,a)$. Now, in order to estimate the expectation, we take a sample of the expression within the expectation, and apply the update rule as shown below:
Starting with any $J_0$, the TD(0) algorithm iteratively updates an estimate $J_{n+1}$  at iteration $n+1$ using the observed sample cost $k(x_n,a_n), ~a_n \sim \mu^\theta\left(\cdot | x_n\right)$, and previous estimate $J_n$ as follows:
\boxed{
\vspace*{-2pt}
\begin{equation} \label{eq:td-fullstate}
\hspace*{-8pt}
J_{n+1}(x) \hspace*{-2pt}=\hspace*{-2pt} J_n(x) + \zeta(\nu(x,n))\indic{x_n=x} \hspace*{-3pt}[k(x_n,a_n) + \gamma J_n(x_{n+1}) - J_n(x_n)],
\end{equation}
}
\vspace*{-3pt}
where 
$x_{n+1} \sim P(\cdot | x_n,a_n)$, 
$\nu(x,n) = \sum\limits_{m=0}^n \indic{x_m = x}$ is the number of visits to state $x$, 
and $\{\zeta(\cdot)\}$ is a step-size sequence.
Note that \eqref{eq:td-fullstate} is an iterative means of trying to find the zero of the fixed-point equation given by \eqref{eq:TD-fixed_point} by taking the difference between estimates of the value function given by $J_n(x_n)$ for the LHS of \eqref{eq:TD-fixed_point} 
and that given by $k(x_n,a_n) + \gamma J_n(x_{n+1})$ for the RHS of \eqref{eq:TD-fixed_point}.

Since $\EE{k(x,a)+\gamma J(y)} = T_{\mu}J(x)$, the TD(0) update rule 
\eqref{eq:td-fullstate} is equivalent to
\begin{align}
	J_{n+1}(x) = J_{n}(x) + \zeta(\nu(x,n))\indic{x_n=x}(T_{\mu}J_{n}(x)  - J_{n}(x) + \xi_n(x)),
\end{align}
where $\xi_n$ is the noise term. 
From the equation above, we can draw the parallel to the stochastic approximation algorithm presented in the previous section, in particular, to observe the $\xi_n$ term is conditionally zero mean, and satisfies the linear growth condition in Theorem \ref{thm:contract-sto-approx}, leading to the following convergence result:
\boxed{\begin{theorem}
	For a discounted MDP with bounded single-stage cost, the TD(0) algorithm  \eqref{eq:td-fullstate} using step sizes satisfying \ref{ass:stepsize-squaresummable}
	converges a.s., i.e.,
	\[ J_n \rightarrow J_\mu \textrm{ a.s as } n\rightarrow \infty.\]	
\end{theorem}	}
A similar claim can be made for the case of other MDPs (SSP and average cost), and we omit the details.}

\subsection{TD learning with linear function approximation}
While the TD(0) algorithm described above is provably convergent to the true value $J_\mu(x_0)$, this algorithm employs full-state representation, i.e., it requires a lookup table entry for each state $x\in\X$, and thus would be subject to the curse of dimensionality, in terms of potentially intractable growth of the size of the state space.
A practical approach to address this problem is to employ feature-based representations and function approximation by approximating the value function as follows:
\[J_\mu(x) \approx v\tr \phi(x),\]
where
$\phi(x)$ is a $d$-dimensional feature (column) vector corresponding to the state
 $x$, with $d \ll |\X|$ and $v$ is a tunable $d$-dimensional  parameter.
Given this approximation architecture, an important question is how to choose $v$ so that we obtain a good enough approximation of $J_\mu$ within a linear space. The TD approach is to find a $v$ that solves the following projected system of equations: 
\begin{align}
\Phi v =& \Pi T_\mu(\Phi v), \label{eq:td-fixedpt-projected}
\end{align}
where $\Phi$ is a matrix with rows $\phi(x)\tr ~\forall x \in \X$, $T_\mu J = k + \gamma P_\mu J$ is the discounted-cost MDP Bellman operator \eqref{T_eqn_mu} underlying the fixed-point equation for policy $\mu$ given in Proposition \ref{prop-SSPfixed},  
with $P_\mu$ representing the transition probability matrix of the Markov chain generated by $\mu$, 
and $\Pi$ is an operator that projects onto the linear space $\S=\{\Phi v | v \in \R^{d}\}$. More precisely,
assuming a stationary distribution, say $\psi$, exists for the Markov chain generated by policy $\mu$, we have $\Pi=\Phi(\Phi\tr D \Phi)^{-1}(\Phi\tr D)$, with $D$ denoting a diagonal matrix with entries from the distribution $\psi$. Define a weighted $\ell_2$-norm as 
\[ \| J \|^2_{\psi} = \sum_{i=1}^{|\X|} \psi(i) J(i)^2, \textrm{ for any } J \in \R^{|\X|}.\]
Then, $\Pi$ can be seen as the orthogonal projection operator onto the set $\S$ under the norm defined above, i.e., for any $J$,
\[\Pi J=\underset{\bar J \in \S}{\arg\min}  ||J- \bar J||_{\psi}^2.\]

The projected fixed-point relation in \eqref{eq:td-fixedpt-projected} can be written equivalently as a linear system of equations, i.e.,  
\begin{align*}
\Phi v &= \Pi T_\mu(\Phi v)	\Leftrightarrow C v = d, \textrm{ where }\\
C &= \Phi\tr D (I - \gamma P_\mu) \Phi, d=\Phi\tr D \mathbf{k},
\end{align*}
and $\mathbf{k}$ is a $|\X|$-dimensional vector with elements $\sum_a k(x,a)\mu(a | x)$. 
The above equivalence can be seen by noting the following:  
\begin{eqnarray*}
		C v - d &=& \Phi\tr D (I - \gamma P_\mu) \Phi v - \Phi\tr D \mathbf{k} \\
		&=& \displaystyle \sum_{x,y} \psi(x) P_{xy} \phi(x) \left( \phi(x)\tr v - \gamma \phi(y)\tr v - \sum_a k(x,a)\mu(a | x) \right) \\
		&=& \E_\psi \left[\phi(x) \left( \phi(x)\tr v - \gamma \phi(y)\tr v - \sum_a k(x,a)\mu(a | x) \right)\right],
\end{eqnarray*}
where 
$\E_\psi$
denotes expectation with respect to one step in a Markov chain generated by policy $\mu$ that starts in the stationary distribution $\psi$
and $P_{xy}$ is the $x$-$y$th entry in the one-step transition matrix $P_\mu$. 

Thus, finding the TD fixed point is equivalent to obtaining a $v$ such that the expectation on the RHS above is zero. The obvious method for finding such a $v$ requires sampling a state $x$ from the stationary distribution $\psi$ and the next state $y$ from $P_{x \cdot} \triangleq P(\cdot | x, \mu(x))$. However, in practice, the stationary distribution $\psi$ is unknown, so sampling from it may not be feasible. Instead,
assuming an initial distribution $\nu_0$, the samples seen by TD(0) would be coming from $\nu_0 P_\mu^n$, which under suitable mixing assumptions converges to the stationary distribution $\psi$. 
This motivates the following update rule for the TD(0) algorithm:
\boxed{
\vspace*{-10pt}
\begin{align}
v_{n+1}  &= v_n + \zeta(n) \phi(x_n)(k(x_n,a_n) +
              \gamma v_n\tr \phi(x_{n+1}) \!-\! v_n\tr \phi(x_n)),~~~~~~~ \label{eq:td-fa}
\end{align}}
where $v_0$ is set arbitrarily, $a_n \sim \mu(\cdot | x_n)$ and $\{\zeta(n)\}$ is a step-size sequence satisfying standard SA conditions. 

\diff{The convergence analysis of the TD(0) algorithm utilizing \eqref{eq:td-fa} is more complicated than the usual RM-based TD algorithms presented previously. The stochastic approximation schemes presented in earlier sections assumed that the noise elements came from a martingale difference sequence, whereas iteration \eqref{eq:td-fa} has a transient mixing phase before the samples are seen from the stationary distribution $\psi$.

To show that TD(0) with linear function approximation converges to solution of $Cv-d=0$, we make the following assumptions:

\begin{assumption}
\label{ass:ergodic}
The Markov chain induced by the policy $\mu$ is irreducible and aperiodic. Moreover, there exists a stationary distribution $\psi(=\psi_\mu)$ for this Markov chain. Let $\E_\psi$ denote the expectation w.r.t. this distribution.
\end{assumption}

\begin{assumption}
\label{ass:fullColFeatures}
The matrix $\Phi$ with rows $\phi(x)\tr, \forall x \in \X$ has full column rank. 
\end{assumption}

\begin{assumption}
\label{ass:bddCost}
The single-period cost function satisfies $\E_\psi(k^2(x,\mu(x))) < \infty$, $\forall x \in \X$.
\end{assumption}

\begin{assumption}
\label{ass:bddFeature}
The feature vector $\phi_i(x)$ satisfies $\E_\psi(\phi_i^2(x)) < \infty,
~\forall x\in\X$, $i=1,\ldots,d$.
\end{assumption}


\begin{assumption}
\label{ass:mixing}
For the Markov chain $\{x_t\}$ with stationary distribution $\psi$ induced by policy $\mu$, 
there exists a non-negative bounded function $B(\cdot)$ such that for any $q>1$, there exists a non-negative constant $K_q < \infty$ satisfying
$\E[B^q(x_m)\mid x_0] \le K_q B^q(x_0)$ 
for all $x_0 \in \X$.  
Furthermore, 
	\begin{align}  
		&\sum\limits_{n=0}^\infty \left\| \E[k(x_{n},\mu(x_{n}))\left.\phi(x_n)\right| x_0] - \E_\psi[k(x_{n},\mu(x_{n}))\phi(x_n)]\right\| \le B(x_0), \label{eq:mixing-bd1}\\
		&\sum\limits_{n=0}^\infty \left\| \E[\phi(x_n)\phi(x_{n+m})\tr \mid x_0] - \E_\psi[\phi(x_n)\phi(x_{n+m})\tr]\right\|\le B(x_0). \label{eq:mixing-bd}
	\end{align}
\end{assumption}

We briefly discuss the assumptions.
\ref{ass:ergodic} is an ergodicity requirement that is necessary to ensure that the operator $\Pi T_\mu$ is a contraction mapping w.r.t. the weighted max-norm $\l \cdot \r_\psi$, with the weights coming from the stationary distribution vector $\psi$. A full column rank feature matrix (assumed in \ref{ass:fullColFeatures}) together with the fact that $\Pi T_\mu$ is a contraction mapping imply that the fixed point $v$ is unique. Next, \ref{ass:bddCost} and \ref{ass:bddFeature} are integrability requirements, which ensure that the effect of noise in the TD(0) update vanishes asymptotically. 
The conditions in \ref{ass:mixing} are related to the mixing of the Markov chain generated by the given policy $\mu$. These conditions ensure that taking expectation of quantities relevant to the update \eqref{eq:td-fa} w.r.t. the distribution $\nu_0 P^n$, where $\nu_0$ is the initial distribution, is close to the one with the stationary distribution.

The main result establishing asymptotic convergence of TD(0) with linear function approximation is given below.
\boxed{\begin{theorem}
For a discounted MDP, assume \ref{ass:ergodic}--\ref{ass:mixing}. Then the TD(0) algorithm for $v_n$ governed by \eqref{eq:td-fa} using step sizes satisfying \ref{ass:stepsize-squaresummable} converges a.s. to the fixed-point solution of the following projected Bellman equation:
\[  \Phi v = \Pi T_\mu(\Phi v), \]
where $T_\mu$ is the Bellman operator corresponding to policy $\mu$ and $\Pi$ is the orthogonal projection onto the linearly parameterized space $\{\Phi v \mid v\in \R^d\}$, with $\Phi$ denoting the feature matrix with rows $\phi(x)\tr, \forall x \in \X$.
\end{theorem}}
}
\subsection{Average-cost TD learning}
TD-learning can be employed to estimate the differential value function $V(\theta,x)$ in an average-cost MDP setting, with the Poisson equation in \eqref{eq:DifV-Q-Poisson} as the basis. Notice that the latter equation contains the average cost $J(\theta)$, which has to be estimated from sample data, and then plugged into the TD-learning update rule for estimating $V$. The TD(0) variant in this case, with policy $\mu^\theta$, would update the estimate $V_{n+1}$ as follows:
\boxed{
\vspace*{-10pt}
\begin{align}
 V_{n+1}(x) &= V_n(x) + \zeta(\nu(x,n))\indic{x_n=x}\Big(k(x_n,a_n) - \hat J_n ~~~~~~~~~ \nonumber\\
 & ~~~~~~~~~~~~~~~~~~~~~~~~~~~~~~+ V_n(x_{n+1}) - V_n(x_n)\Big),\label{eq:td-fullstate-avg}  \\
    \hat{J}_{n+1} &= (1-\alpha_n) \hat{J}_n + \alpha_n k(x_n,a_n),~~~~~ \label{eq:avcost-est}
\end{align}
}
where $\hat J_n$ is the average of the sample single-stage costs seen up to $n$,
$a_n \sim \mu^\theta(\cdot  | x_n)$, and $\alpha_n \in (0,1)$. 

The parameter update rule in \eqref{eq:td-fullstate-avg} can be understood using arguments analogous to those used in the discounted-cost setting for \eqref{eq:td-fullstate}, modulo the additional need to estimate the average cost of the given policy, a quantity denoted by $\hat J_n$, which is also updated iteratively using \eqref{eq:avcost-est}, 
necessitated by the fact that the differential value function satisfies a fixed-point relationship that involves the true average cost 
(refer to Proposition \ref{prop:Poission} and \eqref{eq:DifV-Q-Poisson}.)

\diff{The function-approximation variant of TD in the average-cost setting would involve the following update iterations:
\boxed{
\vspace*{-10pt}
\begin{align*}
\delta_n&= k(x_n,a_n)-\hat{J}_{n+1}+v_n\tr \phi(x_{n+1})-v_n\tr \phi(x_n),\\
v_{n+1} & =  v_n +\alpha_n\delta_n \phi(x_n),
\end{align*}
}where $\hat{J}_n$ is updated using \eqref{eq:avcost-est}, as in the tabular TD(0) case.}

\section{Simultaneous perturbation stochastic approximation (SPSA)}
\label{sec:spsa}
Suppose we want to solve the optimization problem 
$$
\min_\theta h(\theta) \triangleq \E\left[\hat{h}(\theta, \xi)\right],
$$
where $\hat{h}$ denotes a noisy unbiased estimator for $h$, which itself is not directly available, and $\xi$ denotes the underlying randomness (noise). 
For this stochastic optimization problem,  
the Kiefer-Wolfowitz (KW) algorithm performs gradient descent using a finite-differences estimate for $\nabla h$ as follows: 
\boxed{
\vspace*{-10pt}
\begin{align}
\theta_{n+1} &= \theta_n - \zeta(n) \widehat \nabla h(\theta_n), \label{eq:sg-alg}\\
\widehat \nabla_i h(\theta_n) &= \left(\frac{\hat{h}(\theta_{n} + \delta_n e_i, \xi_{n,i}^+)
- \hat{h}(\theta_{n} - \delta_n e_i, \xi_{n,i}^-)}{2\delta_n} \right)\hspace*{-1pt}, i=1,\ldots,d,~~~~~~~ \label{eq:FD}
\end{align}
}
where $\{\zeta(n)\}$ is a step-size sequence satisfying standard SA conditions, 
$\widehat \nabla_i h$ denotes the $i$th element of the gradient estimator, 
$\{\delta_n\}$ is a sequence of positive perturbation constants,
$\xi^+_{n,i}$ and $\xi^-_{n,i}$ are the noise components, and
$e_i$ is the unit vector in the $i$th direction.
It can be shown that $\widehat \nabla_i h(\theta)$ approaches $\nabla_i h(\theta)$ if $\delta_n \rightarrow 0$ as $n \rightarrow \infty$.
However, this symmetric finite-differences gradient estimator \eqref{eq:FD} requires $2d$ samples of 
$\hat{h}$ for each iteration of \eqref{eq:sg-alg}, whereas 
the SPSA algorithm requires {\it only two samples in each iteration, regardless of the parameter dimension $d$}, estimating the gradient as follows:
\boxed{
\vspace*{-10pt}
\begin{align}
\widehat \nabla_i h(\theta_n) =  \left(\frac{\hat{h}(\theta_n + \delta_n \Delta(n), \xi_n^+)
- \hat{h}(\theta_n - \delta_n \Delta(n), \xi_n^-)}{2\delta_n
\Delta_{i}(n)} \right),~~~~~ \label{eq:spsa-grad}
\end{align}
}
where 
$\Delta(n) = (\Delta_{1}(n),\ldots, \Delta_{d}(n))\tr$ is a random perturbation vector, 
with each $\Delta_{i}(n)$ chosen to be symmetric $\pm 1$-valued Bernoulli r.v.s, and 
$\xi_n^+$ and $\xi_n^-$ are the analogous noise components as before in \eqref{eq:sg-alg}. 
Note that in \eqref{eq:spsa-grad}, the numerator is the same for each component of the gradient estimate vector, and only the denominator is changed, so that in each iteration, there are only two distinct values among the gradient estimate components. One can also rescale the gradient estimate if needed by generalizing the scalar $\delta_n$ to a direction-dependent vector. 

Like all gradient estimators based on finite differences, 
the SPSA gradient estimator is generally biased, but the bias can be controlled using the perturbation constant $\delta_n$. 
In the following, 
we present a result that establishes that the bias is only of order $O(\delta_n^2)$, hence asymptotically unbiased.  
For this result, we make the following assumptions:
\diff{\begin{assumption}
\label{ass:spsa-noise} Let $\eta_n^{\pm} = \hat{h}(\theta_n \pm \delta_n \Delta(n), \xi_n^{\pm}) - h(\theta_n)$.
Let $\F_n = \sigma(\theta_m, m<n)$ denote the underlying $\sigma$-field.
 For all $n\ge 1$, the $\eta_n^\pm$ satisfy 
	\begin{align}
		\E[\eta_n^+-\eta_n^- |\, \F_n] &= 0, \text{~~ and ~~}
		\E [ (\eta_n^{+} - \eta_n^-)^{2} |\, \F_n] \le \sigma^2 <\infty\,.
		\label{eq:noiseass}
	\end{align}
\end{assumption}

\begin{assumption}
\label{ass:spsa-smoothf}   The function $h$ is three times continuously differentiable, with $\left|\nabla^3_{i_1 i_2 i_3} h(\theta) \right| < \tilde B < \infty$, for $i_1, i_2, i_3=1,\ldots, d,\ \theta\in \R^d$, for some positive constant $\tilde B$. 
Furthermore, the function estimator $\hat{h}$ satisfies
$\E [ \hat{h}(\theta_n \pm \delta_n \Delta(n))^{2}] \le B <\infty\,~n\ge 1$, for some positive constant $B$.
\end{assumption}
}

\boxed{
\vspace*{-8pt}
	\begin{proposition}
		\label{prop:grad-2pt-c3}
 Assume \ref{ass:spsa-noise}--\ref{ass:spsa-smoothf}. Then the SPSA gradient estimator defined by \eqref{eq:spsa-grad} satisfies
		\begin{align*}
			&\left|\EE{\widehat\nabla_i h(\theta_n)} \!-\! \nabla_i h(\theta_n) \right| \le C_1\delta_n^2,  \textrm { for } i=1,\ldots, d, \textrm{ and }\\ 
		&\EE{\norm{ \widehat\nabla h(\theta_n) -  \EE{\widehat\nabla h(\theta_n)\mid \F_n} }^2} \le \frac{C_2}{\delta_n^2},
		\end{align*}
		for some positive constants $C_1$ and $C_2$.
\end{proposition}}

\medskip
\begin{proof}
	Using a Taylor series expansion of $h$ around $\theta_n$, we obtain
	\begin{align}
		h(\theta_n \pm \delta_n \Delta(n)) &=
		h(\theta_n)
		\pm\delta_n\,  \Delta(n)\tr\,\nabla h(\theta_n)
		+ \frac{\delta_n^2}{2}\, \Delta(n)\tr \nabla^2 h(\theta_n) \Delta(n)\nonumber\\
		&\quad\pm  \frac{\delta_n^3}{6} \nabla^3 h(\tilde  \theta_n^{\pm})(\Delta(n) \otimes \Delta(n) \otimes \Delta(n)),\label{eq:taylor}
	\end{align}
	where $\otimes$ denotes the Kronecker product, and $\tilde \theta_n^+$ (resp. $\tilde \theta_n^-$) is on the line segment between $\theta_n$ and $(\theta_n + \delta_n \Delta(n))$ (resp. $(\theta_n - \delta_n \Delta(n))$).
	
	Thus, we have
	\begin{align}
		    & \EE{\left.\dfrac{h(\theta_n+\delta_n \Delta(n)) - h(\theta_n-\delta_n \Delta(n))}{2\delta_n \Delta_i(n)} \right| \F_n} \nonumber\\
			&= \mathbb{E}\bigg(\frac{\Delta(n)^{\tr} \nabla h(\theta_n)}{\Delta_i(n)}   \\
			&\qquad+
			\frac{\delta_n^2}{12 \Delta_i(n)}  (\left.\nabla^3 h(\tilde  \theta_n^+)+\nabla^3 h(\tilde  \theta_n^-)(\Delta(n) \otimes \Delta(n) \otimes \Delta(n))\right| \F_n\bigg)\nonumber\\						&= \nabla_i h(\theta_n)    +
\EE{\frac{\delta_n^2}{12 \Delta_i(n)}  (\nabla^3 h(\tilde  \theta_n^+)+\nabla^3 h(\tilde  \theta_n^-)(\Delta(n) \otimes \Delta(n) \otimes \Delta(n))\mid \F_n}.\nonumber
	\end{align}
    To arrive at the final inequality, we used 
    \[\EE{\left.\frac{\Delta(n)^{\tr} \nabla h(\theta_n)}{\Delta_i(n)} \right| \F_n}= 
    \nabla_i h(\theta_n) + \EE{\sum_{j=1,j\not=i}^{d} \frac{\Delta_j(n)}{\Delta_i(n)}\nabla_j h(\theta_n)} = \nabla_i h(\theta_n),\]
    since $\Delta(n)$ is a vector of i.i.d. symmetric $\pm 1$-valued Bernoulli r.v.s.
    
	Using  \\
	(i) $\E[\widehat\nabla_i h(\theta_n)] =  \E\left[ \dfrac{h(\theta_n+\delta_n \Delta(n))  -h(\theta_n-\delta_n \Delta(n))}{2\delta_n \Delta_i(n)} \right]$;\\
	 (ii) $|\nabla^3 h(\tilde \theta_n^\pm)| < \tilde B$; and \\
	 (iii) $\EE{\frac{1}{\Delta_i(n)}|\nabla^3 h(\bar\theta_n) (\Delta(n) \otimes \Delta(n) \otimes \Delta(n))|}\le
	\tilde B d^3$ for any $\bar\theta_n$, \\
	we have
	\begin{align*}
		\left| \EE{ \widehat\nabla_i h(\theta_n) } - \nabla h(\theta_n) \right|
		\le C_1\, \delta_n^2 \,, \textrm{ where }C_1 = \frac{\tilde B d^3}{6}.
	\end{align*}
	Next, we prove the second claim concerning the variance of $\widehat\nabla h(\theta_n) $.
	Notice that
	\begin{align}
		&\E \left|\widehat\nabla_i h(\theta_n) \right|^2 \\
		&=  \E\left[ \left(\dfrac{\eta_n^+ - \eta_n^-}{2\delta_n\Delta_i(n)}\right)^2  \right.\\
        &\left.\qquad\qquad		+ 2 \left(\dfrac{\eta_n^+ - \eta_n^-}{2\delta_n\Delta_i(n)}\right) \left(\dfrac{h(\theta_n+\delta_n \Delta(n)) - h(\theta_n-\delta_n \Delta(n))}{2\delta_n \Delta_i(n)}\right)\right.\nonumber\\
		&\left.\qquad\qquad+ \left( \dfrac{h(\theta_n+\delta_n \Delta(n)) - h(\theta_n-\delta_n \Delta(n))}{2\delta_n \Delta_i(n)} \right)^2 \right] \nonumber \\
		&=  \E\left(  \left(\dfrac{\eta_n^+ - \eta_n^-}{2\delta_n }\right)^2\right)
		+  \E \left(\left( \dfrac{h(\theta_n+\delta_n \Delta(n)) - h(\theta_n-\delta_n \Delta(n))}{2\delta_n} \right)^2\right)~~ \\ 
		& \le  \frac{C_2}{\delta_n^2}\,  
	\end{align}
	where $C_2 = \left( \sigma^2+2 B^2\right)/4$.
	The last equality above 
	uses $\Delta_i(n)^2=1$ and  \ref{ass:spsa-noise}, while the final inequality 
	uses \ref{ass:spsa-noise} and \ref{ass:spsa-smoothf}.
Then the second claim in the proposition follows by using
$\E\left\| \widehat\nabla h(\theta_n)  - \E \widehat\nabla h(\theta_n) \right\|^2
		\le 4 \E \left\|\widehat\nabla h(\theta_n) \right\|^2$ in conjunction with the inequality above. 
\end{proof}

\begin{remark}
A variant of \eqref{eq:spsa-grad} is to use a one-sided estimate, i.e., given sample observations at $\theta_n + \delta_n \Delta(n)$ and $\theta_n$,  
use the following gradient estimator:
\begin{align}
\widehat \nabla_i \hat{h}(\theta_n) =  \left(\frac{\hat{h}(\theta_n + \delta_n \Delta(n), \xi_n^+)
- \hat{h}(\theta_n, \xi_n)}{\delta_n
\Delta_{i}(n)} \right), \label{eq:spsa-grad-onesided}
\end{align}
where $\delta_n$ and $\Delta(n)$ are as defined earlier. 
For solving constrained optimization problems, one-sided estimates are efficient, since a sample observation at the unperturbed value of the underlying parameter is necessary for performing the dual ascent on the Lagrange multiplier. The overall SPSA-based policy gradient algorithm would estimate the necessary gradient, as well as the risk measure, using two sample observations corresponding to $\theta_n+ \delta_n \Delta(n)$ and $\theta_n$. On the other hand, using a balanced estimate, as defined in \eqref{eq:spsa-grad} would require an additional observation with the underlying parameter set to  $\theta_n- \delta_n \Delta(n)$.
\end{remark}

\diff{The following result establishes that the SPSA algorithm converges to a zero of the gradient, where the SPSA algorithm is defined by the gradient-based SA iteration in \eqref{eq:sg-alg} driven by biased (but asymptotically unbiased) stochastic gradient estimates given by \eqref{eq:spsa-grad}.
\boxed{
\vspace*{-10pt}
\begin{theorem}
	\label{thm:spsa-conv}
	Assume \ref{ass:spsa-noise}--\ref{ass:spsa-smoothf}.	
	Let $\Theta^* \triangleq\{ \theta \mid \nabla h(\theta) = 0 \}$.
	Then $\theta_n$ governed by  \eqref{eq:sg-alg} using step sizes satisfying \ref{ass:stepsize-squaresummable}, with $\widehat \nabla h(\theta) \triangleq [\widehat \nabla_i h(\theta)]$ defined by \eqref{eq:spsa-grad}, converges a.s. to a zero of the gradient of $h$, i.e., 
	\[\theta_n \rightarrow \Theta^* \text{ a.s. as } n\rightarrow \infty.\]
\end{theorem}}
The proof involves an application of Theorem \ref{thm:sa-conv}, and we omit the details. A policy gradient algorithm with SPSA-based gradient estimates is analyzed in Section \ref{sec:convergence-unconstrained}, and the proof of Theorem \ref{thm:spsa-conv} would go through using arguments similar to those employed in the proof of Theorem \ref{thm:theta-convergence-unconstrained} in Section \ref{sec:convergence-unconstrained}.}

\section{Direct single-run gradient estimation using the likelihood ratio (LR) method}
\label{sec:lr}

When the system is a complete black box, SPSA is an effective way to carry out gradient-based policy optimization. 
However, in many settings, more is known about the system, and more efficient {\it direct} gradient estimation techniques may be applicable, where ``direct'' means that the gradient estimator is unbiased (as opposed to asymptotically unbiased when finite difference methods such as SPSA are used).  
The main approaches are perturbation analysis, the likelihood ratio method (also known as the score function method), and weak derivatives (also known as measure-valued differentiation). 
\diff{
Here, we consider only the likelihood ratio (LR) method, which is a single-run method for gradients where $\theta$ parameterizes the input distribution(s) of the system. 
``Single-run'' means that a gradient estimate can be obtained using a single sample path (or simulation) of the system, in contrast to SPSA, which requires multiple sample paths to estimate the gradient. 
To motivate the more general case, we first illustrate the idea using a single (discrete-valued) random variable example: 
$$
\mathbb{E}[X] = \sum_x x \mathbb{P}_{\theta}(X=x)  = \sum_x x p_\theta(x),
$$
where $p_\theta$ denotes the probability mass function of $X$. 
Differentiating with respect to $\theta$ (assuming the differentiation operator can be brought inside the summation), 
\[
\frac{d\mathbb{E}[X]}{d\theta} 
= \sum_x x \frac{d\mathbb{P}_{\theta}(X=x)}{d\theta}  
= \sum_x x \frac{d \ln p_\theta(x)}{d\theta} p_\theta(x) 
= \mathbb{E} \left[ X \frac{d \ln p_\theta(X)}{d\theta} \right],
\]
and thus the LR derivative estimator for this simple example is given by 
\[
X \frac{d \ln p_\theta(X)}{d\theta}.
\]
Now consider a Markov chain $\{X_n\}$ with a single recurrent state $0$, transient states $1,\ldots,r$, 
and (one-step) transition probability matrix $P(\theta):=[p_{ij}(\theta)]_{i,j=0}^{r}$, where $p_{ij}(\theta)$ denotes the (one-step transition) probability of going from state $X_n = i$ to $X_{n+1} = j$ and is parameterized by $\theta$. 
Let $\tau$ denote the first passage time to the recurrent state $0$ and 
$X:=(X_0, \ldots, X_{\tau-1})\tr$ denote the corresponding sequence of states (sample path).
Assuming $\theta$ occurs only in the transition probabilities,
an unbiased single-run sample path LR gradient estimator for 
$\nabla h(\theta)$ is given by
\boxed{
\vspace*{-10pt}
\[
\widehat{\nabla} h(\theta) = 
\hat{h}(X)  \nabla \ln p_{X_0 X_1 \cdots X_{\tau}}(\theta) = 
\hat{h}(X) \sum\limits_{m=0}^{\tau-1} \dfrac{\nabla p_{X_m X_{m+1}}(\theta)}{p_{X_m X_{m+1}}(\theta)},
\]
}
where the single random variable is replaced by a function of the Markov chain states visited and the single probability mass function is replaced by a joint distribution $p_{X_0 X_1 \cdots X_{\tau}}$ that is the product of the individual one-step transition probabilities $p_{ij}$. 

It can be shown under mild conditions on the transition probabilities that the LR gradient estimator is unbiased, 
i.e., 
\begin{align}
\E[\widehat{\nabla} h(\theta)] = \nabla h(\theta),    \label{eq:lr-gd}
\end{align} 
which follows by invoking the dominated convergence theorem to interchange expectation and differentiation operators. 
Unbiasedness is a desirable property, because it generally leads to a faster convergence rate for gradient-based algorithms, e.g., an asymptotic rate of $1/\sqrt{n}$ in (\ref{sec:sa}) rather than $1/n^{1/3}$ for the typical finite-difference gradient estimate (see, e.g., Theorem \ref{thm:biased_sp}).
However, it should also be noted that if $\theta$ is a common parameter that appears in all of the probabilities, then this LR estimator will have the undesirable property of its variance increasing linearly with the sample path length.

\begin{remark}
Using the LR gradient estimator in the stochastic gradient algorithm \eqref{eq:sg-alg} would ensure that the resulting algorithm converges to the set of stationary points of the objective, i.e., Theorem \ref{thm:spsa-conv} holds for the LR case, as well.
\end{remark}
}

\section{Bibliographic remarks}
\diff{Stochastic approximation has a long history, starting with the seminal paper of 
\textcite{robbins1951stochastic}, where the iterative algorithm \eqref{eq:sa} is used to solve a stochastic root-finding problem. 
For a proof of the asymptotic convergence claim in Theorem \ref{thm:sa-conv} for a Robbins-Monro stochastic approximation scheme, the reader is referred to Theorem 2.3.1 of \textcite{kushner-clark}, which is what is referred to in the SA literature as the Kushner-Clark Lemma. For a rigorous introduction to the ODE approach for analyzing stochastic approximation algorithms, the reader is referred to \textcite{borkar2008stochastic}. The claim in Theorem \ref{thm:contract-sto-approx} is a special case of the result in Theorem \ref{thm:sa-conv} for a general stochastic approximation scheme, and the interested reader is referred to Chapters 4 and 5 of \textcite{BertsekasT96} for the proof as well as RL applications. The convergence result for the projected stochastic approximation scheme in Theorem \ref{thm:kc} is the projected form of the Kushner-Clark Lemma, which is Theorem 5.3.1 of \textcite{kushner-clark}. The noise conditions referred to in Section \ref{sec:sa} for avoiding traps (e.g., local maxima, saddle points) are given in \textcite{pemantle1990nonconvergence}; see also Section 4.3 of \textcite{borkar2008stochastic}. 
The stability result presented in Theorem \ref{thm:borkarmeyn} is the Borkar-Meyn theorem; see Theorems 2.1-2.2(i) of \textcite{borkar2000ode} and also Chapter 3 of \textcite{borkar2008stochastic}. A popular idea that improves the convergence guarantees for general stochastic approximation algorithms is iterate averaging, proposed independently by Polyak \citep{polyak1992acceleration} and Ruppert \citep{ruppert1991stochastic}. The idea behind this scheme is to use larger step-sizes of $\Theta(1/n^\varsigma)$ for some $\varsigma \in (1/2,1)$ to perform the update iteration \eqref{eq:sa-general}, and then use the averaged iterate $\bar \theta_{n+1} = \frac{1}{n} \sum_{k=1}^n \theta_k$ instead of the last iterate $\theta_n$ for providing the convergence guarantees. For the special class of stochastic gradient algorithms that solve a minimization problem, popular approaches for improving the convergence rate is to incorporate second-order information and/or variance reduction, see \textcite{bottou2018optimization,gower2020variance} for recent surveys on these topics. 

The method of temporal differences for policy evaluation was proposed by \textcite{sutton1988learning}. For an analysis of TD learning, the reader is referred to either Chapters 5 and 6 of \textcite{BertsekasT96} or Chapter 6 of \textcite{sutton2018reinforcement}. 
An analysis of the extension of TD to incorporate linear function function approximation can be found in \textcite{tsitsiklis1997analysis}.  The stability of the TD iterate can also be inferred using the Borkar-Meyn theorem, referred above.
Non-asymptotic analysis of TD with linear function approximation has received a lot of attention recently, and a few representative works are    \textcite{prashanth2021td,dalal2018finite,bhandari2018finite,srikant19td}.
}
Average-cost TD learning algorithm with full state representations and its convergence analysis can be found in \textcite{konda1999actor}, in particular, the faster timescale recursion in Algorithm 4 there. Also related is the average-cost Q-learning algorithm with full state representations, which has been proposed/analyzed by \textcite{abounadi2001learning}. For general conditions to infer stability of TD/Q-learning algorithms in an average-cost MDP, see \textcite{abounadi2002stochastic}. Finally, for the extension of average-cost TD algorithm to incorporate feature-based representations and linear function approximation, see \textcite{tsitsiklis1999average}. 

\textcite{Kiefer1952} extended the Robbins-Monro algorithm for root finding to optimizing a function through gradient search. The Kiefer-Wolfowitz algorithm requires $2d$ function evaluations/estimates per iteration to estimate the gradient, whereas the SPSA algorithm proposed by \textcite{Spall1992} estimates the gradient using only two function evaluations/estimates per iteration, regardless of the problem dimension $d$. For a closely related random directions SA algorithm, 
see \textcite[pp.~58-60]{kushner-clark} and \textcite{prashanth2017RDSA}, and 
for a detailed introduction to such gradient estimation methods, see \textcite{Bhatnagar13SR}. The convergence analysis of the SPSA algorithm presented here, in particular, the main result in Theorem \ref{thm:spsa-conv} can be inferred by an application of the Kushner-Clark Lemma \citep{kushner-clark}.  


The likelihood ratio (LR) method for gradient estimation, also known as the score function method, has its roots in a 1968 Russian paper \citep{AlSySh68}, but the technique appeared to be unknown to the rest of the world until it was ``rediscovered'' by several different researchers in parallel
\citep{Rubinstein1989,glynn1987likelihood,Reiman1989}; see \textcite{Fu2006b,Fu2015} for an introduction and overview.
For the mild conditions justifying the unbiasedness of \eqref{eq:lr-gd}, the reader is referred to Section VII.3 of \citet{Asmussen2007}.
The weak derivatives (WD) method, also known as measure-valued differentiation \citep{Pflug1989,Pflug1996,Heidergott2000}, could also be applied in the Markov chain context, but it would generally require multiple sample paths of the chain or appropriate randomization; however, unlike finite-difference-based methods, including the gradient estimators used in SPSA, LR and WD gradient estimators are generally unbiased (as opposed to asymptotically unbiased). 
    
%
%
\chapter{Policy Gradient Templates for Risk-sensitive RL}
\label{sec:rpg-template}
\diff{In this chapter, we present and analyze template algorithms for two risk-sensitive RL settings. The first setting, described in Section \ref{sec:template-unconstrained}, incorporates the risk measure directly in the objective function, whereas the second setting, described in Section \ref{sec:template-constrained}, considers the risk measure in a constrained formulation, with the usual (risk-neutral) cost function as the objective.
Sections \ref{sec:convergence-unconstrained} and \ref{sec:convergence-constrained} analyze the convergence of the template algorithms for the risk-as-objective and risk-as-constraint settings, respectively. 

Recall that  $\left\{\mu^\theta(\cdot|x),x\in\X,\theta\in\Theta\subseteq\R^{d}\right\}$
is a parameterized set of randomized policies, 
and the goal is to find a policy that optimizes a risk measure as an objective, or optimizes the usual risk-neutral objective (cost/reward) function while satisfying a risk constraint. The policy parameterization is assumed to be smooth (cf. Sections \ref{sec:convergence-unconstrained} and \ref{sec:convergence-constrained}), and a commonly used class of distributions that ensures a smooth parameterization is the `Boltzmann family' taking the form
\begin{equation}
	\mu^\theta(a|x) = \frac{\exp(\theta^{\tr} \phi_{x,a})}{\sum_{a' \in {\A(x)}} \exp(\theta^{\tr} \phi_{x,a'})},\ \forall x \in \X\;,\;\forall a \in \A(x),
	\label{eq:boltzmann}
\end{equation}
where $\theta$ is constrained to be in a convex and compact set $\Theta \subset \R^d$ and $\{ \phi_{x,a} \}$ is a set of state-action features.

The proposed policy gradient algorithms attempt to find a `good enough' policy that optimizes a risk-sensitive objective by performing gradient descent in the policy space, where the concept of a `good enough' policy will be made precise in the convergence analysis of these algorithms.

For notational convenience, we define
\[
J(\theta) \triangleq J_{\mu^{\theta}}, ~~D(\theta) \triangleq D_{\mu^{\theta}}, ~~G(\theta) \triangleq G_{\mu^{\theta}}
\]
in MDP contexts where the objective is independent of the initial state, e.g., average-cost MDPs. Analogously, for the discounted-cost MDPs and SSP settings, which depend on the initial state, we define
\[
J(\theta,x) \triangleq J_{\mu^{\theta}}(x), D(\theta,x) \triangleq D_{\mu^{\theta}}(x), G(\theta,x) \triangleq G_{\mu^{\theta}}(x),~~\forall x\in \X.
\]

}

\section{Template for the  setting with risk as objective}
\label{sec:template-unconstrained}

This setting considers the following optimization problem:
\begin{align}
	\min_{\theta\in\Theta} G(\theta),\label{eq:pb1a}
\end{align}
where 
$G$ involves one of the risk measures presented in Chapter \ref{sec:risk-measures}. 
Solving the problem (\ref{eq:pb1a}) via a policy gradient algorithm invokes the following stochastic approximation (SA) iterative update:
\boxed{
\vspace*{-10pt}
	\begin{align}
		\label{eq:theta_descent_unconstrained}
		\theta_{n+1} &= \Gamma\left[\theta_n - \zeta(n)  \widehat \nabla G(\theta_n)\right],
\end{align}} 
where $\{\zeta(n)\}$ is a step-size sequence,  $ \widehat \nabla G(\theta_n)$ is an estimate of $\nabla G(\theta_n)$, and $\Gamma$ is a projection operator that keeps the iterate $\theta_n$ bounded within a convex and compact set $\Theta$. \diff{A simple choice for the projected region is $\Theta:= \prod_{i=1}^{d} [\theta^{(i)}_{\min},\theta^{(i)}_{\max}]$, which leads to the following simple implementation for the projection operator: For any $\theta \in \R^d$,  $\Gamma(\theta) = (\Gamma^{(1)}(\theta^{(1)}),\ldots, \Gamma^{(d)}(\theta^{(d)}))^T$, with $\Gamma^{(i)}(\theta^{(i)}) := \min(\max(\theta^{(i)}_{\min},\theta^{(i)}),\theta^{(i)}_{\max})$.}

\diff{The convergence analysis of the policy update algorithm in \eqref{eq:theta_descent_unconstrained} is presented in Section \ref{sec:convergence-unconstrained}. In particular, we provide both asymptotic and non-asymptotic convergence guarantees there. }

\section{Template for the setting with risk as constraint}
\label{sec:template-constrained}

Recall that the setting incorporating risk as constraint is given by the following problem:
\begin{align}
	\min_{\theta\in\Theta} J(\theta)\quad\quad \text{subject to} \quad\quad G(\theta)\leq \kappa,\label{eq:pb1}
\end{align}
where  $J$ is the usual risk-neutral MDP objective, while the constraint $G$ involves one of the risk measures presented in Chapter \ref{sec:risk-measures}. 
In an average-cost formulation, the objective/constraint do not depend on the initial state, whereas they do in total-cost formulations such as SSP and discounted problems. In either case, the template for solving the problem remains the same, and to keep the presentation simple, we have chosen to have  the policy parameter only in $J$ and $G$. In the special cases of Section \ref{sec:risk-algos}, we shall include the initial state as necessary.

If there is a policy in $\Theta$ that satisfies the constraint in \eqref{eq:pb1}, then it can be inferred that there exists an optimal policy.

Using the Lagrangian approach, we consider the following relaxed MDP problem:
\begin{equation}
	\max_\lambda\min_\theta\left(L(\theta,\lambda) \triangleq J(\theta)+\lambda\big(G(\theta)-\kappa\big)\right),
\end{equation}
where $\lambda$ is the Lagrange multiplier.
The goal here is to find the saddle point of  $L(\theta,\lambda)$, i.e.,~a point  $(\theta^*,\lambda^*)$ that satisfies  
$$L(\theta, \lambda^*) \ge L(\theta^*, \lambda^*) \ge L(\theta^*, \lambda),\forall\theta\in\Theta,\forall \lambda>0.$$ For a standard convex optimization problem where the objective $L(\theta,\lambda)$ is convex in $\theta$ and concave in $\lambda$,  one can ensure the existence of a unique saddle point under mild regularity conditions. Further, convergence to this point can be achieved by descending in  $\theta$ and ascending in  $\lambda$ using $\nabla_\theta L(\theta,\lambda)$ and $\nabla_\lambda L(\theta,\lambda)$, respectively.

However, in the risk-sensitive RL setting, the Lagrangian $L(\theta,\lambda)$ is not necessarily convex in $\theta$, which implies there may not be an unique saddle point.
Hence, performing primal descent and dual ascent, one can only get to a local saddle point, i.e., a point $(\theta^*, \lambda^*)$ that is a local minima w.r.t. $\theta$, and local maxima w.r.t $\lambda$ of the Lagrangian.  
\diff{ The problem is further complicated by the fact in a {\em typical RL} setting, closed-form evaluation of the Lagrangian for any given policy parameter $\theta$ and Lagrange multiplier $\lambda$ is not feasible. Instead, one can run sample trajectories after fixing the parameters $\theta$ and $\lambda$, and obtain estimates of the Lagrangian corresponding to the policy parameter.}

For the purpose of finding an optimal risk-sensitive policy, a standard procedure would update the policy parameter $\theta$ and Lagrange multiplier $\lambda$ in two nested loops:
an inner loop that descends in $\theta$ using the gradient of the Lagrangian $L(\theta,\lambda)$ w.r.t. $\theta$, and
an outer loop that ascends in $\lambda$ using the gradient of the Lagrangian $L(\theta,\lambda)$ w.r.t. $\lambda$.

We operate in a setting where we only observe simulated costs of the underlying MDP. Thus, it is required to estimate both $J$ and $G$ for a given $\theta$ and then use these estimates to compute an estimate of the gradient of the Lagrangian w.r.t. $\theta$ and $\lambda$. 
The gradient $\nabla_\lambda L(\theta,\lambda)$ has a particularly simple form of $(G(\theta)-\kappa)$, suggesting that a sample of the risk measure can be used to perform the dual ascent for Lagrange multiplier $\lambda$. On the other hand, the policy gradient $\nabla_\theta L(\theta,\lambda) = \nabla J(\theta) + \lambda \nabla G(\theta)$ is usually complicated and does not lend itself to stochastic programming techniques in a straightforward fashion. 
We shall address the topic of gradient estimation in the next section, but for presenting the template of the risk-sensitive policy gradient algorithm, we assume the availability of estimators $\widehat \nabla J(\theta)$, $\widehat \nabla G(\theta)$ and $\widehat G(\theta)$ of $\nabla J(\theta)$, $\nabla G(\theta)$ and $G(\theta)$, respectively. Then, using two-timescale stochastic approximation,  
the inner and outer loops mentioned above can run in parallel, as follows (please refer to Figure \ref{fig:algorithm-flow}):
\boxed{
	\vspace*{-10pt}
	\begin{align}
		\lambda_{n+1} &= \left[\lambda_n + \zeta_1(n) \left(\widehat G(\theta_n) - \kappa\right)\right]^+, \label{eq:lambda-ascent}  \\
		\theta_{n+1} &= \Gamma\left[\theta_n - \zeta_2(n)  \left(\widehat\nabla J(\theta_n) + \lambda_n \widehat \nabla G(\theta_n)\right)\right],          
		\label{eq:theta_descent_det} 
\end{align}}
where $[x]^+=\max(0,x)$ for any real $x$,
$\Gamma$ is a projection operator that keeps the iterate $\theta_n$ stable by projecting  onto a compact and convex set $\Theta$, as in the setting considered in Section \ref{sec:template-unconstrained}; and
$\{\zeta_1(n), \zeta_2(n)\}$ are step-size sequences selected such that the $\theta$ update is on the faster timescale, and the $\lambda$ update is on the slower timescale. 

\begin{figure*}
	\centering
	\tikzstyle{block} = [draw, fill=white, rectangle,
	minimum height=5em, minimum width=6em]
	\tikzstyle{sum} = [draw, fill=white, circle, node distance=1cm]
	\tikzstyle{input} = [coordinate]
	\tikzstyle{output} = [coordinate]
	\tikzstyle{pinstyle} = [pin edge={to-,thin,black}]
	\scalebox{0.5}{\begin{tikzpicture}[auto, node distance=2cm,>=latex']
			\node (theta) at (-6,0) {$\bm{\theta_n,\lambda_n}$};
			\node [block,minimum height=7em, minimum width=7em, fill=blue!20,right=2cm of theta,label=below:{\color{bleu2}\bf Simulation},align=center] (sample) {\textbf{Using policy $\mu^{\theta_n}$, }\\[1ex] \textbf{simulate the }\\[1ex]\textbf{underlying MDP}}; 
			\node [block, fill=green!20,above right=3cm of sample,label=below:{\color{darkgreen!90}\bf Policy Gradient},align=center] (obj) {\textbf{Estimate} $\bm{\nabla J(\theta)}$};
			\node [block, fill=green!20,right=2cm of sample,label=below:{\color{darkgreen!90}\bf Risk Estimation},align=center] (estcvar) {\textbf{Estimate} $\bm{G(\theta)}$};
			\node [block, fill=green!20,below right=3cm of sample,label=below:{\color{darkgreen!90}\bf Risk Gradient},align=center] (cvar) {\textbf{Estimate} $\bm{\nabla G(\theta)}$};
			\node [block, fill=red!20,right=8cm of sample, minimum height=8em,label=below:{\color{violet!90}\bf Policy Update},align=center] (update) {\textbf{Update} $\bm{\theta_n}$ \textbf{using} \eqref{eq:theta_descent_det} \\[1ex] \textbf{Update } $\bm{\lambda_n}$ \textbf{using \eqref{eq:lambda-ascent}}};
			\node [right=2cm of update] (end) {$\bm{\theta_{n+1},\lambda_{n+1}}$};
			\draw [thick,-triangle 45] (theta) --  (sample);
			\draw [thick,-triangle 45] (sample) -- (obj);
			\draw [thick,-triangle 45] (sample) -- (estcvar);
			\draw [thick,-triangle 45] (sample) -- (cvar);
			\draw [thick,-triangle 45] (obj) -- (update);
			\draw [thick,-triangle 45] (estcvar) -- (update);
			\draw [thick,-triangle 45] (cvar) -- (update);
			\draw [thick,-triangle 45] (estcvar) -- (cvar);
			\draw [thick,-triangle 45] (update) -- (end);
	\end{tikzpicture}}
	\caption{Overall flow of risk-sensitive policy gradient algorithm.}
	\label{fig:algorithm-flow}
\end{figure*}

The template for a risk-sensitive policy gradient algorithm would involve the following components:
\begin{enumerate} 
	\item A two-timescale update rule for the policy parameter $\theta$ and the Lagrange multiplier $\lambda$;
	\item Estimates of the objective  $J(\cdot)$  and the risk measure $G(\cdot)$, which can be obtained by sampling from the underlying MDP with the current policy parameter, and then using a suitable estimation scheme, usually based on stochastic approximation (in RL, this would be equivalent to some form of TD-learning), or based on Monte Carlo averaging; the estimate of the objective would feed into estimating the policy gradient (see step below), while the estimate of the risk measure is necessary for the gradient of the Lagrangian, as well as for the dual ascent procedure; 
	\item Estimates of the gradients \diff{$\nabla J(\cdot)$} and $\nabla G(\cdot)$ for primal descent, which may be challenging to obtain if the underlying risk measure has no structure that can be exploited in an MDP framework.
\end{enumerate}

The convergence analysis of the two-timescale SA algorithm given by 
\eqref{eq:lambda-ascent} -- \eqref{eq:theta_descent_det} is presented in Section \ref{sec:convergence-constrained}.

\section{Convergence analysis in the setting with risk as objective}
\label{sec:convergence-unconstrained}
\diff{
The following assumptions are used for the convergence analysis of the SA recursive parameter update given by \eqref{eq:theta_descent_unconstrained} in Section \ref{sec:template-unconstrained}.
First, let $\F_n = \sigma\left(\theta_m, m<n\right)$ denote the underlying $\sigma$-field.

\begin{assumption}
\label{ass:smooth-policies}
The risk measure $G$ is a continuously differentiable function of the policy parameter $\theta$.
\end{assumption}
\begin{assumption}
\label{ass:unbiased-gradG}	
For all $n$, the gradient estimator $\widehat\nabla G(\theta_n)$ satisfies
$$ 
\E\left(\widehat\nabla G(\theta_n) \mid \F_n\right) = \nabla G(\theta_n), 
	\textrm{ and } \E\left\| \widehat\nabla G(\theta_n) - \nabla G(\theta_n) \right\|^2 \le \sigma^2 < \infty.
$$
\end{assumption}
\begin{assumption}
		\label{ass:biased-gradG}
For all $n$, the gradient estimator  $\widehat\nabla G(\theta_n)$ depends on a parameter $\delta_n>0$ and satisfies
$$
\left\| \E\left(\widehat\nabla G(\theta_n) \mid \F_n\right) - \nabla G(\theta_n) \right\| = C_1\delta_n^2,
	\textrm{ and }\E\left\| \widehat\nabla G(\theta_n) \right\|^2< \frac{C_2}{\delta_n^2},
$$
where 
$C_1$ and $C_2$ are dimension-dependent constants.
\end{assumption}
\begin{assumption}
		\label{ass:stepsize-onetimescale}
The step-size sequence $\{\zeta(n)\}$ satisfies
\begin{align*}
	&\sum\limits_n \zeta(n)  = \infty,
	\textrm{ and ~} \sum\limits_n \zeta(n)^2\;<\infty.
\end{align*}
\end{assumption}
\begin{assumption}
		\label{ass:stepsize-onetimescale-variant}
$\delta_n \rightarrow 0 \textrm{ as } n \rightarrow \infty$, and 
the step-size sequence $\{\zeta(n)\}$ satisfies
\begin{align*}
	&  \sum\limits_n \zeta(n) = \infty, \sum\limits_n \left[\frac{\zeta(n)}{\delta_n}\right]^2<\infty.
\end{align*}
\end{assumption}}
	
\diff{We now discuss the assumptions.
	Assumption \ref{ass:smooth-policies} imposes a smoothness requirement on the risk measure $G$, when viewed as a function of the policy parameter $\theta$. A similar requirement, i.e., the value function $J$ is smooth in $\theta$, holds when the underlying policy parameterization is smooth, i.e., $\mu^\theta$ is a continuously differentiable function of $\theta$.  While a smooth policy parameterization ensures $J$ is smooth, one cannot infer the same for an abstract risk measure $G$. As an example, one could consider the VaR risk measure. Hence, the second part of \ref{ass:smooth-policies} explicitly imposes the smoothness condition on $G$, and this condition together with the fact that $J$ is smooth is necessary for the ordinary differential equation (ODE) underlying the $\theta$-recursion to be well-posed.
Assumption \ref{ass:unbiased-gradG} requires that the risk gradient estimator be unbiased with finite variance, which is a standard assumption in the analysis of stochastic gradient schemes, and is generally satisfied using likelihood ratio-based gradient estimators. 
Assumption \ref{ass:biased-gradG} is a relaxed variant of \ref{ass:unbiased-gradG}, where the gradient estimators may not be unbiased, as in the case of finite-difference estimators that depend on a difference (perturbation) parameter $\delta_n$, which can be used to control the bias-variance tradeoff --- lower $\delta_n$ results in lower gradient estimation bias but higher variance, and vice versa. The SPSA approach presented in Section \ref{sec:spsa}, as well as the general class of simultaneous perturbation schemes, meet the conditions in \ref{ass:biased-gradG}. For convergence, $\delta_n$ has to vanish asymptotically, but not too fast, as outlined in the second part of \ref{ass:stepsize-onetimescale-variant}.   
The conditions on the step-size sequence $\{\zeta_1(n)\}$ in \ref{ass:stepsize-onetimescale} are standard stochastic approximation requirements (see \ref{ass:stepsize-squaresummable}). 
Assumption \ref{ass:stepsize-onetimescale-variant} is a variant of \ref{ass:stepsize-onetimescale}, and has to be coupled with \ref{ass:biased-gradG}, in the sense that the gradient estimates have a $O(\delta_n^2)$ bias, but $\sum_n \left(\frac{\zeta_2(n)}{\delta_n}\right)^2 < \infty$. The latter requirement ensures that the effect of noise in gradient estimation can be (eventually) ignored, while ensuring asymptotic convergence. Such a requirement that couples the step-size sequence and finite-difference gradient estimator perturbation sequence arises when a biased gradient estimation scheme such as SPSA is employed. 

Consider the following ODE underlying the policy update in \eqref{eq:theta_descent_unconstrained}:
		\begin{align}
	\dot{\theta}(t) = \check{\Gamma}\left( -\nabla G(\theta(t))\right),\label{eq:theta-ode-unconstrained}
\end{align}
where $\check{\Gamma}(\cdot)$ is a projection operator  that ensures the evolution of $\theta$ via the ODE (\ref{eq:theta-ode-unconstrained}) stays within 
the set $\Theta$ and is defined as follows: For any bounded continuous function $f(\cdot)$,
\begin{align}
	\label{eq:Pi-bar-operator}
	\check{\Gamma}\big(f(\theta)\big) = \lim\limits_{\tau \rightarrow 0}
	\dfrac{\Gamma\big(\theta + \tau f(\theta)\big) - \theta}{\tau}.
\end{align}
The limit defined above may not exist and in that case, one can define $\check{\Gamma}(f(\theta))$ to be the set of all possible limit points. 
	Further, for $\theta$ in the interior of $\Theta$, $\check{\Gamma}(f(\theta)) = f(\theta)$, while for $\theta$ on the boundary of $\Theta$, $\check{\Gamma}(f(\theta))$ is the projection of $f(\theta)$ onto the tangent space of the boundary of $\Theta$ at $\theta$.

The main result establishing asymptotic convergence of the policy gradient algorithm \eqref{eq:theta_descent_unconstrained} is given below.
\boxed{
\vspace*{-10pt}
	\begin{theorem}
		\label{thm:theta-convergence-unconstrained}
		Assume that \ref{ass:smooth-policies}, (\ref{ass:unbiased-gradG} + \ref{ass:stepsize-onetimescale})  or (\ref{ass:biased-gradG} + \ref{ass:stepsize-onetimescale-variant}) hold. For $\theta_n$ governed by \eqref{eq:theta_descent_unconstrained}, 
		\[\theta_n \rightarrow \Z \textrm{ a.s. as } n \rightarrow \infty,\]
		where $\Z=\big\{\theta\in \Theta:\check\Gamma\big(\nabla G(\theta(t))\big)=0\big\}$ is the set of limit points of the ODE \eqref{eq:theta-ode-unconstrained}.
\end{theorem}}
\begin{proof}
	The proof involves an application of the Kushner-Clark lemma for projected stochastic approximation, provided in Section \ref{sec:projSA} as Theorem \ref{thm:kc}.

	We first rewrite the recursion \eqref{eq:theta_descent_det} as follows:
	\begin{align}
		\theta_{n+1} = &\Gamma \bigg( \theta_n + \zeta(n) \Big(\nabla G(\theta_n) + \xi_{n}  \Big)\bigg)\label{eq:theta-update-unconstrained-equiv},\\
		\textrm{where	}\xi_{n} =& \widehat\nabla G(\theta_n)  - \nabla G(\theta_n).\nonumber
	\end{align}
	Application of Theorem \ref{thm:kc} requires the conditions \ref{ass:LipschitzF}--\ref{ass:noise-sa} to hold, and we verify these conditions for the recursion in \eqref{eq:theta-update-unconstrained-equiv} in the case where we assume \ref{ass:smooth-policies}, \ref{ass:unbiased-gradG}, and \ref{ass:stepsize-onetimescale}. We shall later provide the deviations necessary to handle the case when \ref{ass:biased-gradG}, and \ref{ass:stepsize-onetimescale-variant} are used in place of \ref{ass:unbiased-gradG}, and \ref{ass:stepsize-onetimescale}, respectively.
	\begin{itemize}
		\item The smoothness condition in \ref{ass:smooth-policies} implies \ref{ass:LipschitzF}.
		\item Since \ref{ass:unbiased-gradG} implies an unbiased gradient estimator, the assumption \ref{ass:betaVanish} trivially holds.
		\item \ref{ass:stepsize-onetimescale} implies \ref{ass:stepsize-sa}.
		\item The verification of \ref{ass:noise-sa} requires the application of a martingale inequality attributed to Doob, which is given in \eqref{eq:doob}.
We apply this inequality in our setting to the martingale sequence $\{\sum_{n=k}^{l} \zeta(n) \xi_{n}\}_{l\ge k}$ to obtain
		\begin{align*}
		\lim_{k\rightarrow \infty}	\mathbb{P}\left( \sup_{l\geq k}   \left\|\sum_{n=k}^{l} \zeta(n) \xi_{n}\right\| \geq \epsilon \right) \le & \dfrac{1}{\epsilon^2}\lim_{k\rightarrow \infty} \sum_{n=k}^{\infty} \zeta(n)^2 \E\left\| \xi_{n}\right\|^2 \\
			\le & \dfrac{C_3}{\epsilon^2} \lim_{k\rightarrow \infty}\sum_{n=k}^{\infty} \zeta(n)^2 =0, \end{align*}
		where the final inequality uses $\E\left\| \xi_{n}\right\|^2\le C_3 < \infty$, which can be inferred from the second condition in \ref{ass:unbiased-gradG}, while the final equality follows from the square summability of the step-size $\zeta(n)$, which is assumed in \ref{ass:stepsize-onetimescale}.	
		Thus, \ref{ass:noise-sa} is satisfied.
		\item $\Z_\lambda$ is an asymptotically stable attractor for the ODE~\eqref{eq:theta-ode-unconstrained}, with $G(\theta)$ itself serving as a strict Lyapunov function. This can be inferred as follows:
		\begin{align*}
			\dfrac{d G(\theta)}{d t}  
			= \nabla G(\theta) \dot \theta
			= \nabla  G(\theta) \check\Gamma\big(-\nabla  G(\theta)\big) < 0,\; \forall \theta \notin \Z_\lambda.  
		\end{align*}
	\end{itemize}
	The claim now follows by an application of Theorem \ref{thm:kc}. 
		
	For the case when \ref{ass:biased-gradG} and \ref{ass:stepsize-onetimescale-variant} are used in place of \ref{ass:unbiased-gradG}, and \ref{ass:stepsize-onetimescale}, the proof again follows by verifying conditions \ref{ass:LipschitzF}--\ref{ass:noise-sa} of Theorem \ref{thm:kc}. Of these, \ref{ass:LipschitzF} and \ref{ass:stepsize-sa} hold as shown above. The conditions \ref{ass:betaVanish} and \ref{ass:noise-sa} require a few deviations from the proof above, and we provide the details below.
	We rewrite the $\theta$-recursion as follows:
	\begin{align*}
			\theta_{n+1} = &\Gamma \bigg( \theta_n + \zeta(n) \Big(\nabla G(\theta_n,\lambda) + \xi_{1,n} + \xi_{2,n}  \Big)\bigg),
		\end{align*}
	where
	\begin{align*}
			\xi_{1,n} =& \E\left(\widehat\nabla G(\theta_n)  \mid \F_n\right) - \nabla  G(\theta_n),\\
			\xi_{2,n} =& \widehat\nabla G(\theta_n) - \E\left(\widehat \nabla G(\theta_n) \mid \F_n\right).
		\end{align*}
	\begin{itemize}
			\item From \ref{ass:biased-gradG},  we have that $\xi_{1,n} = O(\delta_n^2)$. Using \ref{ass:stepsize-onetimescale-variant}, we have $\xi_{1,n}\rightarrow 0$ as $n\rightarrow \infty$. This verifies \ref{ass:betaVanish}.
			\item For verifying \ref{ass:noise-sa}, first note that $	\E\left\| \xi_{2,n}\right\|^2 \le C_4 < \infty$ using the last condition in \ref{ass:biased-gradJG}. As before, applying Doob's martingale inequality and using square-summability of step-sizes $\{\zeta(n)\}$, we obtain
			\begin{align*}
					\lim_{k\rightarrow \infty} \Prob{ \sup_{l\geq k}   \left\|\sum_{n=k}^{l} \zeta(n) \xi_{2,n}\right\| \geq \epsilon} \le & \dfrac{1}{\epsilon^2} \lim_{k\rightarrow \infty}\sum_{n=k}^{\infty} \zeta(n)^2 \E\left\| \xi_{2,n}\right\|^2 \\
					\le & \dfrac{C_4}{\epsilon^2} \lim_{k\rightarrow \infty}\sum_{n=k}^{\infty} \frac{\zeta(n)^2}{\delta_n^2} =0. \end{align*}
			Thus \ref{ass:noise-sa} holds.
		\end{itemize}
	Hence, the claim follows for the case with assumptions \ref{ass:smooth-policies}, \ref{ass:biased-gradG}, and \ref{ass:stepsize-onetimescale-variant}.	
\end{proof}}

\diff{Interestingly, one can derive a non-asymptotic bound for the policy gradient algorithm in a setting where optimizing a risk measure is the objective. The asymptotic result in the theorem above guarantees convergence to a point $\theta^*$ where the gradient of the objective vanishes. The non-asymptotic bound that we present next establishes a non-asymptotic bound on the norm of the gradient $\E\|\nabla G(\theta_R)\|^2$, where $\theta_R$ is a point picked uniformly at random from the set $\{\theta_1,\ldots,\theta_n\}$. Notice the resemblance of this scheme of picking a random iterate uniformly to the well-known Polyak-Ruppert averaging scheme for stochastic approximation. In the latter, the averaging is explicit, while the former achieves the same in expectation. However, we note that the non-asymptotic bounds presented below are for a constant step-size, while the Polyak-Ruppert averaging scheme uses diminishing step-sizes of the form $\frac{1}{k^\alpha}$, for some $\alpha\in (1/2,1)$.}

Recall that the update rule of the algorithm in the unconstrained involved a projection operator, which is not needed for the non-asymptotic bound that we present below. In other words, we shall analyze the following projection-free variant of the risk-sensitive policy gradient algorithm from a non-asymptotic viewpoint:
	\begin{align}
	\theta_{n+1} &= \theta_n - \zeta(n)  \widehat \nabla G(\theta_n).
	\label{eq:risk-pg-noproj}
	\end{align}

We require the following assumption for deriving the non-asymptotic bound:
\begin{assumption}
	\label{ass:smoothG}
There exists a constant $ L_1 > 0 $ such that
\[	\| \nabla G (\theta ) - \nabla G ( \theta' ) \|  \leq L_1 \| \theta - \theta' \|,
\quad \forall \theta , \theta' \in \mathbb { R } ^ {d}.\]
\end{assumption}
The smoothness requirement on the risk objective is not stringent, considering one would require such an assumption for obtaining the gradient estimates in \ref{ass:unbiased-gradG}/\ref{ass:biased-gradG}. 

The main result providing the non-asymptotic bound of the policy gradient algorithm in \eqref{eq:theta_descent_unconstrained} is given below.
\boxed{
\vspace*{-10pt}
\begin{theorem} \label{thm:biased_sp}\ \\
(i)	Assume  \ref{ass:unbiased-gradG} and \ref{ass:smoothG} hold.  Set $\zeta(k) =  \min \bigg\{\frac{1}{L_1}, \frac{1}{\sqrt{n}}\bigg\}$ for $k=1,\ldots,n$.
	Let $\theta_R$ be chosen uniformly at random from $\{\theta_1,\ldots,\theta_n\}$, and let $\theta^*$ be a global optima of $G$.
	Then, for any $ n \ge 1 $, 
	\begin{align} 
		& \mathbb { E }  \left\| \nabla G \left( \theta_{ R } \right) \right\| ^ { 2 }   \le \frac{ 2 L_1 \left(G(\theta_1) - G(\theta^*)\right)}{{ n }} + \frac{\left[ 2 \left(G(\theta_1) - G(\theta^*)\right) +  L_1 \sigma^2 \right]}{\sqrt{n}}. \label{eq:riskpg_likelihood}
	\end{align}
(ii) If instead \ref{ass:biased-gradG} and \ref{ass:smoothG} hold and $ \| \nabla G\left( \theta \right) \|_1 \leq B $ for any $\theta \in \R^d$, then setting $\zeta(k) =  \min \bigg\{\frac{1}{L_1}, \frac{1}{n^{2/3}}\bigg\}$ and $\delta_k = \frac{1}{n^{1/6}}$,  for $k=1,\ldots,n$, 
	\begin{align} 
	& \mathbb { E }  \left\| \nabla G \left( \theta_{ R } \right) \right\| ^ { 2 }   \le \nonumber\\
&	\frac{ 2 \left(G(\theta_1) - G(\theta^*)\right)}{{ n }} \max\bigg\{L_1, n^{2/3}\bigg\}  + \frac{4 B  C_1 +L_1 C_2 }{n^{1/3}} +   \frac{L_1 dC_1^2 }{n^{4/3}},~~~~~~ \label{eq:riskpg_spsa}
\end{align}
where the constants $C_1$ and $C_2$ are given in assumption \ref{ass:biased-gradG}.  
\end{theorem}}
\diff{
The non-asymptotic bound in the result above of the order $O\left(\frac{1}{\sqrt{n}}\right)$ for the case when unbiased gradient estimates are available. On the other hand, the corresponding bound for the case of biased gradient estimates is $O\left(\frac{1}{n^{1/3}}\right)$. The weaker rate is a result of the fact that the gradient estimates exhibit a bias variance tradeoff via the perturabation constant $\delta_n$, i.e., decreasing $\delta_n$ leads to a gradient estimate that has lower bias, though this is at the cost of increasing variance.
\begin{proof}
	We first prove part (i), where \eqref{eq:riskpg_likelihood} provides a finite-time bound for the SA iterative update \eqref{eq:risk-pg-noproj} using unbiased gradient estimates with bounded variance, i.e., gradient estimators satisfying \ref{ass:unbiased-gradG}.

	Using \ref{ass:smoothG}, we have
	\begin{align}
		G \left( \theta_ { k + 1 } \right) 	& \leq G \left( \theta_ { k } \right) + \left\langle \nabla G \left( \theta_ { k } \right) , \theta_ { k + 1 } - \theta_ { k } \right\rangle + \frac { L_1 } { 2 } \left\| \theta_ { k + 1 } - \theta_ { k } \right\| ^ { 2 } \nonumber \\ 
		& = G \left( \theta_ { k } \right) - \zeta(k) \left\langle \nabla G \left( \theta_ { k } \right) , \widehat \nabla G(\theta_k) \right\rangle  + \frac { L_1 } { 2 } \zeta(k)^2 \left\| \widehat \nabla G(\theta_k) \right\| ^ { 2 } \label{eq:take_exp1} .
	\end{align}	
	Note that the search point $ \theta_k $ is a function of the history $\{\theta_m, m<k\}$, and is random. 
	Let $\E_{k}$ denote the expectation w.r.t. the sigma field $\F_k=\sigma(\theta_m, m<k)$.  Taking expectations with respect to $ \E_{k} $ on both sides of above equation, and noting that from \ref{ass:unbiased-gradG}, we have $ \mathbb { E }_{k}  \left[ \widehat \nabla G(\theta_k) \right] =  \nabla G \left( \theta _ { k } \right) ,  $ and 	$ \mathbb { E }_{k} \left[ \left\| \widehat \nabla G(\theta_k) \right\|^{2} \right] \le \left\| \nabla G(\theta_n) \right\| ^{2} + \sigma^2 $ , we obtain
	\begin{align}
		\mathbb { E }_{k}  [G \left( \theta_ { k + 1 } \right) ]
		& \leq G \left( \theta_ { k } \right) - \zeta(k) \left\| \nabla G \left( \theta_ { k } \right) \right\| ^ { 2 }  + \frac { L_1 } { 2 } \zeta(k) ^ { 2 } \left[ \left\| \nabla G \left( \theta_ { k } \right) \right\| ^ { 2 } +\sigma^2 \right]\nonumber\\
		& = G \left( \theta_ { k } \right) - \left( \zeta(k) - \frac { L_1 } { 2 } \zeta(k) ^ { 2 } \right) \left\| \nabla G \left( \theta_ { k } \right) \right\| ^ { 2 } + \frac { L_1 } { 2 } \zeta(k) ^ { 2 } \sigma^ { 2 }. \nonumber
	\end{align}
	Rearranging the terms, we obtain	
	\begin{align*}
&		\left( \zeta(k) - \frac { L_1 } { 2 } \zeta(k) ^ { 2 } \right) \left\| \nabla G \left( \theta_ { k } \right) \right\| ^ { 2 }  \leq G \left( \theta_ { k } \right) - 	\mathbb { E }_{k} [G \left( \theta_ { k + 1 } \right)]  + \frac { L_1 } { 2 } \zeta(k) ^ { 2 } \sigma ^ { 2 } \\
	& \Longrightarrow	\zeta(k)  \left\| \nabla G \left( \theta_ { k } \right) \right\| ^ { 2 } 
		\leq \frac{2 \left[ G \left( \theta_ { k } \right) - 	\mathbb { E }_{k} [G \left( \theta_ { k + 1 } \right)] \right]}{\left( 2-  { L_1 }  \zeta(k) \right)}   +  \frac{L_1  \zeta(k) ^ { 2 } \sigma ^ { 2 }}{\left( 2-  { L_1 }  \zeta(k) \right)}.
	\end{align*}
	Now summing up the above inequality from $ k = 1 $ to $ n $, we obtain
	\begin{align*}
		&\sum _ { k = 1 } ^ { n } \zeta(k)  \left\| \nabla G \left( \theta_ { k } \right) \right\| ^ { 2 }  \\
		&\leq 2 \sum _ { k = 1 } ^ { n } \frac{ \left[ G \left( \theta_ { k } \right) - 	\mathbb { E }_{k} [G \left( \theta_ { k + 1 } \right)] \right]}{\left( 2-  { L_1 }  \zeta(k) \right)} + L_1 \sigma^2 \sum _ { k = 1 } ^ { n } \frac{  \zeta(k) ^ { 2 } }{\left( 2-  { L_1 }  \zeta(k) \right)}.
	\end{align*}
	Taking expectations on both sides of the equation above, we obtain	
	\begin{align*}
		&\sum _ { k = 1 } ^ { n } \zeta(k) \mathbb{E}_{n} \left\| \nabla G \left( \theta_ { k } \right) \right\| ^ { 2 }\\
		 & \leq 2 \sum _ { k = 1 } ^ { n } \frac{ \left[ \mathbb{E}_{n} \left[G \left( \theta_ { k } \right)\right] - 	\mathbb{E}_{n}  \left[G \left( \theta_ { k + 1 } \right)\right] \right] }{ \left( 2-  { L_1 }  \zeta(k) \right) } +  L_1 \sigma^2 \sum _ { k = 1 } ^ { n }\frac{ \zeta(k)^2}{\left( 2-  { L_1 }  \zeta(k) \right)} \\
		& =  2 \left[\frac { G \left( \theta_ { 1 } \right) } {  \left( 2-  { L_1 }  \zeta _ { 1 } \right) } - \sum _ { k = 2 } ^ { n } \left( \frac { 1 } { \left( 2-  { L_1 }  \zeta _ { k-1 } \right) } - \frac { 1 } { \left( 2-  { L_1 }  \zeta(k) \right) } \right) \mathbb { E }_{n}  \left[ G \left( \theta_ { k } \right) \right]\right.\\
		&\qquad\left. - \frac { \mathbb { E }_{n}  \left[ G \left( \theta_ { n + 1 } \right) \right] } { \left( 2-  { L_1 }  \zeta _ { n } \right) } \right] +  L_1 \sigma^2 \sum _ { k = 1 } ^ { n }\frac{ \zeta(k)^2}{\left( 2-  { L_1 }  \zeta(k) \right)}.
	\end{align*}
	Notice that since step sizes $ \{\zeta(k)\}_{k \ge 1} $ are non-increasing, we have \\$ \left( \frac { 1 } { \left( 2-  { L_1 }  \zeta _ { k-1 } \right) } - \frac { 1 } { \left( 2-  { L_1 }  \zeta(k) \right) } \right)  \ge 0 $ and using the fact that $ \mathbb { E }_{n}  \left[ G \left( \theta_ { k } \right) \right] \ge G(\theta^*) $, where $\theta^*$ is a local optima, we obtain
	\begin{align*}
		&\sum _ { k = 1 } ^ { n } \zeta(k) \mathbb{E}_{n} \left\| \nabla G \left( \theta_ { k } \right) \right\| ^ { 2 }  \\
		& \leq 2 \left[ \frac { G(\theta_1) } { \left( 2-  { L_1 }  \zeta _ { 1 } \right) } - G(\theta^*) \sum _ { k = 2 } ^ { n } \left( \frac { 1 } { \left( 2-  { L_1 }  \zeta _ { k-1 } \right) } - \frac { 1 } { \left( 2-  { L_1 }  \zeta(k) \right) } \right) - \frac { G(\theta^*)  } { \left( 2-  { L_1 }  \zeta _ { n } \right) } \right]\\
		& \qquad+  L_1 \sigma^2 \sum _ { k = 1 } ^ { n }\frac{ \zeta(k)^2}{\left( 2-  { L_1 }  \zeta(k) \right)}\\
		& = \frac { 2 \left(G(\theta_1) - G(\theta^*)\right)} { \left( 2-  { L_1 }  \zeta _ { 1 } \right)} +  L_1 \sigma^2 \sum _ { k = 1 } ^ { n }\frac{ \zeta(k)^2}{\left( 2-  { L_1 }  \zeta(k) \right)}.
	\end{align*}
 Let $R$ be a r.v. with the following mass function:
 	\begin{align} 
 	P_R(k) := \Prob{R=k} = \frac{\zeta(k)}{\sum _ { k = 1 } ^ { n} { \zeta(k) }},  \quad k = 1,\dots,n. \label{eq:prob}
 \end{align}
	It follows from the definition of $ P_R $ above that,
	\begin{align*} 
		\mathbb { E } \left[ \left\| \nabla G \left( \theta_ { R } \right) \right\| ^ { 2 } \right]  = \frac{\sum _ { k = 1 } ^ { n} { \zeta(k) } \mathbb{E}_{n} \left\| \nabla G \left( \theta_ { k } \right) \right\| ^ { 2 } }{\sum _ { k = 1 } ^ { n }\zeta(k) } .
	\end{align*}
	Thus, we conclude
	\begin{align} 
		\mathbb { E } \left[ \left\| \nabla G \left( \theta_ { R } \right) \right\| ^ { 2 } \right]  &\le \frac{1}{\sum _ { k = 1 } ^ { n }\zeta(k) } \left[ \frac { 2  \left(G(\theta_1) - G(\theta^*)\right) } { \left( 2-  { L_1 }  \zeta _ { 1 } \right)} +  L_1 \sigma^2 \sum _ { k = 1 } ^ { n }\frac{ \zeta(k)^2}{\left( 2-  { L_1 }  \zeta(k) \right)} \right].\label{eq:rpg-nonasymp-unbiased-general}
	\end{align}
	The bound in the equation above holds for a general step-size choice. Specializing the bound in \eqref{eq:rpg-nonasymp-unbiased-general} to the case of a constant step size $ \zeta(k) = \left\{\zeta=\min \bigg\{\frac{1}{L_1}, \frac{1}{\sqrt{n}}\bigg\}\right\}, \forall k \geq 1 $, we obtain
	\begin{align*}
		\mathbb { E } \left[ \left\| \nabla G \left( \theta_ { R } \right) \right\| ^ { 2 } \right]  &\le \frac{1}{\sum _ { k = 1 } ^ { n }\zeta(k) } \left[ \frac { 2 \left(G(\theta_1) - G(\theta^*)\right)} { \left( 2-  { L_1 }  \zeta _ { 1 } \right)} +  L_1 \sigma^2 \sum _ { k = 1 } ^ { n }\frac{ \zeta(k)^2}{\left( 2-  { L_1 }  \zeta _ { k } \right)} \right]	\\
		& = \frac{1}{{ n }\zeta } \left[ \frac { 2 \left(G(\theta_1) - G(\theta^*)\right)} { \left( 2-  { L_1 }  \zeta  \right)} +  L_1 \sigma^2  { n }\frac{ \zeta^2}{\left( 2-  { L_1 }  \zeta  \right)}\right]\\
		& \le  \frac{1}{{ n }\zeta } \left[ { 2 \left(G(\theta_1) - G(\theta^*)\right)}  +  L_1 \sigma^2  { n }{ \zeta^2}\right]\\
		& = \frac{ 2 \left(G(\theta_1) - G(\theta^*)\right)}{{ n }\zeta}  +  L_1 \sigma^2 \zeta\\
		& \le \frac{ 2 \left(G(\theta_1) - G(\theta^*)\right)}{{ n }} \max\bigg\{L_1, \sqrt{n}\bigg\}  +  L_1 \sigma^2 \frac{1}{\sqrt{n}} \\
		& \le \frac{ 2 L_1 \left(G(\theta_1) - G(\theta^*)\right)}{{ n }} + \frac{ 2 \left(G(\theta_1) - G(\theta^*)\right)}{{ \sqrt{n} }} +  L_1 \sigma^2 \frac{1}{\sqrt{n}} \\
		& = \frac{ 2 L_1 \left(G(\theta_1) - G(\theta^*)\right)}{{ n }} + \frac{1}{\sqrt{n}}\left[ 2 \left(G(\theta_1) - G(\theta^*)\right) +  L_1 \sigma^2 \right],
	\end{align*}
	establishing \eqref{eq:riskpg_likelihood} to complete the proof of part (i).
	
\medskip
\noindent
\textbf{Proof of part (ii).}
		Now we prove the second part of the theorem, specifically the finite-time bound \eqref{eq:riskpg_spsa} for the SA iterative update \eqref{eq:risk-pg-noproj} using biased gradient estimators satisfying \ref{ass:biased-gradG}.
	As before, using \ref{ass:smoothG}, we have
	\begin{align}
		G \left( \theta_ { k + 1 } \right) 	& \leq G \left( \theta_ { k } \right) - \zeta(k) \left\langle \nabla G \left( \theta_ { k } \right) , \widehat \nabla G(\theta_k) \right\rangle  + \frac { L_1 } { 2 } \zeta(k)^2 \left\| \widehat \nabla G(\theta_k) \right\| ^ { 2 }.~~~ \label{eq:take_exp2} 
	\end{align}	
	Taking expectations w.r.t. $\E_k$ on both sides of \eqref{eq:take_exp2}, followed by an application of the inequalities in \ref{ass:biased-gradG}, we obtain
	\begin{align}
		& \E_{k} \left[G\left( \theta_ { k + 1 } \right)\right] \nonumber \\
		& \leq \E_{k} \left[G\left( \theta_ { k } \right) \right] - \zeta(k) \left\langle \nabla G\left( \theta_ { k } \right) , \nabla G\left( \theta_ { k } \right) + C_1 \delta_k^2  \mathbf{1}_{d \times 1} \right\rangle  \nonumber\\
		& \quad
		+ \frac { L_1 } { 2 } \zeta(k) ^ { 2 } \left[ \left\| \E_{k}  \left[ \widehat \nabla G(\theta_k) \right]  \right\| ^{2} + \frac{C_2}{\delta_k^2} \right] \nonumber \\
		&  \leq G\left( \theta_ { k } \right)
		- \zeta(k) \left\| \nabla G\left( \theta_ { k } \right) \right\| ^ { 2 } + C_1 \delta_k^2 \zeta(k)    \mathbb{E}_{k}  \| \nabla G\left( \theta_ { k } \right) \|_1  \nonumber \\
		& \quad + \frac { L_1 } { 2 } \zeta(k) ^ { 2 } \hspace*{-3pt} \left[ \hspace*{-1pt} \left\| \nabla G\left( \theta_ { k } \right) \right\|^2 \hspace*{-2pt}  + 2 C_1 \delta_k^2  \mathbb{E}_{k}  \| \nabla G\left( \theta_ { k } \right) \|_1 \hspace*{-2pt} + {d}C_1^2 \delta_k^4 + \frac{C_2}{\delta_k^2} \right]~~~ \label{eq:used_1norm_ineq}\\
		& \leq G\left( \theta_ { k } \right) - \left( \zeta(k) - \frac { L_1 } { 2 } \zeta(k) ^ { 2 } \right)  \left\| \nabla G\left( \theta_ { k } \right) \right\| ^ { 2 } + C_1 \delta_k^2 B  \left( \zeta(k) + L_1 \zeta(k) ^ { 2 } \right)  \\
		&\qquad+ \frac { L_1 } { 2 } \zeta(k) ^ { 2 } \left[ d C_1^2 \delta_k^4 + \frac{C_2}{\delta_k^2}\right], \label{eq:f12}
	\end{align}
	where the inequality in \eqref{eq:used_1norm_ineq} uses $ - \|\theta\|_1 \leq \sum_{i=1}^{d} \theta_i $ for any vector $\theta$, while the final inequality uses $ \| \nabla G\left( \theta_ { k } \right) \|_1 \leq B $, which holds by an assumption in the theorem statement.
	A straightforward rearrangement of the terms in \eqref{eq:f12} leads to
	\begin{align*}
		&\zeta(k) \left\| \nabla G\left( \theta_ { k } \right) \right\| ^ { 2 }	
		\leq \frac{2}{\left( 2  -  { L_1 }  \zeta(k) \right) } \bigg[ G\left( \theta_ { k } \right) -  \mathbb{E}_k G\left( \theta_ { k + 1 } \right) \bigg. \\
		& \quad \bigg.  + C_1 \delta_k^2 \left( \zeta(k) + L_1 \zeta(k) ^ { 2 } \right)B \bigg] +  \frac{{ L_1 } \zeta(k) ^ { 2 }}{\left( 2  -  { L_1 }  \zeta(k) \right) } \left[ dC_1^2 \delta_k^4 + \frac{C_2}{\delta_k^2}\right].
	\end{align*}
	Now, summing up the inequality above for $ k = 1 $ to $ n $, and taking expectations, we obtain
	\begin{align*}
		&	\sum _ { k = 1 } ^ { n} \zeta(k) \mathbb{E}_{n}  \left\| \nabla G\left( \theta_ { k } \right) \right\| ^ { 2 } \\
		& \leq 2 \sum _ { k = 1 } ^ { n} \frac{ \left(\mathbb{E}_{n}  G\left( \theta_ { k } \right) - \mathbb{E}_{n}  G\left( \theta_{k+1} \right)  \right)}{\left( 2  -  { L_1 }  \zeta(k) \right) }  + 2 \sum _ { k = 1 } ^ { n} C_1 \delta_k^2 B \left( \frac{ \zeta(k) + L_1 \zeta(k) ^ { 2 } }{ 2  -  { L_1 }  \zeta(k)  }\right) \\
		& \quad + L_1 \sum_{ k = 1 }^{n} \frac{ \zeta(k) ^ { 2 }}{\left( 2  -  { L_1 }  \zeta(k) \right) } \left[ dC_1^2 \delta_k^4  + \frac{C_2}{\delta_k^2}\right]   \\
		& =  2 \left[\frac { G\left( \theta_ { 1 } \right) } {  \left( 2-  { L_1 }  \zeta(1) \right) } - \sum _ { k = 2 } ^ { n } \left( \frac { \E_n   G\left( \theta_ { k } \right) } { \left( 2-  { L_1 }  \zeta(k-1) \right) } - \frac { \E_n  G\left( \theta_ { k } \right) } { \left( 2-  { L_1 }  \zeta(k) \right) } \right)  - \frac { \E_n  \left[ G\left( \theta_ { n + 1 } \right) \right] } { \left( 2-  { L_1 }  \zeta(n) \right) } \right] \nonumber \\
		& \quad + 2 \sum _ { k = 1 } ^ { n} C_1 \delta_k^2 B \left( \frac{ \zeta(k) + L_1 \zeta(k) ^ { 2 } }{ 2  -  { L_1 }  \zeta(k)  }\right)  + L_1 \sum_{ k = 1 }^{n} \frac{ \zeta(k) ^ { 2 }}{\left( 2  -  { L_1 }  \zeta(k) \right) } \left[ dC_1^2 \delta_k^4  + \frac{C_2}{\delta_k^2}\right].
	\end{align*}
	Using $ \E_{n}  \left[ G\left( \theta_ { k } \right) \right] \ge G(\theta^*) $, and $ \left( \frac { 1 } { \left( 2-  { L_1 }  \zeta(k-1) \right) } - \frac { 1 } { \left( 2-  { L_1 }  \zeta(k) \right) } \right)  \ge 0 $, we obtain
	\begin{align}
		&\sum _ { k = 1 } ^ { n } \zeta(k) \mathbb{E}_{n} \left\| \nabla G\left( \theta_ { k } \right) \right\| ^ { 2 } 
		\leq \frac { 2 \left(\left(G(\theta_1) - G(\theta^*)\right) \right)} { \left( 2-  { L_1 }  \zeta(1) \right)}  
		 \nonumber \\
		& + 2 \sum _ { k = 1 } ^ { n} C_1 \delta_k^2 B \left( \frac{ \zeta(k) + L_1 \zeta(k) ^ { 2 } }{ 2  -  { L_1 }  \zeta(k)  }\right) + L_1 \sum_{ k = 1 }^{n} \frac{ \zeta(k) ^ { 2 }}{\left( 2  -  { L_1 }  \zeta(k) \right) } \left[ dC_1^2 \delta_k^4  + \frac{C_2}{\delta_k^2}\right].\nonumber
	\end{align}
	Using the fact that $\Prob{R=i} = 1/n, i=1,\ldots,n$, we have
	\begin{align}
	&\mathbb { E } \left[ \left\| \nabla G \left( \theta_ { R } \right) \right\| ^ { 2 } \right] 
	 \le \frac{1 }{\sum _ { k = 1 } ^ { n } \zeta(k) } \left[\frac { 2 \left(G(\theta_1) - G(\theta^*)\right)} { \left( 2-  { L_1 }  \zeta(1) \right)} + \right.  \nonumber \\
	& \hspace*{-9pt} \left.  2 B \sum _ { k = 1 } ^ { n} C_1 \delta_k^2 \left[ \frac{ \zeta(k) + L_1 \zeta(k) ^ { 2 } }{ 2  -  { L_1 }  \zeta(k)  }\right]      +  \sum_{ k = 1 }^{n} \frac{ L_1\zeta(k) ^ { 2 }}{\left( 2  -  { L_1 }  \zeta(k) \right) } \left[ d C_1^2 \delta_k^4  + \frac{C_2}{\delta_k^2}\right] \right].~~~~~~~ \label{eq:prop_biased_esterr_bound}
	\end{align}	
	We now specialize the result obtained in the equation above, to derive the bound in \eqref{eq:riskpg_spsa}.
Using $ \zeta(k) \triangleq \left\{\zeta =  \min \bigg\{\frac{1}{L_1}, \frac{1}{n^{2/3}}\bigg\}\right\}$, and $\delta_k \triangleq \left\{\delta = \frac{1}{n^{1/6}}\right\}$ in \eqref{eq:prop_biased_esterr_bound}, we obtain
	\begin{align}
	&	\mathbb { E } \left[ \left\| \nabla G \left( \theta_ { R } \right) \right\| ^ { 2 } \right] \nonumber \\
	& \le  \frac{1}{{ n }\zeta } \left[ { 2 \left(G(\theta_1) - G(\theta^*)\right)} + 4n \zeta B C_1 \delta^2 +  { L_1 n \zeta ^ { 2 }}\left[ d C_1^2 \delta^4  + \frac{C_2}{\delta^2}\right] \right] \nonumber \\ 
	& \le \frac{ 2 \left(G(\theta_1) - G(\theta^*)\right)}{{ n }} \max\bigg\{L_1, n^{2/3}\bigg\}  + \frac{4 B  C_1 }{n^{1/3}} +   \frac{L_1}{n^{2/3}} \left[ \frac{dC_1^2 }{n^{2/3}}   + \frac{C_2}{  n^{-1/3}}\right], 
	\end{align}
	where the first inequality above 
	follows by using the fact that $ \zeta \leq 1/L_1 $, while the final inequality 
	follows by using the definition of $\zeta$ and $ \delta $. The final bound in \eqref{eq:riskpg_spsa} then follows by rearranging terms. 
\end{proof}
}

\section{Convergence analysis in the setting with risk as constraint}
\label{sec:convergence-constrained}
In this section, we analyze the convergence properties of the two-timescale SA algorithm given by 
\eqref{eq:lambda-ascent}--\eqref{eq:theta_descent_det} in Section \ref{sec:template-constrained}.
The convergence analysis using the ODE approach on the two timescales in \eqref{eq:lambda-ascent} and \eqref{eq:theta_descent_det} is based on the following intuition, to be made rigorous:  the faster timescale recursion in \eqref{eq:theta_descent_det} sees the iterate $\lambda_n$ on the slower timescale as quasi-static, while the slower timescale recursion in  \eqref{eq:lambda-ascent} sees the iterate $\theta_n$  on the faster timescale as equilibrated. In essence, this viewpoint is equivalent to assuming that slower timescale iterate as constant while analyzing the faster timescale recursion, and using converged values of the faster timescale iterate for analysis of the slower timescale recursion. 

We make the following assumptions for the convergence analysis of the two-timescale SA recursions given by  
\eqref{eq:lambda-ascent} and \eqref{eq:theta_descent_det}.
Again, let $\F_n = \sigma\left(\theta_m, m\le n\right)$ denote the underlying $\sigma$-field.

\begin{assumption}
\label{ass:smoothJG}
The policy $\mu^\theta(\cdot | x)$ is a continuously differentiable function of $\theta$, for any $x \in \X$ and $a \in \A$.
Furthermore, the risk measure $G$ is a continuously differentiable function of the policy parameter $\theta$.
\end{assumption}

\begin{assumption}
	\label{ass:unbiased-G}
	The risk-measure estimator $\widehat G(\cdot)$ satisfies
	$\E\left(\widehat G(\theta_n) \mid \F_n\right) = G(\theta_n).$
\end{assumption}

\begin{assumption}
	\label{ass:unbiased-gradJG}
The gradient estimators $\widehat\nabla J(\theta_n)$ and $\widehat\nabla G(\theta_n)$ satisfy
\begin{align*} 
&\E\left(\widehat\nabla J(\theta_n) \mid \F_n\right) = \nabla J(\theta_n),  \E\left(\widehat\nabla G(\theta_n) \mid \F_n\right) = \nabla G(\theta_n), \\
&\textrm{ and }\E\left\| \widehat\nabla J(\theta_n) \right\|^2 + \E\left\| \widehat\nabla G(\theta_n) \right\|^2 < \infty.
\end{align*}
\end{assumption}

\begin{assumption}
	\label{ass:biased-gradJG}
\diff{The gradient estimators $\widehat\nabla J(\theta_n)$ and $\widehat\nabla G(\theta_n)$ with perturbation $\delta_n>0$ satisfy
\begin{align*}
&\left\| \E\left(\widehat\nabla J(\theta_n) \mid \F_n\right) - \nabla J(\theta_n) \right\| = C_5\ \delta_n^2,\\
&\left\| \E\left(\widehat\nabla G(\theta_n) \mid \F_n\right) - \nabla G(\theta_n) \right\| = C_6\ \delta_n^2,  \\
&\textrm{ and }\left(\E\left\| \widehat\nabla J(\theta_n) \right\|^2 + \E\left\| \widehat\nabla G(\theta_n) \right\|^2\right)< \frac{C_7}{\delta_n^2},
\end{align*}
where $C_5$, $C_6$, and $C_7$ are dimension-dependent constants.}
\end{assumption}

\begin{assumption}
	\label{ass:stepsizes-JG}
The step-size sequences $\{\zeta_1(n),\zeta_2(n)\}$ satisfy
$$
\sum\limits_n \zeta_1(n) = \sum\limits_n \zeta_2(n)  = \infty,~
\sum\limits_n (\zeta_1(n)^2+\zeta_2(n)^2)\;<\infty,~
$$
$$
~~~~~~~~~~~~~~~~~~~~~~\zeta_1(n) = o\big(\zeta_2(n)\big).
$$
\end{assumption}

\begin{assumption}
	\label{ass:stepsizes-JG-variant}
The gradient estimator perturbation $\delta_n \rightarrow 0 \textrm{ as } n \rightarrow \infty$, and 
the step-size sequences $\{\zeta_1(n),\zeta_2(n)\}$ satisfy
$$
\sum\limits_n \zeta_1(n) = \sum\limits_n \zeta_2(n)  = \infty,~
\sum\limits_n \left[\zeta_1(n)^2+\left[\frac{\zeta_2(n)}{\delta_n}\right]^2\right]<\infty,~
$$
$$
~~~~~~~~~~~~~~~~~~~~~~\zeta_1(n) = o\big(\zeta_2(n)\big).
$$
\end{assumption}
\diff{We now discuss the assumptions.
The first part in assumption \ref{ass:smoothJG} is a standard requirement in the analysis of policy gradient-type RL algorithms. The second part imposes a smoothness requirement on the risk measure $G$, when viewed as a function of the policy parameter $\theta$. As mentioned earlier, for an abstract risk measure $G$, one cannot infer smoothness of $G$ based only on the policy parameterization being smooth. This smoothness condition together with the fact that $J$ is smooth ensures  that the ODE underlying the $\theta$-recursion is well-posed.
Assumption \ref{ass:unbiased-G} is an unbiasedness requirement on the risk measure estimate that is used for dual ascent in \eqref{eq:lambda-ascent}. Assumptions \ref{ass:unbiased-gradJG} and \ref{ass:biased-gradJG} are unbiasedness/asymptotic unbiasedness requirements on the estimators of the gradient of the objective and risk measure, and are necessary to ensure  that the $\theta$-recursion in \eqref{eq:theta_descent_det} is descending in the Lagrangian objective.
Assumption \ref{ass:unbiased-gradJG} requires that the estimators of the gradient of $J$ and $G$ be unbiased, and also that the variance of the these gradient estimates is bounded. Such requirements are common in the analysis of stochastic gradient schemes, and estimators formed using the likelihood ratio method (described in the next chapter) usually satisfy \ref{ass:unbiased-gradJG}. 
Assumption \ref{ass:biased-gradJG} is a relaxed version of \ref{ass:unbiased-gradJG}, where the gradient estimators are asymptotically unbiased, i.e., the gradient estimators have a perturbation parameter $\delta_n$, which can be used to control the bias-variance tradeoff. The SPSA technique for gradient estimation, presented in Section \ref{sec:spsa}, as well as the general class of simultaneous perturbation schemes, satisfy the conditions in \ref{ass:biased-gradJG}. For convergence, $\delta_n$ has to vanish asymptotically, but not too fast, as outlined in the second part of \ref{ass:stepsizes-JG-variant}.   }

\diff{The first two conditions on the step-sizes $\{\zeta_1(n), \zeta_2(n)\}$ in \ref{ass:stepsizes-JG} are standard stochastic approximation requirements. The condition $\zeta_1(n) = o\big(\zeta_2(n)\big)$ is required for the standard two-timescale view, i.e., the policy recursion in \eqref{eq:theta_descent_det} views the Lagrange multiplier as quasi-static, while the Lagrange multiplier recursion views the policy parameter as almost equilibrated. This view is made precise in the convergence analysis presented in Section \ref{sec:convergence-constrained} below. Two-timescale updates are convenient because both policy and Lagrange multiplier can be updated in parallel, albeit with varying step-sizes. The latter are chosen carefully so that one is able to mimic a two-loop behavior, with policy updates in the inner loop and Lagrange multiplier updates in the outer loop. 
Assumption \ref{ass:stepsizes-JG-variant} is a variant of \ref{ass:stepsizes-JG}, which when coupled with (A3') ensures that the bias $O(\delta_n^2)$ of gradient estimates vanish asymptotically. Further, the noise in gradient estimates can be ignored in the asymptotic analysis owing to the condition $\sum_n \left(\frac{\zeta_2(n)}{\delta_n}\right)^2 < \infty$. }

We adopt the ODE approach for analyzing the template algorithm in \eqref{eq:lambda-ascent}--\eqref{eq:theta_descent_det}. In particular, 
under the assumptions listed above, the ODE governing the policy update, for any given Lagrange multiplier $\lambda$, is given by
\begin{align}
\label{eq:theta-ode}
\dot{\theta}(t) = \check{\Gamma}\left ( -\nabla J(\theta(t)) - \lambda \nabla G(\theta(t))\right),
\end{align}
where $\check{\Gamma}(\cdot)$ is a projection operator  that ensures the evolution of $\theta$ via the ODE (\ref{eq:theta-ode}) stays within the set $\Theta$ and is defined as in \eqref{eq:Pi-bar-operator}.

\begin{remark}(Two-timescale view)
\diff{In describing the ODE governing the policy recursion, we have assumed that the Lagrange multiplier is constant, and this view can be justified as follows:
First rewrite the $\lambda$-recursion as
\[
\lambda_{n+1} = \bigg[\lambda_n + \zeta_2(n) \left( \frac{\zeta_1(n)}{\zeta_2(n)} \left( G(\theta_n) - \kappa + \varsigma_{1,n}\right) \right)\bigg]^+,\]
where $\varsigma_{1,n}$ is a martingale difference sequence (a consequence of (A1)). Considering that we have a finite-dimensional MDP setting, together with the fact that $\frac{\zeta_1(n)}{\zeta_2(n)} = o(1)$ (see (A4)), it is clear that the $\lambda$-recursion above tracks the ODE $\dot \lambda(t) = 0$.}

\diff{The claim that the $\lambda$-recursion views the policy parameter as almost equilibrated requires a more sophisticated argument, and we provide a proof sketch below.
We first rewrite the $\lambda$ update iteration as follows:
\begin{align*}
	\lambda_{n+1} &= \bigg[\lambda_n + \zeta_1(n)\left( G(\theta_{\lambda_n}) - \kappa + \varsigma_{2,n}\right)\bigg]^+,
\end{align*}
where $\varsigma_{2,n}= G(\theta_n)  -  G(\theta_{\lambda_n}) $. The noise factor $\varsigma_{2,n}$ is defined using the fast timescale parameter $\theta_{\lambda_n}$ with the slow timescale iterate $\lambda_n$. The parameter $\theta_{\lambda_n}$ is a limiting point of the $\theta$-recursion, with the Lagrange multiplier $\lambda_n$. Owing to the convergence of $\theta$-recursion, one can infer that $\varsigma_{2,n} = o(1)$, i.e., $\varsigma_{2,n}$ adds an asymptotically vanishing bias term to the $\lambda$-recursion above.
Thus, it is apparent that the $\lambda$-recursion views the policy parameter as almost equilibrated, and the technical proof proceeds by showing that the $\lambda$-recursion tracks the following ODE: 
\[\dot \lambda(t) \;\;=\;\; \check\Gamma_\lambda\big[G(\theta_{\lambda(t)}) - \kappa\big],\] 
where $\check\Gamma_\lambda$ is a projection operator that is defined as follows: For any bounded continuous function $f(\cdot)$,
\begin{align}
	\label{eq:gamma-lambda}
	\check{\Gamma}_\lambda\big(f(\lambda)\big) = \lim\limits_{\tau \rightarrow 0}
	\dfrac{\left(\lambda + \tau f(\lambda)\right)^+ - \lambda}{\tau}.
\end{align}
The projection operator $\check\Gamma$ ensures that the $\lambda$-recursion stays within $[0,\infty)$.
The tools used in establishing such a claim are classic for stochastic approximation schemes, e.g., Theorem \ref{thm:sa-conv} in the previous chapter. }

\end{remark}
\subsection{Convergence of policy parameter}
\boxed{
\vspace*{-10pt}
\begin{theorem}
\label{thm:theta-convergence}
Assume \ref{ass:smoothJG}, \ref{ass:unbiased-G}, (\ref{ass:unbiased-gradJG} + \ref{ass:stepsizes-JG})  or (\ref{ass:biased-gradJG} + \ref{ass:stepsizes-JG-variant}). 
If $\lambda_n = \lambda~\forall n$, then for $\theta_n$ governed by \eqref{eq:theta_descent_det}, 
\[\theta_n \rightarrow \Z_{\lambda} \textrm{ a.s. as } n \rightarrow \infty,\]
where $\Z_\lambda=\big\{\theta\in \Theta:\check\Gamma\big(-\nabla J(\theta(t)) - \lambda \nabla G(\theta(t))\big)=0\big\}$ is the set of limit points of the ODE \eqref{eq:theta-ode}. 
\end{theorem}}

\begin{proof}
The proof again involves an application of the Kushner-Clark lemma for projected stochastic approximation, i.e., Theorem \ref{thm:kc} in Section \ref{sec:projSA}. 

We first rewrite the recursion \eqref{eq:theta_descent_det} as follows 
(with $\lambda_n = \lambda$):
$$
	\theta_{n+1} = \Gamma \bigg( \theta_n - \zeta_2(n) \Big(\nabla  L(\theta_n,\lambda) + \xi_{n}  \Big)\bigg)\label{eq:theta-update-equiv},
$$
where 
\begin{align*}
L(\theta,\lambda) & = J(\theta)+\lambda\big(G(\theta)-\kappa\big), \\
\xi_{n} &= \widehat\nabla J(\theta_n) + \lambda \widehat \nabla G(\theta_n) - \nabla L(\theta_n,\lambda).
\end{align*}

We now verify conditions \ref{ass:LipschitzF}--\ref{ass:noise-sa} of Theorem \ref{thm:kc} for \eqref{eq:theta-update-equiv} in the case where we assume \ref{ass:smoothJG}, \ref{ass:unbiased-G}, \ref{ass:unbiased-gradJG}, and \ref{ass:stepsizes-JG}. Later we provide the deviations necessary to handle the case when \ref{ass:biased-gradJG} and \ref{ass:stepsizes-JG-variant} are used in place of \ref{ass:unbiased-gradJG} and \ref{ass:stepsizes-JG}, respectively.
\begin{itemize}
	\item From \ref{ass:smoothJG}, we have that the policy $\mu^\theta$ is a continuously differentiable function of $\theta$, which implies the value function $J$ is continuously differentiable in $\theta$, as well. This fact combined with the second part of \ref{ass:smoothJG}, which imposed a smoothness requirement on the risk measure $G$,  imply that the condition  \ref{ass:LipschitzF} follows for $\nabla L(\theta_n, \lambda)$.
 \item Since \ref{ass:unbiased-gradJG} implies unbiased gradient estimators, the assumption \ref{ass:betaVanish} trivially holds.
	\item \ref{ass:stepsizes-JG} implies \ref{ass:stepsize-sa}.
	\item Using the third condition in \ref{ass:unbiased-gradJG}, it is easy to infer that \\$	\E\left\| \xi_{n}\right\|^2 \le C_8 < \infty$.
 Using this fact in conjunction with Doob's martingale inequality stated earlier, we obtain
	\begin{align*}
		\lim_{k\rightarrow\infty} \Prob{ \sup_{l\geq k}   \left\|\sum_{n=k}^{l} \zeta_2(n) \xi_{n}\right\| \geq \epsilon} \le & \dfrac{1}{\epsilon^2} \lim_{k\rightarrow\infty}\sum_{n=k}^{\infty} \zeta_2(n)^2 \E\left\| \xi_{n}\right\|^2 \\
		\le & \dfrac{C_8}{\epsilon^2} \lim_{k\rightarrow\infty}\sum_{n=k}^{\infty} \zeta_2(n)^2 \rightarrow 0, 
	\end{align*}
where the last limit follows from the square summability of the step-size $\zeta_2(n)$, which is assumed in \ref{ass:stepsizes-JG}.	
Thus, \ref{ass:noise-sa} is satisfied.
	\item $\Z_\lambda$ is an asymptotically stable attractor for the ODE~\eqref{eq:theta-ode}, with $L(\theta,\lambda)$ itself serving as a strict Lyapunov function. This can be inferred as follows:
	\begin{align*}
		\dfrac{d L(\theta,\lambda)}{d t}  
		= \nabla L(\theta,\lambda) \dot \theta
		= \nabla  L(\theta,\lambda) \check\Gamma\big(-\nabla  L(\theta,\lambda)\big) < 0,\; \forall \theta \notin \Z_\lambda.  
	\end{align*}
\end{itemize}
The claim now follows by an application of Theorem \ref{thm:kc}. 
\hfill $\Box$

\medskip
For the case when \ref{ass:biased-gradJG} and \ref{ass:stepsizes-JG-variant} are used in place of \ref{ass:unbiased-gradJG} and \ref{ass:stepsizes-JG}, the proof again follows by verifying conditions \ref{ass:LipschitzF}--\ref{ass:noise-sa} of Theorem \ref{thm:kc}. Of these, \ref{ass:LipschitzF} and \ref{ass:stepsize-sa} hold as shown above. The conditions \ref{ass:betaVanish} and \ref{ass:noise-sa} require a few deviations from the proof above, and we provide the details below.

First, we rewrite the $\theta$-recursion as follows:
\begin{align}
	\theta_{n+1} = &\Gamma \bigg( \theta_n - \zeta_2(n) \Big(\nabla  L(\theta_n,\lambda) + \xi_{1,n} + \xi_{2,n} \Big)\bigg)\label{eq:theta-update-equiv-again},
\end{align}
where
\begin{align*}
	\xi_{1,n} =& \E\left(\widehat\nabla J(\theta_n) + \lambda \widehat \nabla G(\theta_n) \mid \theta_n\right) - \nabla  L(\theta_n,\lambda),\\
	\xi_{2,n} =& \widehat\nabla J(\theta_n) + \lambda \widehat \nabla G(\theta_n) - \E\left(\widehat\nabla J(\theta_n) + \lambda \widehat \nabla G(\theta_n) \mid \theta_n\right).
\end{align*}
\begin{itemize}
\item From \ref{ass:biased-gradJG},  we have $\xi_{1,n} = O(\delta_n^2)$. Using \ref{ass:stepsizes-JG-variant}, we have $\xi_{1,n}\rightarrow 0$ as $n\rightarrow \infty$. This verifies \ref{ass:betaVanish}.
\item For verifying \ref{ass:noise-sa}, first note that $	\E\left\| \xi_{2,n}\right\|^2 \le C_9 < \infty$ using the last condition in \ref{ass:biased-gradJG}. As before, applying Doob's martingale inequality and using square-summability of step-sizes $\{\zeta_2(n)\}$, 
	\begin{align*}
	\lim_{k\rightarrow\infty} \Prob{  \sup_{l\geq k}   \left\|\sum_{n=k}^{l} \zeta_2(n) \xi_{2,n}\right\| \geq \epsilon } \le & \dfrac{1}{\epsilon^2} \lim_{k\rightarrow\infty}\sum_{n=k}^{\infty} \zeta_2(n)^2 \E\left\| \xi_{2,n}\right\|^2 \\
	\le & \dfrac{C_9}{\epsilon^2} \lim_{k\rightarrow\infty}\sum_{n=k}^{\infty} \frac{\zeta_2(n)^2}{\delta_n^2}\rightarrow 0. 
	\end{align*}
Thus \ref{ass:noise-sa} holds.
\end{itemize}
Hence, the claim follows for the case with assumptions \ref{ass:smoothJG}, \ref{ass:unbiased-G}, \ref{ass:biased-gradJG}, and \ref{ass:stepsizes-JG-variant}.
\end{proof}

\begin{remark}
The policy update \eqref{eq:theta_descent_det} might not converge to a local minimum, and instead get trapped in undesirable equilibria. As discussed earlier in Section \ref{sec:sa}, stochastic (i.e., inherently noisy) gradient estimators that drive the policy update may ensure avoidance of traps. An alternative is to add extraneous noise, as in \eqref{eq:sg-noiseextra}.   
\end{remark}
\subsection{Convergence of Lagrange multiplier}
We now turn to the analysis of $\lambda$-recursion in \eqref{eq:lambda-ascent}. The ODE underlying the Lagrange multiplier is given below:
\begin{align}
\label{eq:lambda-ode}
 \dot \lambda(t) \;\;=\;\; \check\Gamma_\lambda\big[G(\theta_{\lambda(t)}) - \kappa\big],
\end{align}
where $\theta_{\lambda(t)}$ is the converged value of the $\theta$-recursion, when the Lagrange multiplier is set to $\lambda(t)$, and the operator $\check{\Gamma}_\lambda$ is defined in \eqref{eq:gamma-lambda}.

\boxed{
\vspace*{-10pt}
\begin{theorem}
\label{thm:lambda-convergence}
Assume \ref{ass:smoothJG}, \ref{ass:unbiased-G}, (\ref{ass:unbiased-gradJG} + \ref{ass:stepsizes-JG})  or (\ref{ass:biased-gradJG} + \ref{ass:stepsizes-JG-variant}). Then, for $\lambda_n$ governed by \eqref{eq:lambda-ascent},  
\[\lambda_n \rightarrow \Z \textrm{ a.s. as } n \rightarrow \infty,\]
where $\Z=\left\{\lambda\in [0,\infty):\check\Gamma_\lambda\left(G(\theta_{\lambda}) - \kappa\right)=0, \theta_\lambda \in \Z_\lambda\right\}$ is the set of limit points of the ODE \eqref{eq:lambda-ode} and $\Z_\lambda$ is defined in Theorem \ref{thm:theta-convergence}. 
\end{theorem}}
\begin{proof}
	Let $H(\lambda)= \min_{\theta \in \Theta} L(\theta,\lambda)$.
	Note that the function $H$ is a pointwise infimum of a family of affine functions of $\lambda$, hence concave. Thus it is differentiable except at at most countably many points, where its has right and left derivatives. Furthermore, the right derivative at a point does not exceed the left derivative and both right and let derivatives are monotone decreasing. For simplicity of exposition, we assume below that $H$ is differentiable everywhere, the argument can easily be adapted to the more general case by using super-gradients where needed. 
	Also, $H$ attains its maximum at a unique point, viz., the Lagrange multiplier $\lambda^*$. It follows that $H(\lambda)\downarrow -\infty$ as $\lambda \rightarrow \pm \infty$.  In fact, we have some $C,c>0$ such that
	\begin{align}
		H'(\lambda) &< -c\lambda, \forall \lambda \ge C, \label{eq:l1}\\
		H'(\lambda) &> c\lambda, \forall \lambda\le -C. \label{eq:l2}
	\end{align}
	The Lagrange multiplier iterate $\lambda_n$ tracks the following ODE:
	\begin{align}
		\dot \lambda(t) = H'(\lambda(t)).
	\end{align} 
	In view of \eqref{eq:l1}--\eqref{eq:l2}, a straightforward adaptation of the arguments of Theorem \ref{thm:borkarmeyn} ensure the a.s. stability of the iterates.\\
Given that the $\lambda$ iteration is stable, the rest of the proof follows by using arguments similar to those employed in proving convergence of standard stochastic approximation algorithms, and we omit the details.	
\end{proof}

So far, we have shown that $\left(\theta_n,\lambda_n\right)$ converges to $\left(\theta_{\lambda^*},\lambda^*\right)$, for some $\lambda^*$ satisfying $\check\Gamma_\lambda\left(G(\theta_{\lambda}) - \kappa\right)=0$, and $\theta_{\lambda^*} \in \Z_{\lambda^*}$. For a given $\lambda$, the condition 
	$\check\Gamma_\lambda(G(\theta_{\lambda})-\kappa)
	=0$
	is the same as
	$\check\Gamma_\lambda(\nabla_{\lambda} L(\theta_{\lambda}, \lambda))
	=0$.

We now state a result that helps us understand if the limit $\left(\theta_{\lambda^*},\lambda^*\right)$ of the tuple $(\theta_n,\lambda_n)$ is a local saddle point of the Lagrangian, and if $\theta_{\lambda^*}$ satisfies the risk constraint.
\boxed{
\vspace*{-10pt}
	\begin{theorem}
		Let $H(\lambda)= \min_{\theta \in \Theta} L(\theta,\lambda)$.
The ODE underlying the $\lambda$-recursion in \eqref{eq:lambda-ode} is the same as 	
\begin{align}
	 \dot \lambda(t) \;\;=\;\; \check\Gamma_\lambda\big[\nabla_{\lambda} H(\lambda(t)) \big],\label{eq:lambda-equiv-ode}
	\end{align}
where the latter ODE is to be interpreted in the `Caratheodory' sense, i.e., 
    \vspace*{-10pt}
	\begin{align} 
    \vspace*{-10pt}
	\lambda(t) = \lambda(0) + \int_0^t \check\Gamma_\lambda\big[\nabla_{\lambda} H(\lambda(s))\big]ds, t\ge 0.
	\label{eq:lambda-integral}
	\end{align}
Thus, the iterate $\lambda_n$ governed by \eqref{eq:lambda-ascent} converges to a local minima of $H$. 	
\end{theorem}
}
\begin{proof}(Sketch)
For inferring the main claim in the result above, one invokes the envelope theorem of mathematical economics. In particular, using this theorem, it follows that at every point $\lambda(t)$ where the function $\tilde H = \check\Gamma_\lambda H(\cdot)$ is differentiable, the RHS of the ODE \eqref{eq:lambda-equiv-ode} coincides with that in the integral equation \eqref{eq:lambda-integral}. At points where the function $\tilde H$ is not differentiable, it can be argued that the ODE spends zero time, provided they are not the global minima of $\tilde H$.  
\end{proof}

\begin{remark}\label{rem:constraint}
Notice that the limiting policy $\theta_{\lambda^*}$ corresponding to $\lambda^* \in \Z$ satisfies the risk constraint $G(\theta_{\lambda^*}) \le \kappa$, since $\lambda^*$ corresponds to the equilibrium of the ODE $	 \dot \lambda(t) = \nabla_{\lambda} H(\lambda(t))$ that is constrained to remain in $[0,\infty)$, implying $\nabla_\lambda H(\lambda^*)=0$.  
\end{remark}

To summarize, the two-timescale risk-sensitive policy gradient algorithm with iterate  $(\theta_n,\lambda_n)$ converges to a local saddle point of the Lagrangian $L(\cdot,\cdot)$, i.e., to a point that is a local minimum w.r.t. $\theta$ and a local maximum w.r.t. $\lambda$. Moreover, the limit is a policy that  satisfies the risk constraint.  

\subsection{Projection of Lagrange multiplier onto a finite interval}
In practice, one may want to project $\lambda$ iterate in \eqref{eq:lambda-ascent} onto a finite interval, say $[0,\lambda_{\max}]$, i.e., the following update iteration:
\begin{align}
	\lambda_{n+1} &= \Gamma_\lambda\left[\lambda_n + \zeta_1(n) \left(\widehat G(\theta_n) - \kappa\right)\right],\label{eq:lambda-ascent-proj}
\end{align} 
where $\Gamma_\lambda$ denotes the projection operator onto $[0,\lambda_{\max}]$.

The analysis of this projected $\lambda$ iteration would be similar to the case analyzed before. In particular, a variant of Theorem \ref{thm:lambda-convergence} can be claimed easily, without a detailed argument for stability of iterates owing to the projection onto a finite interval.
However, the observation in Remark \ref{rem:constraint} regarding the risk constraint is not true in the case when there is projection. In other words,  the $\lambda$-recursion in \eqref{eq:lambda-ascent} involves projection on to the interval $[0,\lambda_\textrm{max}]$, and this recursion tracks the ODE given in \eqref{eq:lambda-equiv-ode}. 


However, it is possible to establish the following claims concerning the limit $(\theta_{\lambda^*},\lambda^*)$.
\boxed{
\vspace*{-10pt}
\begin{theorem}
Consider the limit set $\Z$ defined in Theorem \ref{thm:lambda-convergence}. 
\begin{itemize}
\item[(i)] For the following ``truncated'' version of $\Z$,
\[\widehat\Z=\left\{\lambda\in [0,\lambda_{\max}):\check\Gamma_\lambda\left(G(\theta_{\lambda}) - \kappa\right)=0, \theta_\lambda \in \Z_\lambda\right\}, \]
the policy $\theta_{\hat\lambda}$ corresponding to a $\hat \lambda \in \widehat \Z$ satisfies the risk constraint  $G(\theta_{\hat\lambda}) \le \kappa$.
\item[(ii)] For $\lambda^* \in \Z$, if $G(\theta_{\lambda^*}) < \kappa$, then $\lambda^*=0$, and $L(\theta_{\lambda^*}, \lambda^*) = J(\theta_{\lambda^*})$. Thus,  the risk-sensitive policy gradient algorithm converges to a local minimum of $J$ while satisfying the risk constraint.
\item[(iii)] Call $\lambda^*\in \Z$ a spurious stationary point of \eqref{eq:lambda-ode} if $\lambda^*$ is not a stationary point of the ODE $\dot \lambda(t) = G(\theta_{\lambda(t)}) - \kappa$. 
For $\lambda^* \in \Z$, if $G(\theta_{\lambda^*}) > \kappa$, then $\lambda^* = \lambda_{\textrm{max}}$, and such a $\lambda^*$ corresponds to a spurious stationary point. 
\end{itemize}
\end{theorem}
}

\begin{proof}
We prove the claim in (i) by contradiction.
Assume that 
$G(\theta_{\hat\lambda}) >\kappa$ for $\hat\lambda \in \widehat\Z$. Then, we have
\begin{align}
    \check\Gamma_\lambda(G(\theta_{\hat\lambda})-\kappa) &= \lim_{\eta\rightarrow 0}
\frac{{\Gamma_\lambda}(\hat\lambda + \eta(G(\theta_{\hat\lambda})-\kappa)) - \hat\lambda}{\eta}
= G(\theta_{\hat\lambda})-\kappa >0,~~~~~~~\label{eq:lda1}
\end{align}
which leads to a contradiction since $\check\Gamma_\lambda(G(\theta_{\hat\lambda})-\kappa)=0$ as $\hat\lambda \in \widehat\Z$. 
The equality in \eqref{eq:lda1} follows by using the facts that $\hat\lambda\geq 0$, and $G(\theta_{\hat\lambda}) >\kappa$ to infer that for small enough $\eta>0$, 
\[{\Gamma_\lambda}(\hat\lambda + \eta(G(\theta_{\hat\lambda})-\kappa))
= \hat\lambda + \eta(G(\theta_{\hat\lambda})-\kappa).\]

We now proceed to prove the claim in (ii).
In the case where $\lambda^*=0$, 
we have $\Gamma_\lambda(\lambda^* + \eta(G(\theta_{\lambda^*})-\kappa))=0$ for any $\eta>0$, since $G(\theta_{\lambda^*}) < \kappa$.

Next, the case of $\lambda^* >0$ is not possible, and this can be argued as follows:
Suppose that $\lambda^* >0$. Then, for small enough $\eta>0$
\begin{align*} 
&{\Gamma_\lambda}(\lambda^* + \eta(G(\theta_{\lambda^*})-\kappa)) =
\lambda^* + \eta(G(\theta_{\lambda^*})-\kappa) >0, \textrm{ implying }\\ 
&\check\Gamma_\lambda(G(\theta_{\lambda^*})-\kappa) < 0,
\end{align*}
which leads to a contradiction since $\lambda^* \in \Z$.

For the final claim in (iii), observe that ${\Gamma_\lambda}(\lambda^* + \eta(G(\theta_{\lambda^*})-\kappa)) = \lambda_{\max}$ leading to $\check\Gamma_\lambda(G(\theta_{\hat\lambda})-\kappa)=0$, since $\lambda^* + \eta(G(\theta_{\lambda^*})-\kappa) > \lambda_{\max}$ for any $\eta>0$.
\end{proof}
Thus, when one projects the Lagrange multiplier onto a finite interval, the convergence guarantee is to a local saddle point of the Lagrangian, as before. However, the limiting policy may not necessarily satisfy the risk constraint due to finite projection for the Lagrange multiplier. The latter can be avoided in practice by choosing a large enough value for $\lambda_{\textrm{max}}$. 
However, the only way to theoretically guarantee avoidance of spurious limiting policies (i.e., ones that do not satisfy the risk constraint) is to eliminate the projection, i.e., to allow $\lambda$ to be any positive real number. 


\section{Bibliographic remarks}
For constrained MDPs, a textbook reference is \textcite{altman1999constrained}. In Section \ref{sec:template-constrained}, we invoke Theorem 3.8 from there to infer that there exists an optimal policy for the risk-constrained MDP in \eqref{eq:pb1}, whenever there is a policy that satisfies the risk constraint for this problem. A classic reference for the regularity conditions for ensuring the existence of a unique saddle point of the Lagrangian of the problem \eqref{eq:pb1} is \textcite{sion1958general}. Mean-variance optimization of MDPs has been shown to be NP-hard in \textcite{mannor2013algorithmic}.

Our template algorithm for the `risk as constraint' setting incorporates two-timescale stochastic approximation \citep{borkar1997stochastic,borkar2008stochastic}. Two-timescale algorithms are popular for solving the problem of control in the context of reinforcement learning, where they are usually referred to as actor-critic algorithms, cf. \textcite{konda1999actor,borkar2005actor,bhatnagar2009natural,bhatnagar2010actor,prashanth2016mlj}. Several simulation-based optimization algorithms also involve multiple timescales, see the textbook by \textcite{Bhatnagar13SR} for several examples. From the field of simulation optimization, the simultaneous perturbation method for gradient estimation is particularly relevant for solving risk-sensitive MDPs, when the underlying risk measure does not possess the structure to enable direct gradient estimation schemes such as the likelihood ratio method; see the case studies involving variance and CPT risk measures in Chapters \ref{sec:risk-algos} and \ref{sec:risk-algos-unconstrained} for concrete examples.

\textcite[Theorem 2 in Chapter 6]{borkar2008stochastic} provides a justification of the standard two-timescale viewpoint, i.e., the $\lambda$-recursion on the slower timescale sees  the policy parameter as almost equilibrated, while the $\theta$-recursion on the faster timescale sees the Lagrange multiple $\lambda$ as quasi-static. 
The Kushner-Clark Lemma \citep{kushner-clark} is a classic result that can be invoked to establish asymptotic convergence of stochastic approximation schemes. The proof of Theorem \ref{thm:lambda-convergence} follows by using arguments similar to those employed in the proof of Theorem 2 in Chapter 2 of \textcite{borkar2008stochastic}.
For the claim that the two-timescale algorithm in the constrained setting converges to a policy that satisfies the constraint in \eqref{eq:pb1}, one invokes the envelope theorem of mathematical economics \citep{mas1995microeconomic}. The reader is referred to \textcite{borkar2005actor} or \textcite{bhatnagar2010actor} for further details.   

The convergence guarantees in Section \ref{sec:convergence-constrained} for the template algorithm with risk as constraint are asymptotic in nature, and we have not provided any convergence rate results for this algorithm. Deriving such a rate result is challenging, as there are no rate
rate results for general two-timescale stochastic approximation schemes, barring a few exceptions that we note next. In \textcite{konda2004convergence}, the authors handle the case of linear recursions, and provide an asymptotic rate result through a central limit theorem (CLT)-type result. In \textcite{mokkadem2006convergence}, the authors extend this result to handle nonlinear recursions, and it is not clear if the assumptions for invoking this result are satisfied by the template algorithm that uses the update iterations  \eqref{eq:lambda-ascent}--\eqref{eq:theta_descent_det}. More recently, in \textcite{dalal2018finite,dalal2020tale}, the authors derive non-asymptotic bounds for two-timescale stochastic approximation, albeit with linear recursions.

\diff{The non-asymptotic analysis in the `risk as objective' setting is based on the randomized stochastic gradient algorithm proposed in \textcite{ghadimi2013stochastic}. In particular, the proof of the non-asymptotic bound in Theorem \ref{thm:biased_sp} follows by a completely parallel argument to the proof of Theorem 2.1 in \textcite{ghadimi2013stochastic}.

The theoretical guarantees in this chapter establish convergence of the risk-sensitive policy gradient algorithms to a stationary point, a standard notion often employed in the analysis of policy gradient algorithms. As noted in Section \ref{sec:sa}, one can avoid saddle points/local maxima (minima for maximization problems) and ensure convergence to a local minimum (resp. maximum) if the gradient estimator has sufficient noise in all directions; see \textcite{pemantle1990nonconvergence,brandiere1996algorithmes} for a precise set of conditions, and a textbook reference for the topic of avoidance of such undesirable stationary point ``traps'' is Section 4.3 of \textcite{borkar2008stochastic}. A recent result in this direction is \textcite{barakat2021stochastic}. 
If the gradient estimation noise is lacking, extraneous noise can be added in the policy gradient update, as in \eqref{eq:sg-noiseextra}. Such an approach has been explored in a non-RL context in \textcite{ge2015escaping,jin2017escape}. Recent work in the policy gradient literature has tried to go beyond stationary convergence, e.g., using second-order information and/or incorporating variance reduction, e.g., \textcite{papini2018,shen2019hessian,zhangK2020}; however, providing guarantees beyond stationary convergence and characterizing the optima is beyond the scope of our book, as such considerations are not specific to a risk-sensitive context but apply to any policy gradient algorithm. }

%
\chapter{MDPs with Risk as the Constraint}
\label{sec:risk-algos}
Recall that the main ingredients in each iteration $n$ of the risk-sensitive RL algorithm in the constrained setting are as follows:
\begin{enumerate}[label=(\Roman*)]
\item Simulation of the underlying MDP. For the case of non-perturbation-based approaches, such as the likelihood ratio method, one simulation with policy $\theta_n$ would suffice. On the other hand, for SPSA-based approaches, an additional simulation using a perturbed policy parameter would be necessary (see Section \ref{sec:spsa}).
\item Estimation of $\nabla J(\theta_n)$ and $\nabla G(\theta_n)$. These estimates are fed into the primal descent update for $\theta_n$.
\item Estimation of $G(\theta_n)$ using sample data. This estimate is used for dual ascent.
\item Estimation of $J(\theta_n)$ using sample data. This estimate is used for primal descent. Note that in the case of SPSA, we would require estimate of $J$ for the perturbed policy as well, while additional function estimates are not necessary using the likelihood ratio method.
\end{enumerate}
In the four special cases that we discuss in detail below, we shall address the items above in a variety of MDP contexts, under mean-variance, CVaR, and chance constraints.
In particular, Table \ref{tab:splcases} presents the combinations for the objective $J$ and risk measure $G$ in 
\eqref{eq:gen-constrained-opt}.
\begin{table}[h]
\caption{Risk-sensitive MDPs considered in this book.}
\label{tab:splcases}
\vspace*{-12pt}
\centering
\boxed{
\begin{tabular}{c|c|c|c}
 \toprule
& MDP type & objective $J$ & constraint $G$ \\\midrule
\hspace*{-10pt}
Case 1 & discounted cost & cumulative cost   &   overall variance   \\
&  &  (see \eqref{oe})  &   (see \eqref{eq:V1})   \\\midrule
\hspace*{-10pt}
Case 2 & average cost  & average cost  & per-period variance    \\
 &   & (see \eqref{eq:avg-cost}) & (see \eqref{eq:V})   \\\midrule
\hspace*{-10pt}
Case 3 & SSP  & total cost  & CVaR  \\
 &   &  (see \eqref{eq:ssp-cost}) & (see \eqref{eq:cvar-ssp}) \\\midrule
\hspace*{-10pt}
Case 4 & discounted cost & total/discounted  & chance \\
 & or SSP & cost &(see \eqref{eq:chance} )\\\bottomrule
\end{tabular}\\[0.5ex]}
\end{table}

Chapter \ref{sec:background} presented the necessary background material on TD algorithms and two widely used approaches for gradient estimation, which serve as building blocks for the four special cases in Table \ref{tab:splcases} described in the rest of this chapter. 
The presentation presumes the reader is familiar with the theory of risk-neutral RL. 
The references in the bibliographic remarks provide more details for the interested reader. 

\section{Case 1: Discounted-cost MDP + variance as risk}
\label{sec:discounted-MDP-algo}

\diff{We consider the following constrained problem: For a given $\kappa >0$ and initial state $x_0$ of the discounted-cost MDP,
\begin{align*}
& \min_\theta J(\theta,x_0)\quad  \quad\text{subject to} \quad\quad G(\theta,x_0)\leq\kappa,
\end{align*}
where $J(\theta,x)=\E\left[D(\theta,x)\right]$ and $G(\theta,x)=U(\theta,x)-J(\theta,x)^2$  are the expectation and variance of the cumulative cost r.v., respectively, with $D(\theta,x)$ denoting the discounted total cost, which was defined in \eqref{eq:discounted-cost}.}

\subsection*{Gradient of the Lagrangian}
\vspace*{-2pt}
Defining $L(\theta,\lambda) \triangleq J(\theta,x_0)+\lambda\big(G(\theta,x_0)-\kappa\big)$, the necessary gradients of the Lagrangian are given by
\begin{align*}
\nabla_\theta L(\theta,\lambda)&=\nabla J(\theta,x_0)+\lambda\nabla G(\theta,x_0)\\
&=\nabla J(\theta,x_0)+\lambda\left(\nabla U(\theta,x_0)-2J(\theta,x_0)\nabla J(\theta,x_0)\right),\\
\nabla_\lambda L(\theta, \lambda) &= G(\theta,x_0)-\kappa.
\end{align*}
The expressions for $\nabla J(\theta,x_0)$ and $\nabla U(\theta,x_0)$ require a counterpart of $U(\cdot)$ with initial state-action pair $(x,a)$ under policy $\theta$, defined by
\[
	W(\theta,x,a) \stackrel{\triangle}{=} \E\left[ \left.\left(\sum_{n=0}^\infty\gamma^n k(x_n,a_n) \right)^2 \right| x_0=x,\;a_0=a,\;\theta\right].
\]
Similar to $U$, the function $W$ also satisfies a fixed point equation:
\begin{align}
	W(\theta,x,a)&=k(x,a)^2+\gamma^2\sum_{y}P(y|x,a)U(\theta,y) \nonumber\\[-4pt]
	&\qquad+2\gamma k(x,a)\sum_{y}P(y|x,a)J(\theta,y).\nonumber
\end{align}
\vspace*{-12pt}
We now provide expressions for the gradients $\nabla J(\theta,x_0)$ and $\nabla U(\theta,x_0)$. 
\vspace*{-8pt}
\boxed{
\begin{lemma}
\vspace*{-8pt}
\begin{align}
\nabla J(\theta,x_0)
& =\sum_{x,a}\pi^\theta_\gamma(x,a|x_0)\nabla\log\mu^\theta(a|x)Q(\theta,x,a), \label{eg:gradJ}\\
\nabla U(\theta,x_0)
& =\sum_{x,a}\widetilde{\pi}^\theta_\gamma(x,a|x_0)\nabla\log\mu^\theta(a|x)W(\theta,x,a) \nonumber\\
& +2\gamma\sum_{x,a,y}\widetilde{\pi}^\theta_\gamma(x,a|x_0)P(y|x,a)k(x,a)\nabla J(\theta,y),~~~ \label{eq:gradU}  
\end{align}
\vspace*{-12pt}
\begin{align}
\mbox{where~}
Q(\theta,x,a)&=\E\left[\sum\limits_{n=0}^\infty\gamma^n k(x_n,a_n)\; \middle|\; x_0=x,\;a_0=a,\;\theta\right], \nonumber\\ 
\pi^\theta_\gamma(x,a|x_0)&=(1-\gamma)\sum_{n=0}^\infty\gamma^{n}\Prob{x_n=x|x_0;\theta}\mu^\theta(a|x), \nonumber \\
\widetilde{\pi}^\theta_\gamma(x,a|x_0)&=(1-\gamma^2)\sum_{n=0}^\infty\gamma^{2n}\Prob{x_n=x|x_0;\theta}\mu^\theta(a|x).
\end{align}
\end{lemma}}
Note that $Q(\theta,\cdot,\cdot)$ is the Q-value function associated with policy $\mu^\theta$, while $\pi^\theta_\gamma$ and $\widetilde{\pi}^\theta_\gamma$ are the respective $\gamma$ and $\gamma^2$-discounted visiting distributions of the state-action pair $(x,a)$ under policy $\mu^\theta$.

\vspace*{-3pt}
\begin{proof}
We derive the expression for $\nabla U(\theta,x_0)$, as the proof for the case of $\nabla J(\theta,x_0)$ is standard. Using $U(\theta,x) = \sum_a\mu^\theta(x|a)W(\theta,x,a)$, and differentiating w.r.t. $\theta$, we have
\vspace*{-3pt}
	\begin{align}
		\nabla U(\theta,x_0)
		&=\sum_a\nabla\mu^\theta(a|x_0)W(\theta,x_0,a)+\sum_a\mu^\theta(a|x_0)\nabla W(\theta,x_0,a) \nonumber \\
		&=\sum_a\nabla\mu^\theta(a|x_0)W(\theta,x_0,a)+\sum_a\mu^\theta(a|x_0)\nonumber\\
		& ~~~\times \nabla\Big[k(x_0,a)^2+\gamma^2\sum_{y}P(y|x_0,a)U(\theta,y) \nonumber\\
		&\qquad+2\gamma k(x_0,a)\sum_{y}P(y|x_0,a)J(\theta,y)\Big] \nonumber \\
		=\underbrace{\sum_a\nabla\mu^\theta(a|x_0)W(\theta,x_0,a)+2\gamma\sum_{a,y}\mu^\theta(a|x_0)k(x_0,a)P(y|x_0,a)\nabla J(\theta,y)}_{h(\theta,x_0)}   \hspace{-3.7in} \nonumber\\
		&~~~~+\gamma^2\sum_{a,y}\mu^\theta(a|x_0)P(y|x_0,a)\nabla U(\theta,y) \nonumber \\
		&=h(\theta,x_0)+\gamma^2\sum_x\Prob{x_1=x|x_0;\theta} \nabla U(\theta,x) \\ 
		&=h(\theta,x_0)+\gamma^2\sum_x\Prob{x_1=x|x_0;\theta} 
		 \hspace{-2pt} \Big[ h(\theta,x) + \gamma^2 \sum_x \cdots \Big] \hspace{-10pt}
	\end{align}
\vspace*{-4pt}
Repeated application of the above relationship yields 
	\begin{align*}
    \vspace*{-13pt}
		\nabla U(\theta,x_0) &= \sum_{n=0}^\infty\gamma^{2n}\sum_x\mathbb{P}(x_n=x|x_0;\theta)h(\theta,x) \nonumber \\
		&= \sum_{n=0}^\infty\gamma^{2n}\Big[\sum_{x,a}\Prob{x_n=x|x_0;\theta}\mu^\theta(a|x)\nabla\log\mu^\theta(a|x)W(\theta,x,a)\\
		&~~~+2\gamma\sum_{x,a,y}\mathbb{P}(x_n=x|x_0;\theta)\mu^\theta(a|x)k(x,a)P(y|x,a)\nabla J(\theta,y)\Big] \nonumber \\
		&= \frac{1}{1-\gamma^2}\Big[\sum_{x,a}\widetilde{\pi}_\gamma(x,a|x_0)\nabla\log\mu^\theta(a|x)W(\theta,x,a)
	\end{align*}
\hspace*{65pt}
$+2\gamma\sum_{x,a,y}\widetilde{\pi}_\gamma(x,a|x_0)k(x,a)P(y|x,a)\nabla J(\theta,y)\Big]. $
\end{proof}
\begin{figure}[h]
\centering
\tikzstyle{block} = [draw, fill=white, rectangle,
   minimum height=3em, minimum width=6em]
\tikzstyle{sum} = [draw, fill=white, circle, node distance=1cm]
\tikzstyle{input} = [coordinate]
\tikzstyle{output} = [coordinate]
\tikzstyle{pinstyle} = [pin edge={to-,thin,black}]
\scalebox{0.65}{\begin{tikzpicture}[auto, node distance=2cm]
\node (theta) {\large$\bm{\theta_n}$};
\node [sum, fill=blue!20,above right=0.8cm of theta, xshift=1cm] (perturb) {$\bm{+}$};
\node [sum,fill=red!20, below right=0.8cm of theta, xshift=1cm] (perturb1) {};
\node [above=0.5cm of perturb] (noise) {\large$\bm{\delta_n \Delta(n)}$};
\node [block,fill=blue!20, right=5cm of perturb,label=above:{\color{bleu2}\bf Prediction}, minimum height=5em,] (psim) {\makecell{\bf Estimate\\[1ex]  $\bm{J(\theta_n+\delta_n \Delta(n))}$\\[1ex]  $\bm{G(\theta_n+\delta_n \Delta(n))}$}}; 
\node [block,fill=red!20, right=5cm of perturb1] (sim) {\makecell{\bf Estimate\\[1ex] \bf for $\bm{J(\theta_n),G(\theta_n)}$}}; 
\node [block, fill=green!20,below right=2cm of psim,label=above:{\color{bleu2}\bf Control}, minimum height=8em, yshift=3cm,text width=3.6cm] (update) {\bf{Gradient descent}\\[2ex] \bf{using SPSA}};
\node [right=0.7cm of update] (thetanext) {$\bm{\theta_{n+1}}$};

\draw [->] (perturb) -- node[above] {\textbf{Obtain}} node[below] {$\bm{m_n}$ \textbf{length trajectory}}  (psim);
\draw [->] (perturb1) -- node[above] {\textbf{Obtain}} node[below] {$\bm{m_n}$ \textbf{length trajectory}}  (sim);
\draw [->] (noise) -- (perturb);
\draw [->] (psim) -- 
(update.150);
\draw [->] (sim) --  
(update.205);
\draw [->] (update) -- (thetanext);
\draw [->] (theta) --   (perturb);
\draw [->] (theta) --   (perturb1);
\end{tikzpicture}}
\caption{Overall flow of SPSA-based risk-sensitive policy gradient algorithm in a discounted cost MDP setting.}
\label{fig:algorithm-spsa-flow}
\end{figure}

Estimating $\nabla J(\theta,x_0)$ and $\nabla U(\theta,x_0)$ is challenging due to the following reasons:
\begin{enumerate}[label=\textbf{\Roman*})]
\item Two different sampling distributions are used for $\nabla J(\theta,x_0)$ and $\nabla U(\theta,x_0)$. In particular, the distributions $\pi^\theta_\gamma$ and $\widetilde{\pi}^\theta_\gamma$ involve factors $\gamma$ and $\gamma^2$, respectively. 
\item $\nabla J(\theta,y)$ appears in the second summation on the RHS of \eqref{eq:gradU}, and this makes the estimation task hard in practice, as one needs an estimate of the gradient of the value function $J(\theta,y)$ at every state $y$ of the MDP, and not just at the initial state $x_0$. 
\end{enumerate}

To overcome these issues, we use SPSA to estimate $\nabla J(\theta,x_0)$ and $\nabla U(\theta,x_0)$.
As illustrated in Figure~\ref{fig:algorithm-spsa-flow}, such an estimation scheme requires running two trajectories corresponding to policy parameters $\theta_n + \delta_n \Delta(n)$  (where $\delta_n$ and $\Delta(n)$ are described in Section \ref{sec:spsa}) and $\theta_n$. The samples from the trajectories would be used to estimate $J(\theta_n+ \delta_n \Delta(n)), J(\theta_n)$, $U(\theta_n+ \delta_n \Delta(n))$ and $U(\theta_n)$, which in turn help in forming the estimates of the gradient of $J(\theta,x_0)$ and $U(\theta,x_0)$  as follows: For $i=1,\ldots,\|\X\|,$
\begin{align}
\widehat\nabla_i  J(\theta_n,x_0)&= \dfrac{\widehat J(\theta_n+\delta_n\Delta(n),x_0) - \widehat J(\theta_n,x_0)}{\delta_n \Delta_i(n)}, \textrm{ and }\nonumber\\
\widehat\nabla_i  U(\theta_n,x_0)&= \dfrac{\widehat U(\theta_n+\delta_n\Delta(n),x_0) - \widehat U(\theta_n,x_0)}{\delta_n \Delta_i(n)}.
\label{eq:spsa-uj-grad}
\end{align}
In the above, $\widehat J(\theta,x_0)$ (resp. $\widehat U(\theta,x_0)$) denotes an estimate of $J(\theta,x_0)$ (resp. $U(\theta,x_0)$), for any $\theta\in\Theta$.

\subsection*{Policy evaluation using TD}
\diff{The task of estimating $J$ is straightforward, and the regular TD algorithm described earlier in Section \ref{sec:td} can be employed.
Notice that the policy parameter update would use an estimate of $\nabla_\theta L(\theta,\lambda)$ to perform an incremental update. In this case, this gradient features a factor of the form in the template algorithm \eqref{eq:theta_descent_det} that is the product $J(\theta,x_0)\nabla J(\theta,x_0)$, and to ensure independence, we use double sampling to form estimates $\widehat J$ and $\doublehat J$ of  $J(\theta,x_0)$. The first estimate would be employed in \eqref{eq:spsa-uj-grad}, while the second one would be used in the first term of the aforementioned product.}

\diff{
Recall that the value function $J$ satisfies a fixed-point equation. One could possibly combine both $J$ and $G$, or view the fixed-point equation for $J$ in \eqref{eq:J-Bellman} together with the equation \eqref{eq:G-fixedpoint} over $2|\X|$ variables. However, relying on the equation \eqref{eq:G-fixedpoint} for policy optimization is challenging owing to the fact that  variance lacks the monotonicity property. Monotonicity is required in classic policy improvement algorithms, and one cannot derive meaningful convergence guarantees in the absence of monotone operators.
Alternatively, one can derive a fixed-point equation for the squared value function $U$, where the operator underlying this equation is a contraction mapping, and the variance can then be estimated using $U$ and $J$, where estimation of the latter quantities is facilitated through TD-type learning algorithms. The following proposition presents the fixed-point equation for the squared value function. 
\boxed{
\begin{proposition}
The squared value function $U(\theta,x)$ satisfies
\begin{align}
U(\theta,x)&=\sum_a\mu^\theta(a|x) k(x,a)^2+\gamma^2\sum_{a,y}\mu^\theta(a|x)P^\theta(y|x,a)U(\theta,y)\nonumber\\
&\quad+2\gamma\sum_{a,y}\mu^\theta(a|x)P^\theta(y|x,a)k(x,a)J(\theta,y).  \label{eq:U-W-Bellman}
\end{align}
\end{proposition}}
\begin{proof}
	\begin{align*}
          U(\theta,x) &=\E\left[(D(\theta,x))^2\mid x_0=x\right] = \E\left[  \left(k(x,a_0) + \sum_{m=1}^\infty \gamma^m k(x_m,a_m)\right)^2\right]\\
          &= \E\left[  k(x,a_0)^2\right] + \E\left[ \left(\sum_{m=1}^\infty \gamma^m k(x_m,a_m)\right)^2 \mid x_0=x\right]\\
          &\qquad +2\gamma \E\left[ k(x,a_0) \times \left(\sum_{m=0}^\infty \gamma^m k(x_m,a_m)\right) \mid x_0=x\right]\\
          &= \sum_a\mu^\theta(a|x) k(x,a)^2+\gamma^2\sum_{a,y}\mu^\theta(a|x)P^\theta(y|x,a)U(\theta,y)\\
          &\quad+2\gamma\sum_{a,y}\mu^\theta(a|x)P^\theta(y|x,a)k(x,a)J(\theta,y).
	\end{align*}
\end{proof}

Let $\bm{k}^\theta$ be a $|\X|$ vector of single-stage costs for each state, and $\bm{K}^\theta$ be a $|\X|\times|\X|$ matrix with the entries $\sum_a\mu^\theta(a|x) k(x,a)$ for each state $x$ along the diagonal and zeroes elsewhere.
For any $(\bm{J},\bm{U})\in\R^{2|\X|}$, where $\bm{J}$ and $\bm{U}$ denote the first and last $|\X|$ entries, respectively, define
\begin{align}
	T^\theta (\bm{J},\bm{U}) &= \begin{bmatrix}
		T_1^\theta (\bm{J})\\
		T_2^\theta (\bm{J},\bm{U})
	\end{bmatrix}, \text{ where }\\
	T_1^\theta (\bm{J})&=\bm{k}^\theta+\gamma P^\theta \bm{J}, \textrm{ and ~~} \nonumber\\
	T_2^\theta (\bm{J},\bm{U})&=\bm{K}^\theta \bm{k}^\theta+2\gamma \bm{K}^\theta P^\theta \bm{J}+\gamma^2 P^\theta \bm{U}. \label{eq:T-disc}
\end{align}
Using the notation above, the fixed-point equations for $J$ in \eqref{eq:J-Bellman} and $U$ in \eqref{eq:U-W-Bellman} can be combined together as 
\begin{equation}
\label{eq:J-U-fixed-point}
(\bm{J^\theta},\bm{U^\theta}) = T^\theta (\bm{J^\theta},\bm{U^\theta}),
\end{equation}
where $\bm{J^\theta} = [J(\theta,x)]_{x\in\X}$ and $\bm{U^\theta} = [U(\theta,x)]_{x\in\X}$.

Let $d^\theta$ denote the stationary distribution of the Markov chain underlying policy $\theta$. We shall assume that such a distribution exists --- an assumption that is easily satisfied for unichain policies (i.e., the underlying Markov chain is irreducible and positive recurrent).
We now focus on establishing that the operator $T^\theta$ is a contraction mapping w.r.t. a weighted norm, for any policy $\theta$.
The weighted norm, denoted by $\| \cdot,\cdot\|_\nu$, is defined as follows:
for any $(J,U) \in \R^{2|\X|}$, 
\[\| J,U \|_\nu = \nu \| J \|_{d^\theta} + (1-\nu) \| U \|_{d^\theta},\] 
where $\| x \|_{d^\theta} = \sqrt{ \sum_{i=1}^{|\X|} d^\theta(i) x_i^2}$ for any $x \in \R^{|\X|}$.
 
\boxed{\begin{proposition}
\vspace*{-8pt}
	\label{lemma:td-contract}
	There exists a $\nu \in (0,1)$ and $\bar \gamma <1$ such that 
	\[\left\| T^\theta (J,U) - T^\theta(\bar J, \bar U) \right\|_{\nu} \le \bar\gamma \left\| (J,U) - (\bar J,\bar U) \right\|_{\nu}, \forall J, \bar J, U,\bar U \in \R^{|\X|}.\]
\end{proposition}}
\begin{proof}
	First, we show that $T_1^\theta$ is a contraction mapping. This can be inferred as follows: For any $y, \bar y \in\R^{2|\X|}$,
	\begin{align*}
		||P^\theta J||_{d^\theta}^{2}&=\sum_{i=1}^{|\X|}d^\theta(i)(\sum_{j=1}^{|\X|}P^\theta(j | i)J(j))^{2} 
		\leq\sum_{i=1}^{|\X|}d^\theta(i)\sum_{j=1}^{|\X|}(P^\theta(j | i)J(j)^{2}) \tag{Jensen's inequality}\\
		&=\sum_{j=1}^{|\X|}(\sum_{i=1}^{|\X|}d^\theta(i)P^\theta(j | i))(J(j))^{2} 
		=\sum_{j=1}^{|\X|}d^\theta(j)(J(j))^{2} \tag{as $d^\theta=d^\theta P^\theta $}=||J||_{d^\theta}^{2}. 
	\end{align*}
    Using $||P^\theta J||_{d^\theta} \le ||J||_{d^\theta}$, we have
	\[\| T_1^\theta (J) - T_1^\theta (\bar J) \|_{d^\theta} = \gamma \| P^\theta J - P^\theta \bar J \|_{d^\theta}\le \gamma\| J - \bar J \|_{d^\theta}.\]
	
	Now, for any $J, \bar J, U, \bar U \in \R^{|\X|}$, we have
	\begin{align}
		&\| T_2^\theta (J,U) -  T_2^\theta (\bar J,\bar U)\|_{d^\theta}\nonumber\\
		= & \|2\gamma   \bm{K}^\theta P^\theta J - 2\gamma  \bm{K}^\theta P^\theta \bar J + \gamma^2  P^\theta U - \gamma^2  P^\theta \bar U  \|_{d^\theta}\nonumber\\
		\le & 2\gamma \|  \bm{K}^\theta P^\theta J -  \bm{K}^\theta P^\theta \bar J \|_{d^\theta} + \gamma^2 \|U -  \bar U  \|_{d^\theta}\nonumber\\
		\le & \gamma C_1 \|J - \bar J \|_{d^\theta} + \gamma^2 \|U -  \bar U  \|_{d^\theta},\label{eq:piu}
	\end{align}
	for some $C_1 < \infty$.
	The first inequality above follows by using $||P^\theta J||_{d^\theta} \le ||J||_{d^\theta}$, while the second inequality follows by using the equivalence of norms.
	
	Now, setting $\nu = \dfrac{\gamma C_1}{\epsilon + \gamma C_1}$, with $\epsilon$ satisfying $\gamma + \epsilon < 1$, we have
	\begin{align*}
		&\|T^\theta (J,U) - T^\theta (\bar J,\bar U)\|_{\nu}\\
		= & \nu \| T_1^\theta J - T_1^\theta \bar J \|_{d^\theta} + (1-\nu) \| T_2^\theta U - T_2^\theta \bar U \|_{d^\theta}\\
		\le & \nu \gamma \|J - \bar J\|_{d^\theta} +  (1- \nu) \gamma C_1 \|J - \bar J \|_{d^\theta} + (1-\nu) \gamma^2 \|U -  \bar U  \|_{d^\theta}\\
		\le & \nu (\gamma + \epsilon) \|J - \bar J \|_{d^\theta}  + (1-\nu) \gamma \|U -  \bar U  \|_{d^\theta}\\
		\le &  (\gamma + \epsilon) \|(J,U) - (\bar J,\bar U) \|_{\nu}.
	\end{align*}
	The claim follows by setting $\bar \gamma = \gamma + \epsilon$.
\end{proof}
}

\diff{
We now have what we need to estimate the squared value function $U$ using a TD-type algorithm,
as $U$ satisfies the fixed-point equation \eqref{eq:U-W-Bellman}, and the $T^\theta$ operator underlying this equation is well behaved, in the sense that Proposition \ref{lemma:td-contract} establishes that it leads to a contraction mapping that is amenable for stochastic approximation, so the results of Sections \ref{sec:contractive-SA} and \ref{sec:td} are applicable.
On the other hand, note that estimating variance directly would not help, because the corresponding underlying operator is not monotone.
}

From the foregoing, we have the following TD-type update for estimating $U$:
\begin{align}
& U_{n+1}(x) = U_n(x) + \zeta(\nu(x,n))\indic{x_n=x}\left(k(x_n,a_n)^2 \right.\nonumber\\ 
 &\quad\left.+2\gamma k(x_n,a_n) J_n(x_{n+1}) \!+\! \gamma^2 U_n(x_{n+1}) \!-\! U_n(x_n)\right),\label{eq:u-td}
\end{align}
where $\nu(x,n) = \sum\limits_{m=0}^n \indic{x_m = x}$ and $x_{n+1}$ is a r.v. sampled from $P(\cdot\mid x_n,a_n)$.
Notice that the factor $J$ goes into the fixed-point equation for $U$, and hence, the TD algorithm for $U$ has to employ the TD-based estimate of $J$ for estimating $U$.

Algorithm \ref{alg:discounted-PG} presents the pseudocode for the risk-sensitive policy-gradient algorithm for the discounted-cost setting.
In a nutshell, this algorithm uses multi-timescale stochastic approximation to perform the following tasks:
(i) run the TD algorithm on the fastest timescale to estimate both $J$ and $U$;
(ii) use an SPSA-based gradient descent scheme on the intermediate timescale  for solving the primal problem in \eqref{eq:gen-constrained-opt};  and 
(iii) perform dual ascent on the Lagrange multiplier using the sample variance constraint (using the estimate of $U$) on the slowest timescale. 
The latter two-timescale updates follow the template provided in Section \ref{sec:rpg-template}.

\begin{algorithm}
\SetKwInOut{Input}{Input}\SetKwInOut{Output}{Output}
\Input{initial parameter $\theta_0 \in \Theta$, 
perturbation constants $\{\delta_n\}$, trajectory lengths $\{m_n\}$, step sizes $\{\zeta_1(n)\}$, $\{\zeta_2(n)\}$, projection operators $\Gamma$ and $\Gamma_\lambda$, \# iterations $M$.}
\For{$n\leftarrow 0$ \KwTo $M-1$}{
Set $\Delta(n)$ using symmetric $\pm 1$-valued Bernoulli distribution\;
  \For{$m\leftarrow 0$ \KwTo $m_n-1$}{
  \tcc{Unperturbed policy simulation}
    Use the policy $\mu^{\theta_n}$ to draw action $a_m\sim\mu^{\theta_n}\left(\cdot\mid x_m\right)$\;
    Observe next state $x_{m+1}$ and cost $k(x_m,a_m)$\;
    Use \eqref{eq:td-fullstate} and \eqref{eq:u-td} to form estimates $\widehat J(\theta_n,x_0)$ and $\widehat G(\theta_n,x_0)$ of $J(\theta_n)$ and $G(\theta_n)$, respectively\;
    \diff{Use another independent sample trajectory to form the estimate $\doublehat J(\theta_n,x_0)$ of $J(\theta_n)$}\;
        \tcc{Perturbed policy simulation}
        Use the policy $\mu^{\theta_n+\delta_n\Delta(n)}$ to generate the state $x_m^+$, draw action $a_m^+\sim\mu^{\theta_n+\delta_n\Delta(n)}\left(\cdot\mid x_m^+\right)$\;
    Observe next state $x^+_{m+1}$ and cost  $k(x^+_m,a^+_m)$\;
    Use \eqref{eq:td-fullstate} and \eqref{eq:u-td} to form estimates $\widehat J(\theta_n+\delta_n\Delta(n),x_0)$ and $\widehat G(\theta_n+\delta_n\Delta(n),x_0)$ of $J(\theta_n+\delta_n\Delta(n))$ and 
    $G(\theta_n+\delta_n\Delta(n))$, respectively\; 
    }
Gradient estimate for the objective: $\widehat\nabla_i  J(\theta_n,x_0)= \dfrac{\widehat J(\theta_n+\delta_n\Delta(n),x_0) - \widehat J(\theta_n,x_0)}{\delta_n \Delta_i(n)}$\;
Gradient estimate for the constraint: $\widehat\nabla_i U(\theta_n,x_0)= \dfrac{\widehat U(\theta_n+\delta_n\Delta(n),x_0) - \widehat U(\theta_n,x_0)}{\delta_n \Delta_i(n)}$\;
    \tcc{Policy update: Gradient descent using SPSA}
\diff{$\theta_{n+1} = \Gamma\bigg[\theta_n - \zeta_2(n)  \bigg(\widehat\nabla J(\theta_n,x_0) + \lambda_n \left(\widehat \nabla U(\theta_n,x_0)-2\doublehat J(\theta_n,x_0)\widehat \nabla  J(\theta_n,x_0)\right)\bigg)\bigg]$}\;
    \tcc{Lagrange multiplier update: Dual ascent}
\diff{$\lambda_{n+1} = \Gamma_\lambda\big[\lambda_n + \zeta_1(n) \left(\widehat U(\theta_n,x_0)-2\doublehat J(\theta_n,x_0) - \kappa\right)\big]$}\;
 }
 \Output{Policy $\theta_M$} 
\caption{Policy gradient algorithm under variance as a risk measure in a discounted-cost MDP setting}
\label{alg:discounted-PG}
\end{algorithm}

%

\diff{
\subsection*{On the batch size $m_n$}

\tikzstyle{block} = [draw, fill=white, rectangle,minimum height=3em, minimum width=6em]

\begin{figure}
	\centering
	\begin{tabular}{cc}
		\subfigure[Simulation optimization setting]{
			\tabl{c}{\scalebox{0.75}{\begin{tikzpicture}
						\node (theta) {$\boldsymbol{\theta}$};
						\node [block, fill=blue!20,right=0.6cm of theta,align=center] (sample) {\makecell{\textbf{Measurement}\\\textbf{ Oracle}}}; 
						\node [right=0.6cm of sample] (end) {$\boldsymbol{\mathbf{f(\theta) + \xi}}$};
						\node [ above right= 0.6cm of end] (bias) {\textbf{Zero mean}};
						\draw [->] (theta) --  (sample);
						\draw [->] (sample) -- (end);
						\path [darkgreen,->] (bias) edge [bend left] (end);
				\end{tikzpicture}}\\[1ex]}
		}
		\\
		\subfigure[Typical RL setting]{
			\tabl{c}{\scalebox{0.75}{\begin{tikzpicture}
						\node (theta) {$\boldsymbol{\theta, \epsilon}$};
						\node [block, fill=blue!20,right=0.6cm of theta,align=center] (sample) {\makecell{\textbf{TD-based Estimator}}}; 
						\node [right=0.6cm of sample] (end) {$\boldsymbol{\mathbf{f(\theta) + \epsilon}}$};
						\node [ above right= 0.6cm of end] (bias) {\textbf{Controlled bias}};
						\draw [->] (theta) --  (sample);
						\draw [->] (sample) -- (end);
						\path [red,->] (bias) edge [bend left] (end);
				\end{tikzpicture}}\\[1ex]}
		}
	\end{tabular}
	\caption{Illustration of the difference between classic simulation optimization and optimization of the variance risk measure in an RL setting.
	 In the latter setting, the error $\epsilon$ in function estimates can be \textit{controlled} and made very low at the cost of additional simulations. }
	\label{fig:opt-diff}
\end{figure}

To understand the challenge in choosing an appropriate batch size $m_n$ for policy evaluation in Step 2 of Algorithm \ref{alg:discounted-PG}, so that the overall algorithm converges, consider a simpler setting of optimizing a smooth function $f$, i.e., 
\begin{align}
    \textrm{find ~} \theta^* = \argmin_{\theta\in \Theta} f(\theta),\label{eq:opt-pb-f}
\end{align}
where $\Theta$ is a convex and compact subset of $\R^d$. In a classic stochastic optimization setting,  one has an oracle that supplies noisy function measurements, but the noise is usually zero mean. On the other hand, in a typical RL setting, the function $f$ that has to be estimated from sample trajectories is the value $J(\theta)$ associated with a given policy $\theta$, as illustrated in Figure \ref{fig:opt-diff}. For the sake of simplicity, we drop the dependence on the parameter $\theta$, and instead, study the value estimation problem. Subsequently, when we analyze the policy gradient scheme in Algorithm \ref{alg:discounted-PG} for CPT-value optimization, we shall make the dependence on the policy parameter explicit.

For a given policy with a fixed start state $x_0$, let $m$ denote the length of the sample trajectory used to form an estimate $J_m$, using \eqref{eq:td-fullstate}, of the value $J(x_0)$.   
One can derive a bound on the estimation error $\left|J_m - J(x_0)\right|$, and such a bound would aid the proof of asymptotic convergence of the risk-sensitive policy gradient in Algorithm \ref{alg:discounted-PG}. 
An informal statement of such a finite-time bound for TD is as follows:
\textit{ Using a step size $\zeta(n)=c/n$ with a suitable choice for the constant $c$, 
\begin{align}
	\E \left\|\widehat J_m - J(x_0) \right\|  = O\left(\frac{1}{\sqrt{m}}\right).\label{eq:td-expec-bd}
\end{align}
}
We avoid a detailed discussion of the derivation of such a bound, as it is quite technical, and deviates from the focus of optimizing risk in an RL setting. However, in passing, we note that the step size $c$ in the bound above would require information about the underlying transition dynamics, and such a problematic dependence can be avoided by using Polyak-Ruppert iterate averaging, where one employs a bigger step size $c/n^\alpha$, with $\alpha \in (1/2,1)$ together with averaging of the iterates. Such an approach would result in a bound of the order $O\left(\frac{1}{m^{\alpha/2}}\right)$.

In the following discussion, we shall use $f$ to denote the smooth objective function that we want to minimize. 
From the foregoing, it is apparent that we have a setting where $f$ is not perfectly observable, and instead, one can obtain biased measurements of $f$ at any input parameter $\theta$. Choosing larger values of the batch size $m$ leads to an increase in the accuracy of the function measurement. In particular, from \eqref{eq:td-expec-bd}, the estimation bias is of order $O(\frac{1}{\sqrt{m}})$.
Figure \ref{fig:opt-diff} illustrates this difference in estimation between a classic optimization setting, and a typical RL setting, where the policy evaluation is performed for estimation of the value of a given policy.

A stochastic gradient-descent scheme to solve the problem defined in \eqref{eq:opt-pb-f} would update as follows:
\begin{align}
	\theta_{n+1} = \Gamma\left(\theta_n - \gamma_n  \widehat \nabla f(\theta_n)\right),
	\label{eq:theta-update}
\end{align}
where  $\{\gamma_n\}$ is a step-size sequence that satisfies standard stochastic approximation conditions,
$\Gamma=\left(\Gamma_{1},\ldots,\Gamma_{d}\right)$ is an operator that ensures that the update \eqref{eq:theta-update} stays bounded within the compact and convex set $\Theta$, and $\widehat \nabla f(\theta_n)$ is an estimate of the gradient of $f$ at $\theta_n$. 

Suppose that the gradient estimate $\widehat \nabla f(\theta_n)$ in \eqref{eq:theta-update} is formed using SPSA, as described in Section \ref{sec:spsa}, i.e., 
\[\widehat \nabla_i f(\theta_n)=\frac{\widehat f(\theta_n +\delta_n\Delta(n)) - \widehat f(\theta_n)}{\delta_n\Delta_i(n)},\]
where $\widehat f(\theta)$ denotes the estimate of $f(\theta)$, when the underlying parameter is $\theta$. Suppose that the estimation scheme returns $\widehat f(\theta_n) = f(\theta_n) + \varphi_{n}^\theta$, where $\varphi_{n}^\theta$ denotes the error in estimating the objective $f$ using $m_n$ function measurements. 
For the sake of this discussion, suppose that the estimation error vanishes at the rate $\frac{1}{m_n^{1/2}}$.
We first rewrite the update rule in \eqref{eq:theta-update} as follows:
\begin{align*}
	\theta^{i}_{n+1}  =  \Gamma_{i}\bigg( \theta^{i}_n -  \gamma_n \Big( \frac{f(\theta_n +\delta_n\Delta(n)) - f(\theta_n)}{\delta_n\Delta_i(n)} + \kappa_n\Big)\bigg),
\end{align*}
where $\kappa_n = \frac{(\varphi_n^{\theta_n +\delta_n\Delta(n)} - \varphi_n^{\theta_n})}{\delta_n\Delta_i(n)}$.
Let $\zeta_n = \sum_{l = 0}^{n} \gamma_l \kappa_{l}$. Then, a critical requirement that allows us to ignore the estimation error term $\zeta_n$ is the following condition: 
\begin{align}
	\sup_{l\ge0} \left (\zeta_{n+l} - \zeta_n \right) \rightarrow 0 \text{ as } n\rightarrow\infty.
	\label{eq:ns1}
\end{align}
Notice that the estimation error in $\zeta_n$ is a function of number of samples $m_n$ used for estimating the objective value, and it is obviously necessary to increase the number of samples $m_n$ so that the bias vanishes asymptotically. 
In addition to the usual conditions on the step-size sequence and perturbation constant $\delta_n$, one possible choice for $m_n$ that ensures that the bias in the gradient estimate vanishes and the overall algorithm converges is the following:
$\frac{1}{\sqrt{m_n}\delta_n}\rightarrow 0$. 


\begin{remark}
A similar observation holds even for the case where the function $f$ is the squared value function $J$. In this case, the estimation scheme is TD-learning, and order $\frac{1}{\sqrt{m}}$ bound on the root mean-squared error of TD-learning can be derived. In this case, the condition on $m_n$ for ensuring convergence of the overall stochastic gradient scheme would be $\frac{1}{m_n^{1/2}\delta_n}\rightarrow 0$.
Finally, similar considerations on the trajectory lengths hold for the purpose of CVaR estimation, as well as estimation of CPT-value. In both cases, the estimation procedure (see Algorithms  \ref{alg:discounted-PG} and \ref{alg:cvar-PG}) is asymptotically unbiased, but one does not have the luxury of having a very long run of the policy evaluation procedure, considering that the outer loop of incremental policy update needs to perform policy evaluation often. 
\end{remark}
}


\begin{remark}(Extension to incorporate function approximation)
One could parameterize both $J$ and $U$ using linear function approximation and then employ TD-type schemes for policy evaluation. 
Notice that both $J(\cdot)$ and $U(\cdot)$ need to be evaluated for the perturbed policies. Let $J(x)\approx v\tr\phi_v(x)$ and $U(x)\approx u\tr\phi_u(x)$ be the linear approximations to $J$ and $U$, respectively, with features $\phi_v(\cdot)$ and $\phi_u(\cdot)$ from low-dimensional spaces. 
It can be shown that an appropriate operator can be defined for $U$ using the above equation and an operator that projects orthogonally onto the linear space $\{ \Phi_u u \mid u \in \R^d\}$. Such a projected Bellman operator turns out to be a contraction mapping, and hence, a TD-type scheme can be arrived at, along the lines of that for the regular cost $J$. 
\end{remark}

\section{Case 2: Average-cost MDP + variance as risk}
\label{sec:avg-MDP-algo}
We consider the following constrained optimization problem for average cost MDPs: 
\begin{align*}
&\min_\theta J(\theta)\quad\quad \text{subject to} \quad\quad G(\theta)\leq\kappa,
\end{align*}
where $J(\theta)$ is the long-run average cost and $G(\theta)$ is the variance, as defined in Sections  \ref{sec:mdp-average} and \ref{sec:risk-average}.

\subsection*{Gradient of the Lagrangian}
Letting $L(\theta,\lambda) \triangleq J(\theta)+\lambda\big(G(\theta)-\kappa\big),$
and noting that $\nabla G(\theta)=\nabla\eta(\theta)-2J(\theta)\nabla J(\theta)$, it is apparent that   $\nabla J(\theta)$ and $\nabla\eta(\theta)$ are enough to calculate the necessary gradients of the Lagrangian. Let $U^\theta$ and $W^\theta$ denote the differential value and action-value functions associated with the squared cost under policy $\mu^\theta$, respectively. These two quantities satisfy the following Poisson equations:
\begin{align*}
\eta(\theta)+U(\theta,x) &= \sum_a\mu^\theta(a|x)\big[k(x,a)^2+\sum_{y}P(y|x,a)U(\theta,y)\big],  \\
\eta(\theta)+W(\theta,x,a) &= k(x,a)^2+\sum_{y}P(y|x,a)U(\theta,y).
\end{align*}
As mentioned earlier, we consider finite state-action space MDPs, which together with an irreducibility assumption implies the existence of a stationary distribution for the Markov chain underlying any policy $\theta$. Denote by $d^\theta(x)$ and $\pi^\theta(x,a)=d^\theta(x)\mu^\theta(a|x)$, the stationary distribution of state $x$ and state-action pair $(x,a)$ under policy $\mu^\theta$, respectively.  
We now present the gradients of $J(\theta)$ and $\eta(\theta)$ below.
\boxed{
\vspace*{-10pt}
\begin{align}
\nabla J(\theta)&=\sum_{x,a}\pi^\theta(x,a)\nabla\log\mu^\theta(a|x)Q(\theta, x,a), \label{eq:grad-rho}\\
\nabla\eta(\theta)&=\sum_{x,a}\pi^\theta(x,a)\nabla\log\mu^\theta(a|x)W(\theta,x,a),
\label{eq:grad-eta}
\end{align}}
where $Q(\theta,x,a)=\sum_{n=0}^\infty\E\big[k(x_n,a_n)-J(\theta)\mid x_0=x,a_0=a,\mu\big]$, with actions $a_n \sim \mu^\theta\left(\cdot\mid x_n\right)$.

The above relationships follow from parameterizing the policies, and hence, the gradient of the transition probabilities can be estimated from the policy alone. This is the well-known policy gradient technique  that makes it amenable for estimating the gradient of a performance measure in MDPs, since the values of the transition probabilities are not required and one can work with policies and simulated transitions from the MDP.

An important observation concerning $\nabla J(\theta)$ is that   
any function $b:\X\rightarrow\R$ can be added or subtracted to $Q(\theta,x,a)$ on the RHS of \eqref{eq:grad-rho}, and the resulting summation stays as $\nabla J(\theta)$. In a risk-neutral setting, a popular choice is to replace $Q(\theta,x,a)$ with the advantage function $A(\theta, x,a)=Q(\theta,x,a)-V(\theta,x)$. 

\diff{\subsection*{Policy evaluation using TD}
In a typical RL setting, $\nabla J(\theta)$ has to be estimated, and from the discussion before, this implies estimation of the advantage function using samples -- TD-learning is a straightforward choice for this task. 
Using the expression on the RHS of \eqref{eq:grad-rho}, one can arrive at a decrement factor for the policy update as follows:  substitute a TD-based empirical approximation to the advantage function, calculate the likelihood ratio $\nabla\log\mu(\cdot)$, and perform a gradient descent using the product of the advantage estimate with the likelihood ratio, and arrive at an empirical approximation to the RHS of \eqref{eq:grad-rho} with the advantage function $A$ instead of $Q$ there.
The TD algorithm-based estimate for the value function is given below.
\begin{align}
	\delta_n&=k(x_n, a_n) - J_{n+1}+V_n(x_{n+1}) - V_n(x_n),\nonumber\\
	V_{n+1}&=V_n + \zeta_3(n) \delta_n,
\end{align}
 The idea described above, i.e., to use the advantage function in place of $Q$, can be used for the case of $\nabla\eta(\theta)$ as well, with the advantage function variant $B(\theta,x,a)=W(\theta,x,a)-U(x;\theta)$ on the RHS of~\eqref{eq:grad-eta}. 
The TD algorithm-based estimate for the squared value function is given below.
\begin{align}
	\epsilon_n &=k(x_n, a_n)^2-\eta_{n+1}+U_n(x_{n+1}) - U_n(x_n),\nonumber\\
	U_{n+1}&= U_n + \zeta_3(n) \epsilon_n.
\end{align}

The pseudocode of the overall algorithm in the average reward setting is given in Algorithm \ref{alg:average-PG}. As in the discounted case discussed in the previous section, an extra sample trajectory is needed to form an independent estimate $\widehat J_n$, so that the product $\widehat J_n \delta_n \psi_n$ that we have in the policy parameter update can be separated after taking expectations to obtain $J(\theta_n) \nabla J(\theta_n)$ in the convergence analysis.
}
\begin{algorithm}
\SetKwInOut{Input}{Input}\SetKwInOut{Output}{Output}
\Input{initial parameter $\theta_0 \in \Theta$, where $\Theta$ is a compact and convex subset of $\R^d$,  step sizes $\{\zeta_1(n)\}$, $\{\zeta_2(n)\}$, $\{\zeta_3(n)\}$, $\{\zeta_4(n)\}$, projection operators $\Gamma$ and $\Gamma_\lambda$, ~~~~ \#~iterations $M\gg 1$.}
\For{$n\leftarrow 0$ \KwTo $M-1$}{
    Draw action $a_m \sim \mu^{\theta_n}(\cdot|x_m)$, observe next state $x_{m+1}$ and cost $k(x_m,a_m)$\;
\tcc{Estimate for average cost}
    $
J_{n+1}=\big(1-\zeta_4(n)\big)J_n+\zeta_4(n)k(x_n, a_n)
$\;
\diff{\tcc{Another estimate for average cost}
    Draw action $\hat a_m \sim \mu^{\theta_n}(\cdot|\hat x_m)$, observe next state $\hat x_{m+1}$ and cost $k(\hat x_m,\hat a_m)$\;
$
\widehat J_{n+1}=\big(1-\zeta_4(n)\big)\widehat J_n+\zeta_4(n)k(\hat x_n,\hat a_n)
$\;}
\tcc{Estimate for average squared cost}
$ \eta_{n+1}=\big(1-\zeta_4(n)\big)\eta_n +\zeta_4(n)k(x_n, a_n)^2      
 $   \;
 \tcc{TD estimate for the value function}
$ \delta_n=k(x_n, a_n) - J_{n+1}+V_n(x_{n+1}) - V_n(x_n)$ \;
$V_{n+1}=V_n + \zeta_3(n) \delta_n$\;
   \tcc{TD estimate for the squared value function}
$\epsilon_n =k(x_n, a_n)^2-\eta_{n+1}+U_n(x_{n+1}) - U_n(x_n)$\;
$U_{n+1}= U_n + \zeta_3(n) \epsilon_n$\;
Set $\psi_n=\nabla\log\mu^{\theta_n}(a_n|x_n)$\tcp*{Likelihood ratio}    
    \tcc{Policy update}
\diff{    $ 
\theta_{n+1}=\Gamma\Big(\theta_n-\zeta_2(n)\big(-\delta_n\psi_n+\lambda_n(\epsilon_n\psi_n-2 \widehat J_{n+1}\delta_n\psi_n)\big)\Big)$}\;
\tcc{Lagrange multiplier update}    
$    \lambda_{n+1}=\Gamma_\lambda\Big(\lambda_n+\zeta_1(n)(\eta_{n+1}-J_{n+1}^2-\kappa)\Big) $  \;
 }
 \Output{Policy $\theta_M$} 
\caption{Policy gradient algorithm under variance as a risk measure in an average cost MDP setting}
\label{alg:average-PG}
\end{algorithm}

In addition to the step-size requirements in (A4), we require that $\zeta_2(n) = o\big(\zeta_3(n)\big)$ and $\zeta_4(n)$ is a constant multiple of $\zeta_3(n)$. Such choices ensure that the TD-critic and average cost updates are on the fastest timescale, the policy update is on an intermediate timescale, and the Lagrange multiplier update is on the slowest timescale. 

\begin{remark}
The variance notion employed in this section involved measuring the deviations of the single-stage cost from its average. As we demonstrated in Algorithm \ref{alg:average-PG}, the per-period variance as a risk measure lends itself to policy gradient techniques well, since the likelihood ratio method can be employed to solve the risk-constrained problem. In contrast, the variance notion in the discounted cost setting involved the variance of the cumulative discounted cost, i.e., the (overall) variance of the underlying r.v. and not the per-period one. Such a measure is hard to optimize (see discussion below \eqref{eq:gradU}), though SPSA could be employed. The flip side to the latter approach is that we do not exploit the structure of the underlying problem in forming the gradient estimates, e.g., using the likelihood ratio method. More importantly, SPSA requires simulation of two independent trajectories (corresponding to unperturbed and perturbed policy parameters), and this may not be feasible in many practical applications. 

One could consider swapping the risk measures of discounted and average cost settings, i.e., employ per-period variance in a discounted cost MDP, and overall variance in the average cost MDP. Leaving the question of which is the best risk measure for a given MDP aside, we believe that such a swap of risk measures would make solving the average cost problem difficult, and discounted cost problem easy in comparison. 
\end{remark}

\vspace*{-4pt}
\begin{remark}
 As in the discounted setting, incorporating function approximation for the functions $J$ and $U$ is straightforward, and we omit the details. 
\end{remark}

\section{Case 3: Stochastic shortest path + CVaR as risk}
\label{sec:cvar-MDP-algo}
\vspace*{-8pt}

We again consider the following constrained optimization problem:
\diff{For a given $\kappa >0$ and initial state $x_0$ of the SSP MDP,}
\begin{align*}
& \min_\theta J(\theta,x_0)\quad  \quad\text{subject to} \quad\quad G(\theta,x_0)\leq\kappa,
\end{align*}
where $J(\theta,x_0)$ and $G(\theta,x_0)$  are the expectation and $\text{CVaR}_{\beta}, \beta \in (0,1),$ of the total cost r.v. $D(\theta, x_0)$, respectively (see Sections \ref{sec:mdp-ssp} and \ref{sec:risk-cvar}).

\subsection*{Gradient of the Lagrangian}
\vspace*{-4pt}

With the Lagrangian $L(\theta,\lambda) \triangleq J(\theta,x_0)+\lambda\big(G(\theta,x_0)-\kappa\big)$, the necessary gradients for solving the constrained problem above are $\nabla J(\theta,x_0)$ and $\nabla \text{CVaR}_{\beta}(D(\theta, x_0))$. Using the likelihood ratio method,  
the first gradient is obtained as follows:
\begin{align*}
 \nabla J(\theta,x_0) \!=\! \E\!\left[ \left[\left.\sum\limits_{n=0}^{\tau-1} k(x_n,a_n)\right]\sum\limits_{m=0}^{\tau-1} \nabla \log \mu^\theta(a_m\left|x_m\right.) \right| x_0 \right]\!.
\end{align*}

To estimate the gradient of the CVaR of $D(\theta, x_0)$ for a given policy parameter $\theta$, we use the following variation of the policy gradient theorem for CVaR:
\boxed{
\vspace*{-10pt}
\begin{align}
&\nabla \text{CVaR}_{\beta}(D(\theta, x_0)) = \E\left[ \left[ D(\theta, x_0)- \text{VaR}_{\beta}(D(\theta, x_0)) \right]\right.\nonumber\\
&\left.\left.\times\sum\limits_{m=0}^{\tau-1}\! \nabla \log \mu^\theta(a_m\left|x_m\right.) \right|\! D(\theta, x_0) \geq \text{VaR}_{\beta}(D(\theta, x_0)) \right].\label{eq:cvar-pg-expr}
\end{align}}
We shall provide a derivation of the expression above in the next chapter. In particular, we shall specialize the gradient expression for an abstract coherent risk measure to handle the case of CVaR, and the reader is referred to Section \ref{sec:coherent-algo} for the details.
 
\subsection*{VaR and CVaR estimation}
What remains to be specified is the technique employed for estimating VaR and CVaR for a given policy $\theta$. Notice that CVaR estimation is required for dual ascent, since  $\nabla_\lambda L(\theta,\lambda)= \text{CVaR}_{\beta}(D(\theta, x_0))-\kappa$. 
VaR is required for estimating CVaR and the CVaR gradient.
A well-known result is that both VaR and CVaR can be obtained from the solution of a certain convex optimization problem. More precisely,
for any r.v. $X$, let 
\[v(\xi,X):=\xi + \frac{1}{1-\beta}(X-\xi)_{+} \textrm{  and }V(\xi)=\E\left[v(\xi,X)\right].\]
Then, $\textnormal{VaR}_{\beta}(X)$ is the minimizer of $V$, i.e., a point $\xi^*_\beta$ that satisfies $V'(\xi^*_\beta)=0$ and $\textnormal{CVaR}_{\beta}(X)=V(\xi^{*}_{\beta})$. 

Since $v(\xi,\cdot)$ is continuous w.r.t. $\xi$, $V'(\xi)=\E\left(1-\frac{1}{1-\beta}\indic{X\geq\xi}\right)$. The minimizer $\xi^*$ would be a VaR, and $V(\xi^*)$ would be the CVaR of the r.v. $X$.  Observing that $V$ is convex, a stochastic approximation-based procedure can be derived for estimating VaR and CVaR.  In an SSP context, the r.v. is $D(\theta, x_0)$. Suppose that we can obtain i.i.d. samples from the distribution of $D(\theta, x_0)$, i.e., we can simulate the underlying SSP using the policy $\theta$. Let $D_{k}, k=1,\ldots$ denote these samples. Then, VaR and CVaR can be estimated as follows:
\begin{align}
\textrm{VaR: } \xi_{m}&= \xi_{m-1}-\zeta_{3}(m)\left(1-\frac{1}{1-\beta}\indic{D_m\geq\xi_m}\right),\label{eq:RMVaR}\\
\textrm{CVaR: } C_{m}&=C_{m-1}-\frac{1}{m}\left(C_{m-1} - v(\xi_{m-1},C_{m-1})\right).\label{eq:cvar-update}
\end{align}
In the above, \eqref{eq:RMVaR} can be seen as a gradient descent rule, while \eqref{eq:cvar-update} can be seen as a plain averaging update. The step-size sequence $\{\zeta_3(m)\}$ is required to satisfy standard stochastic approximation conditions, \\i.e., $\sum\limits_m \zeta_3(m) = \infty,$ and 
$\sum\limits_m \zeta_3(m)^2 <\infty$.

The complete algorithm, along with the update rules for various parameters, is presented in Algorithm \ref{alg:cvar-PG}.

\begin{algorithm}
	\SetKwInOut{Input}{Input}\SetKwInOut{Output}{Output}
	\Input{initial parameter $\theta_0 \in \Theta$, where $\Theta$ is a compact and convex subset of $\R^d$,  $\beta\in (0,1)$, trajectory lengths $\{m_n\}$, step sizes $\{\zeta_1(n)\}$, $\{\zeta_2(n)\}$, $\{\zeta_3(n)\}$, projection operators $\Gamma$ and $\Gamma_\lambda$, \#~iterations $M\gg 1$.}
	\For{$n\leftarrow 0$ \KwTo $M-1$}{
		\For{$m\leftarrow 0$ \KwTo $m_n-1$}{
			Simulate the SSP for an episode to generate the state sequence $\{x_{n,j}\}$ using actions $\{a_{n,j} \sim \mu^{\theta_n}\left(\cdot \mid x_{n,j}\right)\}$. Let $\tau_m$ denote the time instant when state $0$ was visited in this episode\;
			Observe total cost $D_{n,m} = \sum\limits_{j=0}^{\tau_m-1} k(x_{n,j},a_{n,j})$\;
			Calculate likelihood ratio: $\psi_{n,m} = \sum\limits_{j=0}^{\tau_m-1} \nabla \log \mu^{\theta_n}(a_{n,j}|x_{n,j})$\; 
		}
		\tcc{Policy evaluation}
		Use the scheme in  \eqref{eq:RMVaR}--\eqref{eq:cvar-update} to obtain the VaR estimate $\xi_n$ and $\textnormal{CVaR}_{\beta}$-estimate $C_n$\;
		$\text{Total cost estimate: }      \bar D_{n}= \dfrac{1}{m_n} \sum\limits_{j=1}^{m_n}D_{n,j}$\;
		$\text{Likelihood ratio: }      \bar\psi_{n}= \dfrac{1}{m_n} \sum\limits_{j=1}^{m_n}\psi_{n,j}$\;
		\tcc{Gradient of the objective}
		$    \widehat\nabla J(\theta_n) =  \bar D_n \bar\psi_n$\;
		\tcc{Gradient of the risk measure}
		$\widehat \nabla \text{CVaR}_{\beta}\left(\theta_n\right) =  (C_{n} - \xi_n) \bar\psi_n \indic{C_n\geq\xi_{n}}$\;
		
		\tcc{Policy and Lagrange Multiplier Update}
		$\theta_{n+1} = \Gamma\left(\theta_{n}-\zeta_2(n)\left(\widehat\nabla J(\theta_n) + \lambda_{n} \widehat \nabla \text{CVaR}_{\beta}\left(\theta_n\right)\right)\right)$\; 
		$\lambda_{n+1}=\Gamma_\lambda\Big(\lambda_{n}+\zeta_1(n)(C_n-\kappa)\Big)$\;
	}
	\Output{Policy $\theta_M$} 
	\caption{Policy gradient algorithm under CVaR as a risk measure in an SSP setting}
	\label{alg:cvar-PG}
\end{algorithm}

\section{Case 4: Stochastic shortest path + chance constraint as risk}
\label{sec:chance-MDP-algo}

The last case studied in this section employs a chance constraint in the optimization problem \eqref{eq:gen-constrained-opt}, i.e.,
\begin{align*}
& \min J(\theta,x_0)\quad  \quad\text{subject to} \quad\quad G(\theta,x_0)\leq\kappa,
\end{align*}
where $J(\theta,x_0)$ is the expectation the total cost r.v. $D(\theta, x_0)$, while 
$G(\theta,x_0) = \Prob{D(\theta, x_0) \ge \beta}$ is the probability that feeds into the chance constraint (see Sections \ref{sec:mdp-ssp} and \ref{sec:risk-chance}).

From the discussion in the previous sections, it is apparent that the main technical challenges in handling any risk measure are as follows: (i) estimation of the risk measure from samples; and (ii) gradient estimation for the policy update iteration.
For the sake of brevity, we provide the necessary details for handling (i) and (ii), and the rest of the pieces of the resulting actor-critic scheme follows in a manner similar to that for variance or CVaR. 

To handle (i), suppose that we are given $n$ i.i.d. samples, say $\{X_1,\ldots,X_n\}$, from the distribution of $X$, and the goal is to  estimate the probability involved in the chance constraint, i.e., $\Prob{X \ge \beta}$. The sample average estimator for the latter probability is given by
\begin{align}
\overline \C_n &=  \frac{1}{n} \sum_{i=1}^{n} \indic{X_i \ge \beta}.
\label{eq:chance-est}
\end{align}
For handling point (ii) concerning the policy gradient for the chance probability, we use the following likelihood ratio gradient expression for the chance probability:
\boxed{
\vspace*{-10pt}
\begin{align*}
&\nabla G(D(\theta, x_0)) = \E\left[ \sum\limits_{m=0}^{\tau-1}\! \nabla \log \mu^\theta(a_m\left|x_m\right.)  \indic{D(\theta, x_0) \ge \beta} \right].
\end{align*}}

\vspace*{10pt}
\noindent
The complete algorithm with chance constraint as the risk measure and the usual value function as the objective is presented in Algorithm \ref{alg:chance-PG}.

\begin{algorithm}
	\SetKwInOut{Input}{Input}\SetKwInOut{Output}{Output}
	\Input{initial parameter $\theta_0 \in \Theta$, where $\Theta$ is a compact and convex subset of $\R^d$,  $\beta\in (0,1)$, trajectory lengths $\{m_n\}$, step sizes $\{\zeta_1(n)\}$, $\{\zeta_2(n)\}$, $\{\zeta_3(n)\}$, projection operators $\Gamma$ and $\Gamma_\lambda$, \#~iterations $M\gg 1$.}
	\For{$n\leftarrow 0$ \KwTo $M-1$}{
		\For{$m\leftarrow 0$ \KwTo $m_n-1$}{
			Simulate the SSP for an episode to generate the state sequence $\{x_{n,j}\}$ using actions $\{a_{n,j} \sim \mu^{\theta_n}\left(\cdot \mid x_{n,j}\right)\}$. Let $\tau_m$ denote the time instant when state $0$ was visited in this episode\;
			Observe total cost $D_{n,m} = \sum\limits_{j=0}^{\tau_m-1} k(x_{n,j},a_{n,j})$\;
			Observe sample chance constraint $C_{n,m} = \indic{D_{n,m} \ge \beta}$\;
			Calculate likelihood ratio: $\psi_{n,m} = \sum\limits_{j=0}^{\tau_m-1} \nabla \log \mu^{\theta_n}(a_{n,j}|x_{n,j})$\; 
		}
		\tcc{Policy evaluation}
		$\text{Total cost estimate: }      \bar D_{n}= \dfrac{1}{m_n} \sum\limits_{j=1}^{m_n}D_{n,j}$\;
		$\text{Likelihood ratio: }      \bar\psi_{n}= \dfrac{1}{m_n} \sum\limits_{j=1}^{m_n}\psi_{n,j}$\;
		$\text{Chance constraint estimate: }      \bar C_{n}= \dfrac{1}{m_n} \sum\limits_{j=1}^{m_n}C_{n,j}$\;
		\tcc{Gradient of the objective}
		$    \widehat\nabla J(\theta_n) =  \bar D_n \bar\psi_n$\;
		\tcc{Gradient of the risk measure}
		$\widehat \nabla G\left(\theta_n\right) =  \bar C_n \bar\psi_n$\;
        \tcc{Policy and Lagrange Multiplier Update}
		$\theta_{n+1} = \Gamma\left(\theta_{n}-\zeta_2(n)\left(\widehat\nabla J(\theta_n) + \lambda_{n} \widehat \nabla G\left(\theta_n\right)\right)\right)$\; 
		$\lambda_{n+1}=\Gamma_\lambda\Big(\lambda_{n}+\zeta_1(n)(C_n-\kappa)\Big)$\;
	}
	\Output{Policy $\theta_M$} 
	\caption{Policy gradient algorithm under the chance constraint in an SSP setting}
	\label{alg:chance-PG}
\end{algorithm}

\clearpage
\section{Bibliographic remarks}
The presentation of the risk-sensitive RL algorithm with variance as the underlying risk measure in discounted and average cost MDPs is based on \textcite{prashanth2016mlj}, while the descriptions for the cases of CVaR and chance constraints are based on
\textcite{prashanth2014cvar} and \textcite{chow2017risk}, respectively.
In the following, we provide additional bibliographic remarks for each case studied.
\begin{description}
 \item[\ref{sec:discounted-MDP-algo}] For a justification of the requirement in \eqref{eq:ns1}, see \textcite[Lemma 1 in Chapter 2]{borkar2008stochastic}.
In \textcite{prashanth2016mlj}, the authors parameterize both $J$ and $U$ using linear function approximation, and show that the underlying
projected Bellman operators are contractive; see \textcite[Lemma 2]{prashanth2016mlj} for a proof. 
The fact that linear parameterization for $J$ leads to a contraction mapping is well known, and a similar approach was shown to work for the squared cost $U$ in \textcite{Tamar13TD} for an SSP setting. In \textcite{prashanth2016mlj}, the authors  extended this idea to include discounted problems. Notice that linear parameterization for $J$ and $U$ implies the underlying variance is also parameterized; however, a direct parameterization of variance is not feasible, as the underlying operator is not monotone, see \textcite{Sobel82VD}. For the finer details of the linear function approximation case in the discounted setting, see \textcite{prashanth2016mlj}.

\item[\ref{sec:avg-MDP-algo}] The expression in \eqref{eq:grad-rho} for the gradient of the average cost was derived independently in \textcite{Marbach2001} and \textcite{sutton1999policy}. This expression leads naturally to policy gradient algorithms, cf. \textcite{bartlett2011infinite}.  There is a corresponding discounted variant of this expression in \textcite{sutton1999policy}, and the policy gradient technique in \textcite{bartlett2011infinite}.
 As in the discounted setting, incorporating function approximation for the functions $J$ and $U$ is straightforward, and we refer the reader to \textcite{prashanth2016mlj} for the case where a linear function approximation architecture is used. 

\item[\ref{sec:cvar-MDP-algo}] 
\textcite{rockafellar2000optimization} first showed that both VaR and CVaR can be obtained from the solution of a certain convex optimization problem,  so that a stochastic approximation-based procedure can be derived for estimating VaR and CVaR, as in \textcite{bardou2009computing}, in turn leading to a specialization for MDPs in \textcite{prashanth2014cvar}. For such a scheme, a non-asymptotic analysis is not available. However, stochastic gradient schemes have received a lot of attention from a non-asymptotic analysis viewpoint, see \citep{bottou2018optimization} for a survey. Since VaR estimation through \eqref{eq:RMVaR} falls under the realm of stochastic gradient schemes, one could use the bounds in the aforementioned reference.   As an alternative, one could consider a direct sample average approximation (SAA) of CVaR, see \citep{serfling2009approximation}.  Concentration bounds for such an SAA-approximation of CVaR has received a lot of attention in past decade or so, cf.  \citep{brown2007large,wang2010deviation,kolla2019concentration,bhat2019concentration,thomas2019concentration,prashanth2020concentration}. Such results would be useful for the analysis of Algorithm \ref{alg:cvar-PG}, considering that one needs to decide on the number of episodes $m_n$ in each policy gradient iteration; see the discussion in Section \ref{sec:mn}. 

The likelihood ratio-based gradient estimate for CVaR was derived by \textcite{tamar2014policy} for the case of continuous distributions. 
An expression for the gradient of an abstract coherent risk measure, 
for both discrete and continuous distributions and  
also specialized to handle the case of CVaR, is derived in \textcite{tamar2015coherent}. 
We shall present this expression as well as the specialization in Section \ref{sec:coherent-algo}. 

\end{description}

%
\chapter{MDPs with Risk as the Objective}
\label{sec:risk-algos-unconstrained}
\diff{In this chapter, we discuss policy gradient algorithms for solving risk-sensitive MDPs where the risk measure is explicitly the objective, i.e., the following optimization problem:
\begin{align}
	\min_{\theta\in\Theta} G(\theta),\label{eq:pb2}
\end{align}
where $G$ is one of the risk measures presented in Chapter \ref{sec:risk-measures} that consider the entire distribution.
Specifically, we consider exponential cost, CPT, and coherent risk measures in Sections \ref{sec:expcost-algo}, \ref{sec:cpt-MDP-algo}, and \ref{sec:coherent-algo}, respectively.
This complements the incorporation of risk measures such as variance or CVaR that are based on the tail of the underlying distribution, which were considered in the constrained optimization formulations in the previous chapter.

Following the template in Chapter \ref{sec:rpg-template}, the main ingredients in each iteration $n$ of a policy gradient algorithm for optimizing a risk objective are as follows:
\begin{enumerate}[label=(\Roman*)]
\item Simulation of the underlying MDP to obtain one or more sample trajectories. 
\item Estimation of $\nabla G(\theta_n)$ from the sample data.
\item Incremental update of the policy parameter in the descent direction using the gradient estimate from the step above, i.e., 
\[\theta_{n+1} = \Gamma\big[\theta_n - \zeta(n)  \widehat \nabla G(\theta_n)\big],\]
where $\zeta(n)$ is the step size and $\Gamma$ is a projection operator that keeps the iterate bounded. 
\end{enumerate}
}

\section{Case 1: Average-cost MDP + Exponential cost as risk}
\label{sec:expcost-algo}
In many cases studied earlier, the recipe for a risk-sensitive policy gradient algorithm is to first derive an expression for the gradient of the risk measure, and then use this expression to form an estimator using sample trajectories of the underlying MDP. 


Recall from Section \ref{sec:exp-cost}, under \ref{ass:avgcost-same}, the exponential cost associated with a policy $\mu^\theta$ is given by
	\begin{align}
	G(\theta) 
	& =  \lim_{T\rightarrow\infty}\frac{1}{T} \frac{1}{\beta} \log \E\left[\exp\left(\beta\sum_{n=0}^{T-1}k(x_n,a_n)\right)\right],\label{eq:expcost-variant-again}
\end{align}
where $\beta$ is the risk-sensitivity parameter.

For the analysis, we also require aperiodicity in addition to irreducibility and positive recurrence, which we specify in the following variant of \ref{ass:avgcost-same}.
\begin{assumption}
	\label{ass:avgcost-expcost}
	For each policy $\mu^\theta$, the underlying Markov chain is irreducible, positive recurrent, and aperiodic.
\end{assumption}
Define the $|\X| \times |\X|$ matrix $A_\theta$ as 
\begin{equation}
\label{eq:A-matrix}
A_\theta \triangleq \frac{1}{\beta}\left[\sum_a \mu^\theta(a | x)\exp\left(\beta k(x,a)\right) P(y | x,a)\right]_{x,y\in\X}.
    \end{equation}

Since the Markov chain underlying $\mu^\theta$ is assumed to be irreducible, and each entry of $ A_\theta$ is non-negative, we can apply Perron-Frobenius theorem to infer that there exists a unique eigenvalue-eigenvector pair $(\lambda_\theta,V(\theta))$ satisfying
\[ \lambda_\theta>0, \ V(\theta,i) >0, \forall i \textrm{ and } |\lambda'| \le \lambda_\theta \textrm{ for any other eigenvalue }\lambda' \textrm{ of }A_\theta.\]

\boxed{
\begin{proposition}
	\label{prop:expcost-eigenval}
	Assume \ref{ass:avgcost-expcost} and that the state-action spaces of the underlying MDP are both finite. Then, for any policy $\mu^\theta$, 
	\[ G(\theta) = \log \lambda_\theta,\]
	where $G(\theta)$ is the exponential cost associated with the policy $\mu^\theta$ given by \eqref{eq:expcost-variant-again} and $\lambda_\theta$ is the Perron-Frobenius eigenvalue of the matrix $A_\theta$ defined by \eqref{eq:A-matrix}.
\end{proposition}	
}
\begin{proof}
	Since $(\lambda_\theta,V(\theta))$ is an eigenvalue-eigenvector pair associated with the matrix $A_\theta$, we have $A_\theta V(\theta) = \lambda_\theta V(\theta)$, or equivalently,
	\begin{align*}
		\lambda_\theta V(\theta,x) = \frac{1}{\beta}\sum_{y }  \sum_a \mu^\theta(a  | x)\exp\left(\beta k(x,a)\right) P(y | x,a) V(\theta,y).
	\end{align*}
Let 
\begin{align}\tilde P_\theta(y | x) = \dfrac{\sum_a \mu^\theta(a | x) \exp\left(\beta k(x,a)\right) P(y | x,a) V(\theta,y)}{\beta \lambda_\theta V(\theta,x)}, \ \forall x,y \in \X. ~~~~~~ \label{eq:ptilde}
	\end{align}
Notice that $\tilde P_\theta(y | x) \ge 0$ and $\sum_{y} \tilde P_\theta(y | x) = 1$, implying $\tilde P_\theta(y | x)$ is a valid transition probability function. 

Let $\{\tilde x_n\}$ be a Markov chain governed by the transition probability function $\tilde P_\theta$. Then, under \ref{ass:avgcost-expcost}, this Markov chain is irreducible, positive recurrent and aperiodic, which in turn implies the existence of a stationary distribution, say $\tilde \psi$. Thus, by ergodicity, we have
\[  \EE{h(\tilde x_n)} \rightarrow \sum_x \tilde \psi(x) h(x) \textrm{ a.s. as } n\rightarrow \infty. \]

\vspace*{10pt}
\noindent
Notice that
\begin{align*}
	&\frac{1}{T}\frac{1}{\beta}\log \E\left[\exp\left(\beta\sum_{n=0}^{T-1}k(x_n,a_n)\right)\middle| x_0\right]\\
	& = \frac{1}{T}\frac{1}{\beta}\log \left[\sum_{\substack{x_1,\ldots,x_T,\\a_0,\ldots,a_{T-1}}}  \prod_{n=0}^{T-1}\exp\left(\beta k(x_n,a_n)\right) \mu^\theta(x_n,a_n) P(x_{n+1} | x_n,a_n)\right]\\
	& = \frac{1}{\beta T}\times\\
	&\log \left[\sum_{\substack{x_1,\ldots,x_T,\\a_0,\ldots,a_{T-1}}}  \prod_{n=0}^{T-1}\dfrac{\exp\left(\beta k(x_n,a_n)\right) \mu^\theta(a_n | x_n)V(\theta,x_{n+1}) P(x_{n+1} | x_n,a_n)}{\lambda_\theta V(\theta,x_n)} \right.\\
	&\qquad\qquad\qquad\qquad \left.\times \dfrac{\lambda_\theta V(\theta,x_0)}{V(\theta,x_T)}\right] \\
	&= \log \lambda_\theta + \frac{1}{T}\frac{1}{\beta} \left( \log V(\theta,x_0) - \log \EE{V(\theta,\tilde x_T)} \right) 
    \rightarrow \log \lambda_\theta \textrm{ as } T\rightarrow \infty.
\end{align*}
The claim follows.
\end{proof}
Since the state and action spaces are assumed to be finite, we have $\lambda^* = \min_{\mu^\theta} \lambda_\theta$, where the minimum is taken over all randomized policies. Let $V^*$ denote the corresponding eigenvector. Then it can be shown that
\begin{align}
	\lambda^* V^*(x) = \min_{\mu} \hspace{-2pt}\left(\frac{1}{\beta}\sum_a \mu(a | x)  \exp(\beta k(x,a))\sum_{y} P(y | x,a) V^*(y) \hspace{-3pt}\right)\hspace{-3pt}. ~~~~~ 
\label{eq:expcost-be}
\end{align}

As in the case of risk-neutral average-cost MDPs, we can define Q-values that assist in solving the problem of control. 
For the exponential cost case, the optimal Q-value is defined as
\begin{align}
	 Q^*(x,a) =  \frac{\exp(\beta k(x,a))}{\beta\lambda^*}\sum_{y} P(y | x,a) V^*(y). 
	\label{eq:expcost-qdef}
\end{align}
$(Q^*,\lambda^*)$ is a solution, unique up to a scalar multiple, of the following eigenvalue problem:
\begin{align}
	Q^*(x,a)\ \lambda^* =  \frac{\exp(\beta k(x,a))}{\beta}\sum_{y} P(y | x,a) \min_b Q^*(y,b). 
	\label{eq:expcost-qbe}
\end{align}
Notice that $V(x) = \min_a Q^*(x,a)$ satisfies \eqref{eq:expcost-be}.

For the special case of a fixed policy $\mu^\theta$, we define the Q-value analogue as 
\begin{align}
	Q(\theta, x,a) =  \frac{\exp(\beta k(x,a))}{\beta \lambda_\theta}\sum_{y} P(y | x,a) V(\theta,y). 
	\label{eq:expcost-qmudef}
\end{align}
$(Q^\theta,\lambda_\theta)$ satisfies the following eigenvalue problem:
\begin{equation}
	Q(\theta, x,a)\ \lambda_\theta =  \frac{\exp(\beta k(x,a))}{\beta}\sum_{y} P(y | x,a) \sum_b \mu(b | y) Q(\theta, y,b). 
	\label{eq:expcost-qmube}
\end{equation}
For the sake of consistent notation, we shall use $Q_\mu(\cdot,\cdot)$ and $V_\mu(\cdot)$ to denote the Q-value and the Perron-Frobenius eigenvector associated with a policy $\mu$ that is not necessarily parameterized.

The results in \eqref{eq:expcost-qbe} and \eqref{eq:expcost-qmube} can be used to derive value and policy iteration algorithms for finding a policy that optimizes the exponential cost.
We present the policy iteration algorithm, as it forms the basis for the risk-sensitive policy gradient algorithm presented later in this section. 

\vspace{-9pt}
\begin{center}
\fbox{
\begin{minipage}{0.96\textwidth}
{\bfseries Policy iteration for exponential cost}\ \\[0.5ex]
{\bfseries Initialization:} Policy $\mu_0$, fixed state $x_f$.\ \\[0.5ex]
{\bfseries For all $n=1,2,\dots$, repeat}
\begin{description}
\item[Policy evaluation:] Solve the eigenvalue problem
\begin{align} 
\hspace{-46pt}
V_n(x) & = \sum_a \mu_n(a | x) \frac{\exp(\beta k(x,a))}{\beta \lambda_n}\sum_{y} P(y | x,a) V_n(y),  \\
\hspace{-46pt}
V_n(x_f) &=1.
\label{eq:expcost-policyeval}
\end{align}
\item[Policy improvement:] Choose action according to 
\[ 
\hspace{-26pt}
\mu_{n+1}(\cdot  | x) \in \argmin_{\mu} \left[ \sum_a \mu_n(a | x) \frac{\exp(\beta k(x,a))}{\beta}\sum_{y} P(y | x,a) V_n(y) \right].
\]
\end{description}
\end{minipage}
}
\end{center}
The policy evaluation step in the algorithm above can be performed in an iterative fashion as follows:
Initializing with $V_n^0=V_{n-1}$, update
\begin{align} 
\tilde V_n^{m+1}(x) & = \sum_a \mu_n(a | x) \frac{\exp(\beta k(x,a))}{\beta }\sum_{y} P(y | x,a) V_n^m(y),\\
V_n^{m+1}(x) &= \frac{\tilde V_n^{m+1}(x)}{\tilde V_n^{m+1}(x_f)}.
\end{align}
The above variation of policy evaluation can be seen as value iteration for a fixed policy. Such a scheme can be shown to converge, and the limit coincides with the solution to the eigenvalue problem in \eqref{eq:expcost-policyeval}.

Next, the policy iteration algorithm could be written using Q-values. Such an algorithm is not really necessary in a context where the transition dynamics of the underlying MDP is known. However, the actor-critic algorithm presented subsequently could be seen as a learning variant of the Q-value based policy iteration, which we present next.

\begin{center}
\fbox{
\begin{minipage}{0.96\textwidth}
{\bfseries Policy iteration using Q-values}\ \\[0.5ex]
{\bfseries Initialization:} Policy $\mu_0$, fixed state $x_f$ and action $a_f$.\ \\[0.5ex]
{\bfseries For all $n=1,2,\dots$, repeat}
\begin{description}
\item[Policy evaluation:] 
With $Q_n^0=Q_{n-1}$, update (until convergence)
\begin{align} 
\hspace{-1.2em}\tilde Q_n^{m+1}(x,a) & = \sum_a \frac{\exp(\beta k(x,a))}{\beta }\sum_{y}P(y | x,a) \sum_b \mu_n(b | y)  Q_n^m(y,b),\\
\hspace{-1.2em}Q_n^{m+1}(x,a) &= \frac{\tilde Q_n^{m+1}(x,a)}{\tilde Q_n^{m+1}(x_f,a_f)}.
\end{align}

\item[Policy improvement:] Choose action according to 
\[ \mu_{n+1}(\cdot  | x) \in \argmin_{\mu} \left[ \sum_a \mu_n(a  | x) Q_n^{m+1}(x,a) \right].\]
\end{description}
\end{minipage}
}
\end{center}

\vspace{1ex}

The result below presents a variant of the policy gradient theorem for the exponential cost risk measure.

\clearpage
\boxed{
\vspace*{-10pt}
\begin{proposition}
Assume \ref{ass:avgcost-expcost}. Then,
\begin{align}
	\nabla \lambda_\theta = \sum_{x,a} \tilde \psi_\theta(x) \nabla \mu^\theta(a | x) \tilde Q(\theta, x,a) \lambda_\theta,\label{eq:expcost-grad}
\end{align}	
$$
\mbox{where~} 
\tilde Q(\theta, x,a) = \dfrac{\exp\left(\beta k(x,a)\right)}{\beta V(\theta, x)\lambda_\theta} \sum_{y }   P(y | x,a) V(\theta, y)
$$   
is the modified Q-value function, 
and $\tilde \psi_\theta$ is the stationary distribution underlying a Markov chain governed by the  transition probability function $\tilde P_\theta(\cdot  | \cdot)$ defined in \eqref{eq:ptilde}.
\end{proposition}	
}
\begin{proof}
Letting $V(\theta)$ denote the eigenvector corresponding to $\lambda_\theta$, the eigenvalue equation can be written as 
\begin{align}
	&\lambda_\theta V(\theta, x) = \frac{1}{\beta}\sum_{y }  \sum_a \mu^\theta(a | x)\exp\left(\beta k(x,a)\right) P(y | x,a) V(\theta, y),\nonumber\\
	&\textrm{ or equivalently, }\nonumber\\
	 &V(\theta, x) =  \sum_a \mu^\theta(a | x)\frac{\exp\left(\beta k(x,a)\right)}{\beta\lambda_\theta}\sum_{y} P(y | x,a) V(\theta, y).\label{eq:eigeneqn}
\end{align}
Setting $V(\theta, x_0)=1$ would ensure that the solution $V(\theta)$ to \eqref{eq:eigeneqn} is unique. 

Differentiating w.r.t. $\theta$ in \eqref{eq:eigeneqn}, we obtain
\begin{align*}
	\nabla V(\theta, x) &= \sum_a \mu^\theta(a | x)\frac{\exp\left(\beta k(x,a)\right)}{\beta\lambda_\theta}\sum_{y} P(y | x,a) \nabla V(\theta, y)\\
	& + \sum_a \nabla \mu^\theta(a | x)\frac{\exp\left(\beta k(x,a)\right)}{\beta\lambda_\theta}\sum_{y} P(y | x,a) V(\theta, y)\\
	&- \sum_a \mu^\theta(a | x)\frac{\exp\left(\beta k(x,a)\right)}{\beta\lambda_\theta^2}\sum_{y} P(y | x,a) V(\theta, y) \nabla \lambda_\theta.
\end{align*}
Dividing by $V(\theta, x)$ and then summing over  the stationary distribution $\tilde \psi_\theta$ on both sides of the equation above, we obtain
\begin{align*}
	&\sum_x \tilde \psi_\theta(x)\frac{\nabla V(\theta, x)}{V(\theta, x)} \\
	&= \underbrace{\sum_x \tilde \psi_\theta(x)\sum_a \mu^\theta(a | x)\frac{\exp\left(\beta k(x,a)\right)}{\beta\lambda_\theta V(\theta, x)}\sum_{y} P(y | x,a) \nabla V(\theta, y)}_{(I)}\\
	& + \underbrace{\sum_x \tilde \psi_\theta(x)\sum_a \nabla \mu^\theta(a | x)\frac{\exp\left(\beta k(x,a)\right)}{\beta\lambda_\theta V(\theta, x)}\sum_{y} P(y | x,a) V(\theta, y)}_{(II)}\\
	&- \underbrace{\sum_x \tilde \psi_\theta(x)\sum_a \mu^\theta(a | x)\frac{\exp\left(\beta k(x,a)\right)}{\beta V(\theta, x)\lambda_\theta^2}\sum_{y} P(y | x,a) V(\theta, y) \nabla \lambda_\theta}_{(III)}.~~~~~~\stepcounter{equation}\tag{\theequation}
\label{eq:gr1}
\end{align*}
Using the fact that $\tilde P_\theta$, defined in \eqref{eq:ptilde}, is a transition probability function, and also that $\tilde \psi_\theta$ is a stationary distribution, we simplify each of the terms on the RHS of \eqref{eq:gr1} as follows:
\begin{align*}
	(I) &= \sum_x \tilde \psi_\theta(x) \sum_{y} \tilde P(y | x) \frac{\nabla V(\theta, y)}{V(\theta, y)} = \sum_x \tilde \psi_\theta(x)\frac{\nabla V(\theta, x)}{V(\theta, x)},\\
	(II) &=  \sum_x \tilde \psi_\theta(x)\sum_a \nabla \mu^\theta(a | x) \tilde Q(\theta, x,a), \\
	(III) &= \sum_{y} \frac{\nabla\lambda_\theta}{\lambda_\theta} \tilde P(y | x) = \frac{\nabla\lambda_\theta}{\lambda_\theta}.
\end{align*}
From the above simplifications, it is apparent that term (I) is the same as the LHS in \eqref{eq:gr1}. A reordering of the simplified terms (II) and (III) leads to 
\begin{align*}
	\frac{\nabla\lambda_\theta}{\lambda_\theta} = \sum_{x,a} \tilde \psi_\theta(x) \nabla \mu^\theta(a | x) \tilde Q(\theta, x,a).
\end{align*}
The claim follows.
\end{proof}

\subsection*{Policy gradient algorithm for exponential cost}
\label{sec:expcost-pgalgo}
In the case of the regular value function, the policy gradient theorem \eqref{eg:gradJ} lends itself to an RL algorithm easily, since one can replace the expectation on the RHS of \eqref{eg:gradJ} with a sample trajectory-based approximation. On the other hand, the formula in \eqref{eq:expcost-grad} is complicated from a sampling viewpoint because the distribution on the RHS of \eqref{eq:expcost-grad} is different from the transition dynamics underlying the given MDP. To elaborate, the averaging in the policy gradient expression for exponential cost involves the stationary distribution $\tilde \psi_\theta$ that underlies the Markov chain governed by transition probabilities $\tilde P(\cdot  | \cdot)$, and it is not practically feasible to obtain samples from this distribution.
However, one can develop a policy gradient-type algorithm without a compact representation, i.e., by treating the policy as a vector of probabilities. 

The main idea that leads to the policy update that we describe next is the following:
Letting $\Lambda_\theta=\log \lambda_\theta$, we have the following variant of \eqref{eq:expcost-grad}: 
\begin{align}
	\nabla \Lambda_\theta = \sum_{x,a} \tilde \psi_\theta(x) \nabla \mu^\theta(a | x) \tilde Q(\theta, x,a),\label{eq:expcost-grad-variant}
\end{align}	
Since $\log$ is monotone, the minimizer of $\lambda_\theta$ coincides with that of $\Lambda_\theta$. 
Next, treating the policy as a probability vector over all states and all but one action, i.e., $\mu^\theta = [\mu^\theta(x,a)]_{x\in\X,a\in \A\setminus a_f}$, where $a_f$ is a fixed action.
Given $\mu^\theta$, we can infer the probability of choosing action $a_f$ in state $x$ as follows:
\begin{align} 
\mu^\theta(a_f | x) = 1 - \sum_{a\ne a_f} \mu^\theta(a | x).
\label{eq:mu-af}
\end{align}
The component of the RHS of \eqref{eq:expcost-grad-variant} corresponding to state-action pair $(x,a)$ is 
\begin{align}
 \frac{\partial  \Lambda_\theta}{\partial \mu^\theta(a | x)}&=\tilde \psi_\theta(x) \nabla \mu^\theta(a | x) \tilde Q(\theta, x,a) + \tilde \psi_\theta(x) \nabla \mu^\theta(a_f | x) \tilde Q(\theta, x,a_f)\\
&=\tilde \psi_\theta(x) \left( \tilde Q(\theta,x,a) - \tilde Q(\theta,x,a_f) \right).\label{eq:partial-expcost}
\end{align}
The  equalities above follow from \eqref{eq:mu-af}.
Thus, it is enough to use the factor $\left( \tilde Q(x,a) - \tilde Q(x,a_f) \right)$ to perform a gradient descent in the policy, while ignoring the multiplicative factor $  \tilde \psi_\theta(x)$, which is not available in practice for an RL algorithm. 

We next describe an algorithm that performs such a gradient descent update in the policy space. Notice that, for the policy update, one would require $\tilde Q(\theta, \cdot,\cdot)$, and the algorithm estimates this modified Q-value on the faster timescale, while performing the policy update on the slower timescale. 
The two-timescale algorithm requires varying step-size sequences $\{\zeta_1(n), \zeta_2(n)\}$ to satisfy the following conditions:
\begin{align*}
	&\sum\limits_{n = 1}^\infty \zeta_1(n) =\sum\limits_{n = 1}^\infty \zeta_2(n) = \infty,
	\sum\limits_{n = 1}^\infty \left ( \zeta_1^2(n) + \zeta_2^2(n) \right ) < \infty, \dfrac{\zeta_1(n)}{\zeta_2(n)} \rightarrow 0.
\end{align*}
Letting $d(n)$ denote either of the step sizes $\zeta_1(n)$ and $\zeta_2(n)$, $\forall\ z \in (0,1)$,
\begin{align*}
	&\sup_n \frac{d(\lceil{zn}\rceil)}{d(n)} < \infty, \text{ and } \sup_n \frac{A(\lceil{z'n}\rceil)}{A(n)}  \rightarrow 1 \text{ uniformly  in } z' \in [z,1], \\
	&\text{ where }A(n) = \sum\limits_{m=0}^n d(m).	
\end{align*}
The first set of conditions on the step-size sequences are standard for two-timescale stochastic approximation, which in particular ensure that $\zeta_2(n)$ would be on the slower timescale, while $\zeta_1(n)$ on the faster timescale. The additional conditions ensure that the both step sizes eventually decrease.

Since we treat the policy as a vector indexed by state-action pairs, we drop the dependence on the parameter $\theta$, and instead work with a randomized policy iterate, say $\mu_n$. 
Using step-size sequences satisfying these conditions, 
the algorithm for optimizing exponential cost would update along two timescales, with fixed $x_f \in \X, a_f \in \A$, as follows: 
\begin{align}
	 Q_{n+1}(x,a) &=  Q_n(x,a) + \zeta_1(\nu(x,a,n)) \indic{x_n=x,a_n=a}\nonumber\\
	&\qquad\times\left(\frac{\exp(\beta k(x,a))  Q_n(x_{n+1},a_{n+1})}{ \beta Q_n(x_f,a_f)}- Q_n(x,a)\right)\hspace{-2pt},~~~~~~\label{eq:q-expcost-pg}\\
	\mu_{n+1}(x) &= \Gamma \left(\mu_n(x) + \zeta_2(\nu(x,a,n)) \indic{x_n=x,a_n=a}\right.\nonumber\\
	&\left.\qquad\times\left( Q_n(x,a) -  Q_n(x,a_f)\right)\right),\label{eq:mu-expcost-pg}
\end{align}
where $\nu(x,a,n) = \sum\limits_{m=0}^n \indic{x_m = x, a_m=a}$ denote the number of times the state-action pair $(x,a)$ has been visited up to time $n$, and $\Gamma$ is a projection operator that ensures that, for any $x \in \X$, the updated policy $\mu_{n + 1}(x)$ 
stays within the simplex $\{ (d_1,\ldots,d_{|\A|-1}) \mid d_i \ge 0, \forall i=1,\ldots,|\A|-1, \sum\limits_{j=1}^{|\A|-1} d_j \le 1\}$.
The policy vector is updated for all but one action $a_f$, and the probability associated with this action can be inferred using \eqref{eq:mu-af}.
Furthermore, the actions $a_n$ are picked using an $\epsilon$-greedy randomized policy, i.e., w.p. $(1-\epsilon)$ choose an action according to $\mu_n$, and w.p. $\epsilon$ pick a random action. Here $\epsilon \in (0,1)$ is an exploration parameter, that is chosen to be small constant, or could be decayed as the algorithm updates. 

The faster timescale recursion in \eqref{eq:q-expcost-pg} can be seen as the stochastic approximation variant of the policy evaluation step in the policy iteration using Q-values. In other words, following the standard two timescale viewpoint to assume the policy $\mu$ is quasi-static, the faster timescale iterate $Q_n$ converges to $Q_\mu$.  
Next, viewing the faster timescale as almost equilibrated, the slower timescale recursion can be seen to perform a gradient step in the policy space with an exponential cost objective. The policy increment in \eqref{eq:mu-expcost-pg} is motivated by the discussion around \eqref{eq:partial-expcost}, and the remark below uses an ODE argument to show that one can ignore the positive multiplicative factor in the gradient expression, and still converge to a local minimum of the exponential cost objective. This argument is facilitated by the fact that we treat the policy as a vector of probabilities over all states and all but one action. 
\begin{remark}
The update iteration for $\mu(i, u)$ in \eqref{eq:mu-expcost-pg} is separate for each fixed $i$, with the following limiting ODE:
\begin{align}
	\dot\mu(i,\cdot) = -\left(Q_\mu(i,\cdot) - Q_\mu(i,a_f)\right).
	\label{eq:muone}
\end{align}
The gradient descent would have been
\begin{align}
	\dot\mu(i,\cdot)=-\tilde \psi(i)(\tilde Q_\mu(i,\cdot))- \tilde Q_\mu(i,a_f) = \frac{\tilde \psi(i)}{V_\mu(i)}(Q_\mu(i,\cdot)-Q_\mu(i,a_f)).
\end{align}
Thus, \eqref{eq:muone} is of the form
\begin{align}
	\dot x_i(t) = -C_i\nabla f(x(t)), \forall i,
\end{align}
with $C_i > 0,\ \forall i$. This still converges to a stationary point for almost all initial conditions, because
\[\frac{d}{dt} f(x(t)) = -\sum_i C_i \left\| \nabla f(x(t))\right\|^2 < 0, \]
except at critical points.
\end{remark}

\clearpage
\begin{algorithm}
	\SetKwInOut{Input}{Input}\SetKwInOut{Output}{Output}
	\Input{initial policy $\mu_0$,  step sizes $\{\zeta_1(n)\}$, $\{\zeta_2(n)\}$, exploration parameter $\epsilon$, projection operator $\Gamma$, \#~iterations $M\gg 1$.}
	\For{$n\leftarrow 0$ \KwTo $M-1$}{
		Draw action $a_n \sim \mu_n(\cdot|x_n)$ w.p. $(1-\epsilon)$ and pick a random action w.p. $\epsilon$\;
		Observe next state $x_{n+1}$ and cost $k(x_n,a_n)$\;
		\tcc{Q-value estimate}
			$ Q_{n+1}(x,a) =  Q_n(x,a) + \zeta_1(\nu(x,a,n)) \indic{x_n=x,a_n=a}$\\
			$\qquad\qquad\qquad\times\left(\frac{\exp(\beta k(x,a))  Q_n(x_{n+1},a_{n+1})}{ Q_n(x_f,a_f)}- Q_n(x,a)\right)$\;
		\tcc{Policy update}
    $\mu_{n+1}(x) = \Gamma \left(\mu_n(x) + \zeta_2(\nu(x,a,n)) \indic{x_n=x,a_n=a}\right.$\\
				$\left.\qquad\qquad\qquad\times\left( Q_n(x,a) -  Q_n(x_f,a)\right)\right)
			$\;
}
\Output{Policy $\mu_M$} 
\caption{Policy gradient algorithm under exponential cost as a risk measure in an average-cost MDP setting}
\label{alg:expcost-PG}
\end{algorithm}
\subsection*{Notes on convergence}
Unlike the other cases studied in this chapter below, the analysis of Algorithm \ref{alg:expcost-PG} does not conform to the template in Section \ref{sec:convergence-unconstrained}, because the policy update in \eqref{eq:mu-expcost-pg} treats the policy as a vector over all states and actions. If one uses a compact representation, say via a parameterized family of policies $\mu^\theta$, then there are two challenges involved in arriving at a policy gradient algorithm. The first relates to sampling. As mentioned before, the policy gradient expression in \eqref{eq:expcost-grad} involves a distribution that is different from the transition dynamics underlying the MDP considered. The second relates to projection. With a parameterized policy class, the projection $\Gamma$ onto the probability simplex should be replaced by a projection onto the set $\{\mu^\theta(\cdot | \cdot), \theta\in \Theta\}$, where $\Theta$ is the set of parameterized policies. However, the aforementioned set of probability vectors used for projection need not be convex.

We now provide a sketch of the convergence analysis for Algorithm \ref{alg:expcost-PG}.
Recall that Algorithm \ref{alg:expcost-PG} employs two-timescale stochastic approximation, i.e., it comprises of iteration sequences that are updated using two different step-size schedules defined via $\{\zeta_1(n)\}$ and $\{\zeta_2(n)\}$, respectively. The analysis follows
a standard sequence of steps needed to show convergence of
two-timescale stochastic approximation algorithms, as discussed in Section \ref{sec:convergence-constrained}. In particular, the faster timescale analysis for the modified Q-value estimate sees the policy update as quasi-static, while the slower timescale analysis for the policy $\mu$ views the modified Q-value updates to have converged.

For the analysis of the faster timescale $\tilde Q$-recursion, consider the following ODE:
$\forall x \in \S, i = 1, 2, \dots, N,$
\begin{align}
	\dot\varphi_{xa}(t) &= \frac{1}{|\X||\A|+1}\left(\frac{\exp(\beta k(x,a))\sum_{y} P(y | x,a) \sum_b \mu(b | y) \varphi_{yb}(t) }{\beta\varphi_{x_f a_f}(t)}\right.\nonumber\\
	&\left.-\varphi_{xa}(t)\right).	\label{eq:qtilde-ode}
\end{align}
where $\mu$ is considered to be time-invariant, since it is updated on the slower timescale. 
It can be shown that \eqref{eq:qtilde-ode} has a globally asymptotically stable equilibrium $Q_\mu$, which is the unique solution to the following system of equations:
\begin{align}
	Q(x,a) &= \left(\frac{\exp(\beta k(x,a))\sum_{y} P(y | x,a) \sum_b \mu(b | y) Q(y,b) }{\beta\lambda_\mu}\right),\ \forall x,a,\\
Q(x_f, a_f) &= \lambda_\mu.  \label{eq:tildeQfixedpt}
\end{align}

Using standard stochastic approximation arguments together with a stability result in Theorem \ref{thm:borkarmeyn}, for any given policy $\mu$, the faster timescale iterate $Q_n$ converges a.s. to $Q_\mu$. 

We now turn our attention to the policy update \eqref{eq:mu-expcost-pg}. Consider the following ODE:
\begin{align}
	\label{eq:expcost-mu-ode}
	\dot \chi_{x\cdot}(t) = Q^{\chi(t)}(x,\cdot) - Q^{\chi(t)}(x,a_f), \forall x \in \X
\end{align}
This ODE can be re-written as 
\begin{align}
	\label{eq:expcost-mu-ode-equiv}
	\dot \chi_{x\cdot}(t) = K_{x\cdot}(\chi(t)) - K_{x a_f}(\chi(t)), 
\end{align}
where $K_{xa}(\mu) = Q_\mu(x,a) - V_\mu(x)$ is the advantage function.
From the definitions of $Q_\mu$ and $V_\mu$, it is apparent that $\sum_a \mu(a | x) K_{x a}(\mu) =0$. Notice that the unique stable point of the ODE \eqref{eq:expcost-mu-ode-equiv} corresponds to an optimal policy, since otherwise, the advantage function is not zero; thus, one can establish that the trajectories of \eqref{eq:expcost-mu-ode-equiv} would converge to the set of optimal policies minimizing the exponential cost, with $\Lambda(\cdot)$ serving as a strict Lyapunov function. Finally, the policy update in \eqref{eq:mu-expcost-pg} used $\epsilon$-greedy exploration, which implies that the ODE tracked by this algorithm would not exactly coincide with \eqref{eq:expcost-mu-ode}. Instead, the algorithm tracks the ODE \eqref{eq:expcost-mu-ode} with an small error for small $\epsilon$, and hence one can claim that the iterate $\mu_n$ converges to a neighbourhood of the set of risk-optimal policies.

\section{Case 2: Discounted-cost/SSP + CPT as risk}
\label{sec:cpt-MDP-algo}
\diff{This case considers  the following optimization problem with CPT risk measure as the objective:} 
\boxed{
\begin{equation}
\min_{\theta \in \Theta} \C(D^\theta),
\label{eq:cpt-unconstrained-opt}
\end{equation}}
\diff{where $\C(D^\theta)$ is the CPT-value associated with the r.v. $D^\theta$, with $\theta$ denoting the policy parameter, and is defined as follows:
\begin{align}
\C(D^\theta) &\triangleq \intinfinity w^+\left(\Prob{u^+(D^\theta)>z}\right) dz  - \intinfinity w^-\left(\Prob{u^-(D^\theta)>z}\right) dz, \label{eq:cpt-general}
\end{align}
where $u^\pm$ are the utility functions and $w^\pm$ are the weight functions (see Section \ref{sec:risk-cpt} for a detailed description of these quantities).}
In a discounted-cost MDP, $D^\theta(x_0)$ would be the total discounted cost, while in an SSP, $D^\theta(x_0)$ would be the total cost. 

\diff{Risk measures such as variance, CVaR, and chance constraints considered in the previous chapter serve as natural constraint functions when optimizing the expected total/discounted cost. On the other hand, CPT is a risk measure that is not purely concerned with the tail behavior or variability of the cost distribution, as it considers the entire distribution, handling gains/losses, as well as incorporating distortions via a weight function. Hence, it is appealing to optimize the CPT value directly in the objective, e.g., in a human-centered decision-making problem. For a concrete example, one could consider a transportation application where $D^\theta$ denotes the delay experienced by a road user, and $\theta$ denotes a policy parameter that governs the traffic light switching strategy.}

From the discussion in the previous sections, it is apparent that the main technical challenges in handling any risk measure are as follows: (i) estimation of the risk measure from samples; and (ii) gradient estimation for the policy update iteration.
For the sake of brevity, we provide the necessary details for handling (i) and (ii), and the rest of the pieces of the resulting actor-critic scheme follows in a manner similar to that for variance or CVaR.

\subsection{CPT-value estimation}
To handle (i), suppose that we are given $m$ i.i.d. samples from the distribution of $X$, and the goal is to  estimate the CPT-value $\C(X)$. Estimating the CPT-value is challenging, because the environment provides samples from the distribution of the r.v. $X$, while the integrals in \eqref{eq:cpt-general} involve a weight-distorted distribution. \diff{Thus, unlike the case of expected value estimation, a sample mean is insufficient for estimating CPT-value $\C(X)$, when the underlying weight functions are nonlinear. CPT-value $\C(X)$ can be estimated if one has an estimate of the \textit{entire} distribution, and a natural candidate to estimate the distribution is the empirical distribution function (EDF).} Using the latter, we estimate $\C(X)$ by
\begin{align}
	\overline \C_m &= \intinfinity w^+\left(1-{\hat F_m}^+\left(x\right)\right)  dx  - \intinfinity w^-\left(1-{\hat F_m}^-\left(x\right)\right)  dx,~~~
	\label{eq:cpt-est}
\end{align}
where 
\diff{\begin{align*}
{\hat F_m}^+(x)&=\frac{1}{m} \sum_{i=1}^m \indic{(u^+(X_i) \leq x)} \textrm{ and }\\ 
{\hat F_n}^-(x)&=\frac{1}{m} \sum_{i=1}^m \indic{(u^-(X_i) \leq x)}.
\end{align*} 
${\hat F_m}^+(x)$ and ${\hat F_m}^-(x)$ are the EDFs of the r.v.s $u^+(X)$ and $u^-(X)$, respectively.  }
The first and second integrals on the RHS of (\ref{eq:cpt-est}) denoted by $\overline \C_m^+$ and $\overline \C_m^-$, respectively, can be computed in a straightforward fashion using the order statistics $X_{[1]}\le X_{[2]} \le \ldots \le X_{[m]}$ as follows:
\boxed{
\vspace*{-10pt}
	\begin{align*}
		\overline \C_m^+&:=\sum_{i=1}^{m} u^+(X_{[i]}) \left(w^+\!\left(\frac{m+1-i}{m}\right)\!-\! w^+\!\left(\frac{m-i}{m}\right) \right),\\
		\overline \C_m^-&:=\sum_{i=1}^{m} u^-(X_{[i]}) \left(w^-\left(\frac{i}{m}\right)- w^-\left(\frac{i-1}{m}\right) \right). 
\end{align*}}
\diff{Notice that the estimates $\overline \C_m^{\pm}$ reduce to  sample means for the case when $w(p)=p$, and in this case, the CPT-value itself is the expectation $\E u^+(X) - \E u^-(X)$. Thus, CPT-value estimation can be seen as a generalization of the classic mean estimation procedure, and the deviations are introduced by a nonlinear weight function that distorts probabilities, which in turn leads to weighing the samples non-uniformly.}

\subsection{Policy gradient for the CPT-value}
As far as handling point (ii) concerning the policy gradient for CPT, we use SPSA, since the CPT-value does not admit a Bellman equation, ruling out a procedure based on the likelihood ratio method. 
The SPSA-based estimate of $\nabla \C(X^{\theta_n})$ with policy $\theta_n$, is given as follows:   
\boxed{
\vspace*{-10pt}
	\begin{align}
		\widehat \nabla_{i} \C(D^\theta) = \dfrac{\overline \C_n^{\theta_n+\delta_n \Delta(n)} - \overline \C_n^{\theta_n}}{ \delta_n \Delta_i(n)}, i=1,\ldots,d,
	\label{eq:grad-est-spsa}
\end{align}}
where $\delta_n$ and $\Delta(n)$ are as described in Section \ref{sec:spsa} and $\overline \C_n^{\theta_n+\delta_n \Delta(n)}$ (resp. $\overline \C_n^{\theta_n}$) denotes the CPT-value estimate that uses $m_n$ samples of the r.v. $X^{\theta_n+\delta_n \Delta(n)}$ (resp. $X^{\theta_n}$).

The complete algorithm with CPT-value as the risk measure and the usual value function as the objective is presented in Algorithm \ref{alg:cpt-PG}.
\begin{algorithm}
	\SetKwInOut{Input}{Input}\SetKwInOut{Output}{Output}
	\Input{initial parameter $\theta_0 \in \Theta$, 
		perturbation constants $\delta_n>0$, \diff{batch sizes} $\{m_n\}$, step sizes $\{\zeta(n)\}$, projection operator $\Gamma$, number of iterations $M\gg 1$.}
	\For{$n\leftarrow 0$ \KwTo $M-1$}{
		\For{$m\leftarrow 0$ \KwTo $m_n-1$}{
			\tcc{Unperturbed policy simulation}
			Use the policy $\mu^{\theta_n}$ to generate the state $x_m$, draw action $a_m\sim\mu^{\theta_n}\left(\cdot | x_m\right)$\;
			Observe next state $x_{m+1}$ and cost $k(x_m,a_m)$\;
			\tcc{Perturbed policy simulation}
			Use the policy $\mu^{\theta_n+\delta_n\Delta(n)}$ to generate the state $x_m^+$, draw action $a_m^+\sim\mu^{\theta_n+\delta_n\Delta(n)}\left(\cdot | x_m^+\right)$\;
			Observe next state $x^+_{m+1}$ and cost  $k(x^+_m,a^+_m)$\;
		}
		\tcc{Monte Carlo policy evaluation}
		Use the scheme in  \eqref{eq:cpt-est} to obtain $\overline \C_n^{\theta_n+\delta_n \Delta(n)}$ and $\overline \C_n^{\theta_n)}$ - the estimates of the CPT-values $\C\left(X^{\theta_n+\delta_n \Delta(n)}\right)$ and $\C\left(X^{\theta_n}\right)$, respectively\;
		
		\tcc{Gradient estimates using SPSA}
		Gradient of the objective: $\widehat \nabla_{i} \C(X^{\theta_n}) = \dfrac{\overline \C_n^{\theta_n+\delta_n \Delta(n)} - \overline \C_n^{\theta_n}}{ \delta_n \Delta_i(n)}$\;
		\tcc{Policy update: Gradient descent using SPSA}
		\begin{align}
			\theta_{n+1} = \Gamma\bigg[\theta_n - \zeta(n)  \bigg(\widehat \nabla \C(X^{\theta_n})\bigg)\bigg].\label{eq:cpt-pg-update}
		\end{align}
	}
	\Output{Policy $\theta_M$} 
	\caption{Policy gradient algorithm under CPT as a risk measure}
	\label{alg:cpt-PG}
\end{algorithm}

\diff{
\subsection{On the batch size $m_n$ per iteration of \eqref{eq:cpt-pg-update}}
\label{sec:mn}

The challenge involved in choosing an appropriate batch size $m_n$ for policy evaluation in Step 2 of Algorithm \ref{alg:cpt-PG} is similar to that in Algorithm \ref{alg:discounted-PG} for optimizing variance as a constraint in a discounted MDP. As in the variance case, the CPT-value has to be estimated from sample trajectories so that the overall policy gradient algorithm \eqref{eq:cpt-pg-update} converges. For the sake of simplicity, we drop the dependence on the parameter $\theta$, and instead, study the CPT-value estimation problem. Subsequently, when we analyze the policy gradient scheme in Algorithm \ref{alg:cpt-PG} for CPT-value optimization, we shall make the dependence on the policy parameter explicit.

For a given r.v. $X$, let $m$ denote the number of sample trajectories used to form the estimate $\overline \C_m$, using \eqref{eq:cpt-est}, of the CPT-value $\C(X)$.  
Notice that $\E\left(\overline \C_m\right) \ne \C(X)$, since the individual components $\overline \C_m^{\pm}$ involve order statistics. However, one can derive a bound on the estimation error $\left|\overline \C_m -C(X)\right|$, and such a bound would aid the proof of asymptotic convergence of the risk-sensitive policy gradient algorithm for CPT. The following result presents a bound in expectation for the estimation error.

\bigskip
\boxed{
\vspace*{-10pt}
\begin{proposition}
\label{prop:cpt-est}
Assume that the weight functions $w^{\pm}$ are H\"{o}lder continuous with common order $\alpha$ and constant $H$, i.e., 
	\[\sup_{x \neq y} \frac{| w^{\pm}(x) - w^{\pm}(y) |}{| x-y |^{\alpha}} \leq H, \forall x,y \in [0,1].\]
Suppose that the utility functions $u^+$ and $-u^-$ are continuous and non-decreasing on their support $\R^+$ and $\R^-$, respectively. Furthermore, the utilities $u^+(X)$ and $u^-(X)$  are bounded by a constant $M$. Then, $\forall \epsilon >0$, we have
\begin{align}
\Prob{\left| \overline \C_n - \C(X) \right|\geq  \epsilon}\leq
2e^{-2n\left(\frac{\epsilon}{HM}\right)^{\frac{2}{\alpha}}},
\label{eq:S-complex-bounded}
\end{align}	
and
\begin{align}
	\E \left|\overline \C_n- \C(X) \right|  \le    \frac{\left(8HM\right) \Gamma\left(\alpha/2\right)}{n^{\alpha/2}}.\label{eq:cpt-expec-bd}
\end{align}
\end{proposition}
}
\begin{proof}
The proof requires the well-known Dvoretzky–Kiefer–Wolfowitz (DKW) inequality, which shows concentration of the empirical distribution function around the true distribution. We recall this result below.

\textbf{\textit{DKW inequality:}} Let $F$ denote the cdf of r.v. $U$ and ${\hat F_n}(u)=\frac{1}{n} \sum_{i=1}^n I_{\left[U_i \leq u\right]}$ denote the empirical distribution of $U$, with $U_1,\ldots,U_n$ sampled from $F$.
Then, for any $\epsilon>0$, we have
\[
\Prob{\sup_{x\in \mathbb{R}}|\hat{F}_n(x)-F(x)|>\epsilon } \leq 2 e^{-2n\epsilon^2}.
\]

Notice that
\begin{align*}
&\left|\int_0^{\infty}\!\!\!\! w^+\left(\Prob{u^+(X)>t}\right) dt- \int_0^{\infty} \!\!\!\! w^+\left(1- {\hat F^+_n}(t)\right) dt\right| \\ = &
    \left|\int_0^{M} w^+\left(\Prob{u^+(X)>t}\right) dt- \int_0^{M} w^+\left(1- {\hat F^+_n}(t)\right) dt\right| \\
    &\leq  HM \sup_{x\in\mathbb{R}}\left|\Prob{u^+(X)<t}-{\hat F^+_n}(t)\right|^\alpha.
\end{align*}
Now, plugging in the DKW inequality, we obtain
\begin{align}
&
P\left(\left|\int_0^{\infty}\!\!\!\! w^+\left(\Prob{u^+(X)>t}\right) dt- \int_0^{\infty} \!\!\!\! w^+\left(1- {\hat F^+_n}(t)\right) dt\right|>\epsilon\right)
\nonumber\\
&
\leq
 P\left(HM\sup_{t\in \mathbb{R}} \left|(P(u^+(X)<t)-{\hat F_n}^+(t)\right|^\alpha>\epsilon\right) \leq  2e^{-2n\left(\frac{\epsilon}{HM}\right)^{\frac{2}{\alpha}}}.~~~~~~~~~
 \label{eq:dkw1}
\end{align}

To derive the bound in expectation in \eqref{eq:cpt-expec-bd}, we integrate the high-probability bound \eqref{eq:dkw1} to obtain
 \begin{align*}
& \E \left|\overline \C_n- \C(X) \right|  
  \le \intinfinity \Prob{\left|\overline \C_n- \C(X) \right| \geq  \epsilon} d\epsilon\\
  &\le 4 \intinfinity \exp\left(-2n\left(\epsilon/HM\right)^{2/\alpha}\right) d\epsilon \le \frac{8HM \Gamma\left(\alpha/2\right)}{n^{\alpha/2}}.
 \end{align*}
\end{proof}
Now, the discussion in Section \ref{sec:mn} applies to the case of CPT-value, with a minor changes. In particular,  from Proposition \ref{prop:cpt-est}, the estimation bias for the case of CPT-value is of the order $\frac{1}{m^{\alpha/2}}$, where $\alpha$ is the H\"{o}lder exponent of the weight functions underlying CPT-value definition. 

Following arguments similar to those employed in Section \ref{sec:mn}, using a SPSA-based gradient estimator for CPT-value would require the batch size $m_n$ to diverge so that the estimation bias does not affect convergence of policy gradient algorithm \eqref{eq:cpt-pg-update}.
In addition to the usual conditions on the step-size sequence and perturbation constant $\delta_n$, one possible choice for $m_n$ that ensures that the bias in the gradient estimate vanishes and the overall algorithm converges is the following:
$\frac{1}{m_n^{\alpha/2}\delta_n}\rightarrow 0$. 


}

\section{Case 3: Any MDP + a coherent risk measure}
\label{sec:coherent-algo}
In this section, we consider optimizing a coherent risk measure. Let 
$(\Omega,\F, \P_\theta)$ denote a probability space, 
where $\P_\theta$ denotes a parameterized probability measure, with $\theta$ as the parameter that belongs to a convex and compact set $\Theta \subset \R^d$. 
Let $D$ be a r.v. with a finite mean, and let $\rho(D)$ denote its coherent risk measure. We consider the following problem:
\boxed{
\begin{equation}
\min_{\theta \in \Theta} \rho(D),
\label{eq:coherent-unconstrained-opt}
\end{equation}}
As in the previous section, $D$ could be the total/discounted cost of the policy parameterized by $\theta$.

The overall algorithm for optimizing the coherent risk measure is given in Algorithm \ref{alg:coherent-PG}. The schema of this algorithm resembles the one used in Algorithm \ref{alg:cpt-PG} for optimizing CPT value, the difference being in the way the coherent risk measure is estimated, and its gradient computed. We elaborate on these two aspects below.

For coherent risk measure estimation, we require the dual representation.  Let $\E \left(D\right) = \int_\Omega  D(\omega) d\P_\theta(\omega)$ denote the expectation of a given r.v. $D$. 
Let 
$\mathfrak{P} = \{\xi \mid \int  \xi d \P_\theta = 1\}$ denote the set of probability densities.
Then there exists a convex and compact subset $\mathfrak{U}$ of $\mathfrak{P}$ such that 
\boxed{
\begin{align}
    \rho(D) = \sup_{\xi \in \mathfrak{U}} \left\{ \E \left(\xi D\right) = \int_\Omega \xi(\omega) D(\omega) d\P_\theta(\omega) \right\}.\label{eq:coherent-dual}
\end{align} 
}
We shall refer to the set $\mathfrak{U}$ as the risk envelope associated with a coherent risk measure.
For the case of CVaR, the risk envelope $\mathfrak{U}$ can be shown to be 
\begin{align} 
\mathfrak{U} = \left\{\xi\; \middle|\; \xi(\omega)\in \left[0,\frac{1}{1-\beta}\right], \E[\xi] =1 \right\}.
\label{eq:cvar-envelope}
\end{align}

Let $\P_{n,\theta}$ denote the empirical distribution function formed from $n$ i.i.d. samples $\{\omega_1,\ldots,\omega_n\}$. Then, the estimate $\hat\rho_n$ of the coherent risk measure $\rho(D)$ is formed as follows:
\begin{align}
    \hat\rho_n = \sup_{\xi \in \mathfrak{U}} \sum_{i=1}^n \xi(\omega_i) D(\omega_i) \P_{n,\theta}(\omega_i).\label{eq:rho-est}
\end{align} 

Next, we turn to the estimate of the gradient of a coherent risk measure. 
\diff{For the purpose of gradient estimation, we shall assume the following  form for the risk envelope:  
\begin{align}
&\mathfrak U(\P_\theta)=\left\{\xi\;\middle|\;g_{k_1}(\xi,\P_\theta)=0,\;k_1=1,\ldots,K_1,
\nonumber\right.\\
&\qquad\qquad\qquad\qquad\left.\;f_{k_2}(\xi,\P_\theta)\leq 0,\;k_2=1,\ldots,K_2,\;\E[\xi]=1,\;\xi(\omega)\geq 0\right\},\label{eq:u-form}
\end{align}
where $g_{k_1}$, $k_1=1,\ldots,K_1$ and $f_{k_2}$, $k_2=1,\ldots,K_2$ are the equality and inequality constraints. 
Using the above form for the risk envelope, the Lagrangian of \eqref{eq:coherent-dual}
turns out to be 
\begin{align}
L_\theta(\xi,\eta,\lambda,\tilde\lambda)
&= \E\left[\xi D\right] -\eta\left(\E\left[\xi\right]-1\right)
-\sum_{k_1=1}^{K_1}\lambda(k_1) g_{k_1}(\xi,\!\P_\theta)\nonumber\\
&\qquad-\sum_{k_2=1}^{K_2}\tilde\lambda(k_2) f_{k_2}(\xi,\!\P_\theta),\label{eq:Lagrangian}
\end{align}
where $\eta$ is the Lagrange multiplier associated with the $\E[\xi]=1$ constraint. Furthermore, $\lambda=(\lambda(1),\ldots,\lambda(K_1))$ and $\tilde\lambda=(\tilde\lambda(1),\ldots,\tilde\lambda(K_2))$ are the Lagrange multipliers associated with equality constraints defined by $(g_1,\ldots,g_{K_1})$, and inequality constraints defined by $(f_1,\ldots,f_{K_2})$, respectively.}

\diff{
We now present the expression for the gradient of the coherent risk measure $\rho(X)$.  For deriving this expression, we make the following assumptions:

\begin{assumption}
	\label{ass:convex}
	The constraints $g_{k_1}$ is an affine function of the parameter $\xi$ for $k_1=1,\ldots,K_1$, and $f_{k_2}$ is a convex function of the parameter $\xi$ for $k_2=1,\ldots,K_2$. 
\end{assumption}
\begin{assumption} 
	\label{ass:feasible}
	There exists a strictly feasible point for the problem in \eqref{eq:coherent-unconstrained-opt}. 
\end{assumption}

\begin{assumption}
	\label{ass:equidiff} 
	The family of functions $\{L_\theta(\xi,\eta,\lambda,\tilde\lambda)\}_{\xi,\eta,\lambda,\tilde\lambda}$ is equi-differentiable in $\theta$\footnote{A family $\{f(x,\cdot)\}_{x\in \X}$ is equi-differentiable at $p\in [0,1]$ if $\frac{(f(x,t')-f(x,t))}{t'-t}$ converges uniformly as $t' \rightarrow t$.}.
\end{assumption}

We now discuss these assumptions. The motivation for \ref{ass:convex} comes from the result that a risk measure $\rho(D)$ is coherent if and only if the underlying risk envelope $\mathfrak{U}$ is convex and weakly compact. The conditions on $g_{k_1}$ and $f_{k_2}$ can be inferred from the convexity requirement on $\mathfrak{U}$. \ref{ass:feasible} ensures strong duality holds for \eqref{eq:coherent-unconstrained-opt}, which in turn implies $\max_\xi \min_{\eta,\lambda,\tilde\lambda} L_\theta(\cdot,\cdot,\cdot,\cdot)=\min_{\eta,\lambda,\tilde\lambda} \max_\xi  L_\theta(\cdot,\cdot,\cdot,\cdot)$. This interchange facilitates the application of the envelope theorem. For the application of the latter theorem, we also require the equi-differentiability condition imposed in \ref{ass:equidiff}.
\boxed{\begin{proposition}
Assume \ref{ass:convex}--\ref{ass:equidiff}.
Let $\lambda_{\theta}^{*}=(\lambda^{*}_{\theta}(1),\ldots,\lambda^{*}_{\theta}(K_1))$,  $\tilde\lambda^{*}_{\theta} =(\tilde\lambda^{*}_{\theta}(1),\ldots,\tilde\lambda^{*}_{\theta}(K_2))$,
and let $(\xi^*_{\theta},\eta^*_{\theta}, \lambda^{*}_{\theta},\tilde\lambda^{*}_{\theta})$ denote a saddle point of~\eqref{eq:Lagrangian}. Then,
\begin{align}
  \nabla\rho(D) &= \E_{\xi_\theta^*}\left[\nabla \log \P_\theta(\omega) (D - \eta^{*}_{\theta})\right]
     - \sum_{k_1=1}^{K_1} \lambda^{*}_{\theta}(k_1) \nabla g_{k_1}(\xi^*_{\theta},\P_\theta)\nonumber\\
     &\qquad\quad-\sum_{k_2=1}^{K_2} \tilde\lambda^{*}_{\theta}(k_2) \nabla f_{k_2}(\xi^*_{\theta},\P_\theta).\label{eq:rho-grad}
\end{align}
\end{proposition}}
\begin{proof}
\ref{ass:feasible} implies  Slater's condition holds for the problem \eqref{eq:coherent-unconstrained-opt}. Furthermore, by \ref{ass:convex}, we have that the Lagrangian $L_\theta(\xi,\eta,\lambda,\tilde\lambda)$ is convex in $\xi$, and concave in the Lagrange multipliers $\eta,\lambda$, and $\tilde\lambda$. Thus, strong duality holds, implying
\[ \max_{\xi \ge 0} \min_{\eta, \lambda, \tilde \lambda \ge 0} L_\theta(\xi,\eta,\lambda,\tilde\lambda) 
=  \min_{\eta, \lambda, \tilde \lambda \ge 0}\max_{\xi \ge 0} L_\theta(\xi,\eta,\lambda,\tilde\lambda).\]
Since the family $\{L_\theta(\xi,\eta,\lambda,\tilde\lambda)\}_{\xi,\eta,\lambda,\tilde\lambda}$ is equi-differentiable by assumption \ref{ass:equidiff}, and $L_\theta(\xi,\eta,\lambda,\tilde\lambda)$ is a smooth function of $\theta$, we obtain the following by an application of the envelope theorem:
\begin{align*}
    &\nabla\max_{\xi \ge 0} \min_{\eta, \lambda_1,\ldots, \lambda, \tilde \lambda \ge 0} L_\theta(\xi,\eta,\lambda,\tilde\lambda) \\
    &= \nabla L_\theta(\xi^*_\theta,\eta^*_\theta,\lambda^*_\theta,\tilde\lambda^*_\theta)\\
    &= \E_{\xi_\theta^*}\left[\nabla \log \P_\theta(\omega) (D - \eta^{*}_{\theta})\right]
     - \sum_{k_1=1}^{K_1} \lambda^{*}_{\theta}(k_1) \nabla g_{k_1}(\xi^*_{\theta},\P_\theta)\nonumber\\
     &\qquad\quad-\sum_{k_2=1}^{K_2} \tilde\lambda^{*}_{\theta}(k_2) \nabla f_{k_2}(\xi^*_{\theta},\P_\theta),
\end{align*}
where the final equality uses the following fact:
\[ \nabla\E\left[\xi D\right] -\eta\nabla\left(\E\left[\xi\right]-1\right) = \E_{\xi}\left[\nabla \log \P_\theta(\omega) (D - \eta)\right].\]
The equality above can be inferred using the likelihood ratio method. To elaborate for the discrete case, 
notice that 
\begin{align*}
\nabla\E\left[\xi D\right] &= \sum_\omega \xi(\omega)  D(\omega)\nabla P_\theta(\omega) \\
&= \sum_\omega \xi(\omega)  D(\omega)\nabla \log P_\theta(\omega) P_\theta(\omega) \\
&= \E_{\xi}\left[\nabla \log \P_\theta(\omega) D\right].
\end{align*}
A similar argument works for the other term involving the $\eta$ factor above.
\end{proof}
}
\diff{
We now specialize the expression derived above for the policy gradient of a coherent risk measure to the case of CVaR. Recall that the risk envelope for CVaR is given by $\mathfrak{U} = \left\{\xi \mid \xi(\omega)\in \left[0,\frac{1}{1-\beta}\right], \E[\xi] =1 \right\}.$
Thus, $\lambda^*_\theta$ and $\tilde\lambda^*_\theta$ are both zero vectors, leading to 
\begin{align*}
  \nabla\text{CVaR}_{\alpha}(D) &= \E_{\xi_\theta^*}\left[\nabla \log \P_\theta(\omega) (D - \eta^{*}_{\theta})\right].
  \end{align*}
 It can be shown that $\xi_\theta^*=\frac{1}{1-\beta}$ for $X>\text{VaR}_{\alpha}(D)$, and $\xi_\theta^*=0$ otherwise. Thus, we have 
 \begin{align}
     \nabla\text{CVaR}_{\alpha}(D) &= \E\left[\nabla \log \P_\theta(\omega) (D - \text{VaR}_{\alpha}(D))\mid D > \text{VaR}_{\alpha}(D)\right].~~~~~~~
     \label{eq:cvar-grad-special}
 \end{align}
}
To obtain an estimate of $\nabla\rho(D^\theta)$, we require an estimate of $\rho(D^\theta)$, which is obtained by solving the convex optimization problem in \eqref{eq:rho-est}. Let $\xi^*_{n,\theta}, \eta^*_{n,\theta}, \lambda^*_{n,\theta}, \tilde \lambda^*_{n,\theta}$ denote the optimal parameter and Lagrange multipliers obtained by solving the estimation problem in \eqref{eq:rho-est}. \diff{Using these quantities, the gradient estimate $\widehat \nabla \rho(D)$ is formed as follows:
\begin{align}
&  \widehat \nabla_{n}\rho(D) = \sum_{i=1}^n\left[ \xi^{*}_{n,\theta}(\omega_i) \P_{n,\theta}(\omega_i) \nabla \log \P_{n,\theta}(\omega_i)  (D(\omega_i) - \eta^{*}_{n,\theta})\right]\nonumber\\
& \hspace*{-3pt} - \hspace*{-3pt} \sum_{k_1=1}^{K_1} \lambda^{*}_{n,\theta}(k_1) \nabla g_{k_1}(\xi^*_{n,\theta},\P_{n,\theta})
     -\sum_{k_2=1}^{K_2} \tilde\lambda^{*}_{n,\theta}(k_2) \nabla f_{k_2}(\xi^*_{n,\theta},\P_{n,\theta}).~~~~~~~
     \label{eq:rho-grad-est}
\end{align}}
%
The complete algorithm using a coherent risk measure is presented in Algorithm \ref{alg:coherent-PG}.

\begin{algorithm}
	\SetKwInOut{Input}{Input}\SetKwInOut{Output}{Output}
	\Input{initial parameter $\theta_0 \in \Theta$, 
		perturbation constants $\delta_n>0$, trajectory lengths $\{m_n\}$, step sizes $\{\zeta(n)\}$, projection operator $\Gamma$, number of iterations $M\gg 1$.}
	\For{$n\leftarrow 0$ \KwTo $M-1$}{
		\For{$m\leftarrow 0$ \KwTo $m_n-1$}{
			Use the policy $\mu^{\theta_n}$ to generate the state $x_m$, draw action $a_m\sim\mu^{\theta_n}\left(\cdot | x_m\right)$\;
			Observe next state $x_{m+1}$ and cost $k(x_m,a_m)$\;
		}
		\tcc{Monte Carlo policy evaluation}
		Use the scheme in  \eqref{eq:rho-est} to obtain the estimate $\hat \rho_n$ of the coherent risk measure\;
		
		\tcc{Gradient estimate using likelihood ratio}
		Form the gradient estimate $\widehat \nabla_{n,\theta}\ \rho(D)$  using \eqref{eq:rho-grad-est} \;
		\tcc{Policy update: Gradient descent}
		$\theta_{n+1} = \Gamma\bigg[\theta_n - \zeta(n)  \bigg(\widehat \nabla_{n,\theta}\ \rho(D)\bigg)\bigg]$\;
	}
	\Output{Policy $\theta_M$} 
	\caption{Policy gradient algorithm under a coherent risk measure}
	\label{alg:coherent-PG}
\end{algorithm}

\clearpage
\section{Bibliographic remarks}
\diff{
\vspace*{-6pt}
We provide below bibliographic remarks for each case studied in this chapter.
\vspace*{-4pt}
\begin{description}
	\item[\ref{sec:expcost-algo}] 
The theory of risk-sensitive control using an exponential utility formulation has a long history; see \textcite{whittle1990risk} for a detailed introduction. 
However, work on the learning side of things is a more recent development; see,  
e.g. \textcite{borkar2001sensitivity,borkar2002q,borkar2002risk,bhatnagar06a,Basu08LA}, and the survey article by \textcite{borkar2010learning}. Our presentation of the policy gradient theorem and the algorithm for optimizing the exponential cost in an average-cost MDP is based on \textcite{borkar2001sensitivity}, where the author provides a sketch of the convergence analysis. For the missing details, the reader is referred to the analysis of a two-timescale policy gradient algorithm in a risk-neutral setting in \textcite{konda1999actor}, in particular, Lemmas 5.6--5.7, and Theorem 5.8 there. More recently, in  \textcite{moharrami2022policy}, the authors propose a policy gradient algorithm for solving a truncated version of the exponential cost MDP; see also Proposition 5 in this work for a variational formula for exponential cost, which establishes stability of the latter risk measure against model uncertainties. 

	\item[\ref{sec:cpt-MDP-algo}]
The presentation of the risk-sensitive RL algorithm with CPT as the underlying risk measure is based on \textcite{prashanth2015cumulative,prashanth2018cpt}. In Proposition \ref{prop:cpt-est}, we presented a concentration bound for CPT-value estimation assuming that the underlying distribution has bounded support. Recent work in \textcite{bhat2019concentration} provides more general concentration bound results assuming sub-Gaussian and sub-exponential distributions. 

\item[\ref{sec:coherent-algo}] The case of coherent risk measure is based on
\textcite{tamar2015coherent}. For an introduction to coherent risk measures and their dual representation, the reader is referred to  \textcite[Section 6.3]{shapiro2014lectures}. In particular, the risk envelope for CVaR presented in \eqref{eq:cvar-envelope} and the justification for the saddle point $\xi_\theta^*$ leading to \eqref{eq:cvar-grad-special} are based on Example 6.16 of \textcite{shapiro2014lectures}.
\end{description}}
\chapter{Conclusions and Future Challenges}
\label{sec:conclusions}
In this book, we considered MDP problems that incorporate a variety of risk measures in discounted-cost, average-cost, and SSP settings.  The risk measures considered were variance (both total and per period), CVaR, chance constraints, CPT, and coherent risk measures. Challenges encountered in the various problem settings include the following: (i) lack of structure, leading to failure of classic DP methods (e.g., policy iteration for variance-constrained MDPs); (ii) lack of gradient information for the risk measures; and (iii) challenges in estimation, in the case of CVaR and CPT.

We briefly summarize some possible future research directions for risk-sensitive MDPs:
\begin{enumerate}[label=\textbf{\Roman*})]
\item  For the discounted MDP setting, the current algorithm employs SPSA only, because a direct gradient estimate cannot be easily obtained for the risk measure. However, a likelihood ratio gradient estimator is available for the cost function, so combining SPSA with direct gradient-based search in a hybrid algorithm might improve the computational efficiency of the algorithm for work along this line (but in the general SA setting, not specific to MDPs). 
Even more critical is that SPSA requires simulation of two system trajectories, which might be infeasible in real-time or online settings, so developing a risk-sensitive algorithm that uses only a single trajectory is of practical interest.
 \item For CVaR-constrained MDPs, variance reduction techniques such as importance sampling and conditional Monte Carlo 
 are essential for keeping CVaR estimation variance at reasonable levels, and as far as we are aware, there is no such \textit{provably convergent} CVaR-estimation algorithm in an RL context. Another important need is incorporating function approximation to handle the curse of dimensionality for large state spaces.
\item A critical challenge is to obtain finite-time bounds for the risk-sensitive RL algorithms, which usually operate on multiple timescales. To the best of our knowledge, there are no non-asymptotic bounds available for multi-timescale stochastic approximation schemes, and hence, for actor-critic algorithms, even in the risk-neutral RL setting.
\item Risk measures that have not been explored in an RL context include spectral risk measures (SRMs) and utility-based shortfall risk (UBSR); see the bibliographic remarks at the end of Chapter \ref{sec:risk-measures} for references. The algorithm presented in Section \ref{sec:coherent-algo} handles a general coherent risk measure, so could in principle be specialized to handle an SRM, but a direct algorithm might be more efficient. On the other hand, a risk-sensitive RL algorithm with either UBSR, or more generally, a convex risk measure as the objective/constraint, has not been developed in the literature as far as we are aware.
\item Finally, CPT-value of the return of an MDP does not have a Bellman equation. 
Here, we proposed treating the CPT-value MDP problem as a black-box stochastic optimization problem, 
but certainly other approaches, especially ones that exploit special structure such as the Markovian property of an MDP, might be more computationally efficient in certain contexts. 
\end{enumerate}

\backmatter  

\printbibliography

\end{document}